\documentclass[10pt,a4paper]{article}
\usepackage[utf8]{inputenc}

\usepackage{mathrsfs,amsthm,xcolor,verbatim,bbm,amsmath,amsfonts,amssymb,nicefrac,enumitem,hyperref,bm,mathtools,xparse,etoolbox}
\usepackage{cite}

\usepackage{thmtools}
\usepackage[margin=1in]{geometry}
\usepackage[capitalise,nameinlink]{cleveref} 
\usepackage{tikz}
\usetikzlibrary{matrix,chains,positioning,decorations.pathreplacing,arrows,shapes,math,arrows.meta,calc, shapes.geometric}
\usepackage[most]{tcolorbox}
\usepackage{footnote}


\tikzset{
    neuron/.style={circle, draw, minimum size=0.5cm, text centered},
    input neuron/.style={neuron, fill=yellow!30},
    hidden neuron/.style={neuron, fill=blue!10, draw=blue, thick},
    output neuron/.style={neuron, fill=green!20},
    conn/.style={-stealth, thick},
    ellipsis/.style={font=\footnotesize},
}
\usepackage{capt-of}
\usepackage{stmaryrd} 

\crefname{enumi}{item}{items}
\crefname{equation}{}{}
\crefname{subsection}{Subsection}{Subsections}
\crefname{figure}{Figure}{Figures}

\usepackage{xurl}

\hypersetup{
    colorlinks,
    linkcolor={red!80!black},
    citecolor={green},
    urlcolor={blue!80!black}
}

\theoremstyle{plain}
\newtheorem{theorem}{Theorem}[section]

\newtheorem{prop}[theorem]{Proposition}

\newtheorem{definition}[theorem]{Definition}



\theoremstyle{definition}

\DeclareMathAlphabet{\mathpzc}{OT1}{pzc}{m}{it}

\DeclareFontEncoding{LS1}{}{}
\DeclareFontSubstitution{LS1}{stix}{m}{n}
\DeclareMathAlphabet{\mathscr}{LS1}{stixscr}{m}{n}

\newcommand{\E}{\mathbb{E}}
\renewcommand{\P}{\mathbb{P}}

\newcommand{\R}{\mathbb{R}}
\newcommand{\N}{\mathbb{N}}

\newcommand{\smalll}{\mathbb{L}}

\renewcommand{\d}{ \mathrm{d}}
\newcommand{\numofsum}{v}
\newcommand{\dimofsum}{w}

\renewcommand{\c}[1]{\mathfrak{c}^{#1}}

\newcommand{\cA}{\mathcal{A}}

\newcommand{\cF}{\mathcal{F}}
\newcommand{\cG}{\mathcal{G}}

\newcommand{\cK}{\mathcal{K}}
\newcommand{\cL}{\mathcal{L}}

\newcommand{\cV}{\mathcal{V}}

\newcommand{\bfm}{\mathbf{m}}

\newcommand{\bfx}{\mathbf{x}}

\newcommand{\globalmin}{\vartheta}

\newcommand{\bfA}{\mathbf{A}}

\newcommand{\bfM}{\mathbf{M}}

\newcommand{\bfV}{\mathbf{V}}

\newcommand{\bfX}{\mathbf{X}}

\newcommand{\scrc}{\mathscr{c}}

\newcommand{\g}{\|\mathfrak{g}\|_{Lip}}

\newcommand{\fC}{\mathfrak{C}}

\newcommand{\fR}{\mathfrak{R}}

\newcommand{\bbX}{\mathbb{X}}

\newcommand{\fd}{d}

\newcommand{\fn}{\mathfrak{n}}

\newcommand{\fr}{\mathfrak{r}}

\newcommand{\dimX}{\mathscr{d}}

\newcommand{\para}{\beta}

\newcommand{\setX}{K}

\newcommand{\bbV}{\mathbb{V}}

\newcommand\restr[2]{{
  \left.\kern-\nulldelimiterspace 
  #1 
  \vphantom{|} 
  \right|_{#2} 
  }}

\DeclarePairedDelimiter{\spro}{\langle}{\rangle}

\newcommand{\ceil}[1]{ \left\lceil #1 \right\rceil}

\newcommand{\qqandqq}{\qquad\text{and}\qquad}

\newcommand{\democrator}{\mathbb{V}}
\newcommand{\mandV}[4]{\Theta^{#1,#2,#3}_{#4}}
\newcommand{\mandVcom}[5]{\Theta^{#1,#2,#3}_{#4,#5}}

\usepackage{cleveref}

\ExplSyntaxOn

\seq_new:N \g_cflist_loaded
\seq_new:N \g_cflist_pending

\NewDocumentCommand{\cfadd} { m } {
  \seq_if_in:NnF \g_cflist_loaded { #1 } {
    \seq_if_in:NnF \g_cflist_pending { #1 } {
      \seq_gput_right:Nn \g_cflist_pending { #1 }
    }
  }
}

\NewDocumentCommand{\cfconsiderloaded} { m } {
  \seq_gput_right:Nn \g_cflist_loaded {#1}
}

\NewDocumentCommand{\cfremove} { m } {
  \seq_gremove_all:Nn \g_cflist_pending { #1 }
}

\NewDocumentCommand{\cfload} { o } {
  \seq_if_empty:NTF \g_cflist_pending {
    \IfValueTF{#1}{\ignorespaces}{\unskip}
  } {
    (cf.\ \cref{\seq_use:Nn \g_cflist_pending {,}})\IfValueTF{#1}{#1~}{\unskip}
    \seq_gconcat:NNN \g_cflist_loaded \g_cflist_loaded \g_cflist_pending
    \seq_gclear:N \g_cflist_pending
    \IfValueT{#1}{\ignorespaces}
  }
}

\NewDocumentCommand{\cfclear} {} {
  \seq_gclear:N \g_cflist_loaded
  \seq_gclear:N \g_cflist_pending
}

\NewDocumentCommand{\cfout} { o } {
  \seq_if_empty:NTF \g_cflist_pending {\unskip\IfValueT{#1}{\ignorespaces}} {
    (cf.\ \cref{\seq_use:Nn \g_cflist_pending {,}})\IfValueTF{#1}{#1~}{\unskip}
    \seq_gclear:N \g_cflist_pending
    \IfValueT{#1}{\ignorespaces}
  }
}

\NewDocumentCommand{\ifnocf} { m } {
  \seq_if_empty:NT \g_cflist_pending { #1 }
}

\ExplSyntaxOff

\ExplSyntaxOn

\bool_new:N \g_noteobserve

\NewDocumentCommand{\setnote}{}{
  \bool_gset_true:N \g_noteobserve
}

\NewDocumentCommand{\setobserve}{}{
  \bool_gset_false:N \g_noteobserve
}

\NewDocumentCommand{\nobs}{ o }{
  \IfValueT{#1}{
    \str_if_eq:noTF {note} {#1} {
      \bool_gset_true:N \g_noteobserve
    } {
      \str_if_eq:noTF {Note} {#1} {
        \bool_gset_true:N \g_noteobserve
      } {
        \bool_gset_false:N \g_noteobserve
      }
    }
  }
  \bool_if:nTF { \g_noteobserve } {
    \bool_gset_false:N \g_noteobserve
    note
  } {
    \bool_gset_true:N \g_noteobserve
    observe
  }
  \IfValueF{#1}{~}
}

\NewDocumentCommand{\Nobs}{ o }{
  \IfValueT{#1}{
    \str_if_eq:noTF {note} {#1} {
      \bool_gset_true:N \g_noteobserve
    } {
      \str_if_eq:noTF {Note} {#1} {
        \bool_gset_true:N \g_noteobserve
      } {
        \bool_gset_false:N \g_noteobserve
      }
    }
  }
  \bool_if:nTF { \g_noteobserve } {
    \bool_gset_false:N \g_noteobserve
    Note
  } {
    \bool_gset_true:N \g_noteobserve
    Observe
  }
  \IfValueF{#1}{~}
}

\ExplSyntaxOff

\ExplSyntaxOn

\bool_new:N \g_hencetherefore

\NewDocumentCommand{\hence}{ o }{
  \IfValueT{#1}{
    \str_if_eq:noTF {hence} {#1} {
      \bool_gset_true:N \g_hencetherefore
    } {
      \str_if_eq:noTF {Hence} {#1} {
        \bool_gset_true:N \g_hencetherefore
      } {
        \bool_gset_false:N \g_hencetherefore
      }
    }
  }
  \bool_if:nTF { \g_hencetherefore } {
    \bool_gset_false:N \g_hencetherefore
    hence
  } {
    \bool_gset_true:N \g_hencetherefore
    therefore
  }
  \IfValueF{#1}{~}
}

\NewDocumentCommand{\Hence}{ o }{
  \IfValueT{#1}{
    \str_if_eq:noTF {hence} {#1} {
      \bool_gset_true:N \g_hencetherefore
    } {
      \str_if_eq:noTF {Hence} {#1} {
        \bool_gset_true:N \g_hencetherefore
      } {
        \bool_gset_false:N \g_hencetherefore
      }
    }
  }
  \bool_if:nTF { \g_hencetherefore } {
    \bool_gset_false:N \g_hencetherefore
    Hence,~we~obtain
  } {
    \bool_gset_true:N \g_hencetherefore
    Therefore,~we~obtain
  }
  \IfValueF{#1}{~}
}

\ExplSyntaxOff

\ExplSyntaxOn

\seq_const_from_clist:Nn \g_prove_mru {
  establish,
  demonstrate,
  prove,
  show,
  imply,
  ensure
}

\prop_new:N \l__verbs
\prop_put:Nnn \l__verbs {show} {shows}
\prop_put:Nnn \l__verbs {imply} {implies}
\prop_put:Nnn \l__verbs {demonstrate} {demonstrates}
\prop_put:Nnn \l__verbs {prove} {proves}
\prop_put:Nnn \l__verbs {establish} {establishes}
\prop_put:Nnn \l__verbs {ensure} {ensures}
\prop_put:Nnn \l__verbs {assure} {assures}

\tl_new:N \g_wordtmp
\seq_new:N \l_mytmps

\cs_generate_variant:Nn \str_if_in:nnTF { nVTF }
\cs_generate_variant:Nn \str_if_in:nnTF { xVTF }

\NewDocumentCommand{\prove}{ o }{
  \IfValueTF{#1}{
    \seq_clear:N \l_mytmps
    \seq_map_inline:Nn \g_prove_mru {
      \str_if_eq:nnTF {##1} {ensure} {
        \str_set:Nn \l_temps {n}
      } {
        \str_set:Nx \l_temps {\str_head_ignore_spaces:n {##1}}
      }
      \str_if_in:xVTF {#1} \l_temps {
        \seq_put_right:Nn \l_mytmps {##1}
      } { }
    }
    \seq_get_right:NN \l_mytmps \g_wordtmp
  } {
    \seq_get_right:NN \g_prove_mru \g_wordtmp
  }
  \tl_use:N \g_wordtmp
  \IfValueTF{#1}{}{~}
  \seq_gput_left:NV \g_prove_mru \g_wordtmp
  \seq_gremove_duplicates:N \g_prove_mru
}

\NewDocumentCommand{\proves}{ o }{
  \IfValueTF{#1}{
    \seq_clear:N \l_mytmps
    \seq_map_inline:Nn \g_prove_mru {
      \str_if_eq:nnTF {##1} {ensure} {
        \str_set:Nn \l_temps {n}
      } {
        \str_set:Nx \l_temps {\str_head_ignore_spaces:n {##1}}
      }
      \str_if_in:xVTF {#1} \l_temps {
        \seq_put_right:Nn \l_mytmps {##1}
      } { }
    }
    \seq_get_right:NN \l_mytmps \g_wordtmp
  } {
    \seq_get_right:NN \g_prove_mru \g_wordtmp
  }
  \str_set:NV \l_tmpa_str \g_wordtmp
  \prop_get:NVN \l__verbs \l_tmpa_str \l_tmpa_tl
  \tl_use:N \l_tmpa_tl
  \IfValueTF{#1}{}{~}
  \seq_gput_left:NV \g_prove_mru \g_wordtmp
  \seq_gremove_duplicates:N \g_prove_mru
}

\newcommand{\llabel}[1]{\savelabel{#1}\label{\loc.#1}\ignorespaces}

\clist_new:N \l_localreflist
\clist_new:N \l_reflist

\NewDocumentCommand{\lref} { m } {
  \clist_set:No \l_localreflist {#1}
  \clist_clear:N \l_reflist
  \clist_map_inline:Nn \l_localreflist { \clist_put_right:Nn \l_reflist {\loc.##1} }
  \cref{\l_reflist}
}

\NewDocumentCommand{\Lref} { m } {
  \clist_set:No \l_localreflist {#1}
  \clist_clear:N \l_reflist
  \clist_map_inline:Nn \l_localreflist { \clist_put_right:Nn \l_reflist {\loc.##1} }
  \Cref{\l_reflist}
}

\NewDocumentCommand{\itref}{ m m }{
  \clist_set:No \l_localreflist {#2}
  \clist_clear:N \l_reflist
  \clist_map_inline:Nn \l_localreflist { \clist_put_right:Nn \l_reflist {#1.##1} }
  \cref{\l_reflist}~in~\cref{#1}
}

\seq_new:N \l_enum_seq
\int_new:N \l_num_items

\bool_new:N \g_commaused_bool

\providecommand{\comma}{}

\cs_new:Nn \enum_it:nn {
  \int_case:nnF {\l_num_items - #1} {
    {0} {
      \renewcommand{\comma}{}
      #2\space
    }
    {1} {
      \bool_gset_false:N \g_commaused_bool
      \renewcommand{\comma}{,~\bool_gset_true:N \g_commaused_bool}
      #2
      \bool_if:NTF \g_commaused_bool {} {,~}
      and~
    }
  } {
    \bool_gset_false:N \g_commaused_bool
    \renewcommand{\comma}{,~\bool_gset_true:N \g_commaused_bool}
    #2
    \bool_if:NTF \g_commaused_bool {} {,~}
  }
}

\cs_new:Nn \enum_it_U:nn {
  \int_case:nnF {\l_num_items - #1} {
    {0} {
      \renewcommand{\comma}{}
      #2
      \space
    }
    {1} {
      \bool_gset_false:N \g_commaused_bool
      \renewcommand{\comma}{,~\bool_gset_true:N \g_commaused_bool}
      #2
      \bool_if:NTF \g_commaused_bool {} {,~}
      and~
    }
  } {
    \bool_gset_false:N \g_commaused_bool
    \renewcommand{\comma}{,~\bool_gset_true:N \g_commaused_bool}
    \int_compare:nTF {#1=1} {
      \text_titlecase_first:n {#2}
    } {
      #2
    }
    \bool_if:NTF \g_commaused_bool {} {,~}
  }
}

\cs_generate_variant:Nn \tl_if_eq:nnTF {onTF}

\cs_new:Nn \enum:nnnn {
  \seq_set_split:Nnn \l_enum_seq ; {#1}
  \seq_remove_all:Nn \l_enum_seq { }
  \seq_remove_all:Nn \l_enum_seq {#2}
  \seq_log:N \l_enum_seq
  \int_set:Nn \l_num_items {\seq_count:N \l_enum_seq}
  \int_log:N \l_num_items
  \int_case:nnF {\l_num_items} {
    { 0 } { 0 }
    { 1 } {
      \IfBooleanTF{#4} {
        \tl_set:Nn \l_text_case_exclude_arg_tl {\cref}
        \bool_set_false:N \l_text_titlecase_check_letter_bool
        \text_titlecase_first:n {\seq_use:Nn \l_enum_seq {}}
      } {
        \seq_use:Nn \l_enum_seq {}
      }
      \space
      \tl_if_eq:onTF{#3}{-}{}{
        \bool_if:NTF \l_plural_bool {
          \prove[#3]~
        } {
          \proves[#3]~
        }
      }
    }
    { 2 } {
      \IfBooleanTF{#4} {
        \tl_set:Nn \l_text_case_exclude_arg_tl {\cref}
        \bool_set_false:N \l_text_titlecase_check_letter_bool
        \text_titlecase_first:n {\seq_use:Nn \l_enum_seq {~and~}}
      } {
        \seq_use:Nn \l_enum_seq {~and~}
      }
      \space
      \tl_if_eq:onTF{#3}{-}{}{
        \prove[#3]~
      }
    }
  } {
    \IfBooleanTF{#4} {
      \tl_set:Nn \l_text_case_exclude_arg_tl {\cref}
      \seq_indexed_map_function:NN \l_enum_seq \enum_it_U:nn
    } {
      \seq_indexed_map_function:NN \l_enum_seq \enum_it:nn
    }
    \tl_if_eq:onTF{#3}{-}{}{
      \prove[#3]~
    }
  }
}

\cs_generate_variant:Nn \enum:nnnn {nxnn}
\cs_generate_variant:Nn \enum:nnnn {nxxn}

\NewDocumentCommand{\enum}{O{} m O{-} s}{
  \IfBooleanTF{#4}{
    \enum:nxnn {#2} {#1} {sindep} \BooleanFalse
  } {
    \enum:nxxn {#2} {#1} {#3} \BooleanFalse
  }
}

\NewDocumentCommand{\dott}{}{\ifnocf{.}\space}

\bool_new:N \g_arg_start_bool
\bool_gset_true:N \g_arg_start_bool

\NewDocumentCommand{\startnewargseq}{}{\bool_gset_true:N \g_arg_start_bool \tl_set:Nn \g_label_tl {}}

\cs_generate_variant:Nn \seq_if_in:NnTF {NxTF}
\cs_generate_variant:Nn \seq_remove_all:Nn {Nx}

\int_new:N \l_random_int

\cs_generate_variant:Nn \tl_if_head_eq_catcode:nNTF {oNTF}
\cs_generate_variant:Nn \tl_if_head_eq_catcode:nNTF {VNTF}

\cs_generate_variant:Nn \tl_if_head_eq_catcode:nNTF {eNTF}
\cs_generate_variant:Nn \tl_log:n {o}
\cs_generate_variant:Nn \tl_log:n {f}
\cs_generate_variant:Nn \tl_log:n {x}
\cs_generate_variant:Nn \tl_log:n {e}

\cs_generate_variant:Nn \tl_if_in:nnTF {onTF}
\cs_generate_variant:Nn \tl_if_in:NnTF {NeTF}

\cs_generate_variant:Nn \tl_if_head_eq_meaning:nNTF {VNTF}

\seq_const_from_clist:Nn \g_arg_mru_this {
  Ahpr,
  Tapr,
  Ctapr,
  H
}

\seq_const_from_clist:Nn \g_arg_mru_nothis {
  Ia,
  Nwc,
  N,
  Itns,
  Fm,
  Itnswc,
  Mo
}

\seq_new:N \l_arg_seq
\tl_new:N \l_cons_tl
\tl_new:N \l_dummy_tl

\bool_new:N \g_debug_bool
\bool_gset_false:N \g_debug_bool

\bool_new:N \l_insidearg_bool

\bool_new:N \g_firstargletter_bool

\sys_gset_rand_seed:n {0903}

\bool_new:N \l_plural_bool
\tl_new:N \l_arg_verbs_tl

\NewDocumentCommand{\argument}{mom}{
\color{black}
  \bool_set_false:N \l_plural_bool
  \tl_set:Nn \l_arg_verbs_tl {sindep}
  \keys_define:nn { benno/argument } {
    plural .value_forbidden:n = true,
    plural .code:n = {\bool_set_true:N \l_plural_bool},
    verbs .value_required:n = false,
    verbs .tl_set:N = \l_arg_verbs_tl,
  }
  \IfValueT{#2}{
    \keys_set:nn { benno/argument } {#2}
  }
  \bool_log:N \l_plural_bool
  \bool_gset_true:N \l_insidearg_bool
  \seq_set_split:Nnn \l_arg_seq ; {#1}
  \seq_remove_all:Nn \l_arg_seq { }
  \seq_log:N \l_arg_seq
  \tl_set:Nn \l_cons_tl {#3}
  \tl_trim_spaces:N \l_cons_tl
  \seq_if_in:NxTF \l_arg_seq {\lref{\g_label_tl}} {
    \seq_remove_all:Nx \l_arg_seq {\lref{\g_label_tl}}
    \seq_get_left:NNTF \l_arg_seq \l_dummy_tl {
      \tl_trim_spaces:N \l_dummy_tl
      \bool_gset_false:N \g_firstargletter_bool
      \tl_if_head_eq_catcode:VNTF \l_dummy_tl a {
        \bool_gset_true:N \g_firstargletter_bool
      } {
        \tl_if_head_eq_meaning:VNTF \l_dummy_tl {\cref} {
          \tl_set:Nx \l_tmpa_tl {\tl_tail:N \l_dummy_tl}
          \tl_set:Nx \l_tmpb_tl {\tl_head:N \l_tmpa_tl}
          \bool_gset_true:N \g_firstargletter_bool
          \tl_if_in:NeTF \l_tmpb_tl {lem\c_colon_str} {} {
            \tl_if_in:NeTF \l_tmpb_tl {thm\c_colon_str} {} {
              \tl_if_in:NeTF \l_tmpb_tl {prop\c_colon_str} {} {
                \tl_if_in:NeTF \l_tmpb_tl {cor\c_colon_str} {} {
                  \bool_gset_false:N \g_firstargletter_bool
                }
              }
            }
          }
        } {
        }
      }
      \bool_if:NTF \g_firstargletter_bool {
        \seq_set_eq:NN \l_tmpa_seq \g_arg_mru_this
        \seq_remove_all:Nn \l_tmpa_seq {H}
        \seq_get_right:NN \l_tmpa_seq \l_tmpa_tl
        \int_case:nnF {\seq_count:N \l_arg_seq} {
          {1} {
            \str_case:VnF {\l_tmpa_tl} {
              {Ahpr} {
                \bool_if:NT \g_debug_bool {C1.1}
                \seq_gput_left:Nn \g_arg_mru_this {Ahpr}
                \seq_gremove_duplicates:N \g_arg_mru_this
                \enum:nxnn {#1} {\lref{\g_label_tl}} {-} {\BooleanTrue}
                \hence~
                \bool_if:NTF \l_plural_bool {
                  \prove[\l_arg_verbs_tl]~\ignorespaces #3
                } {
                  \proves[\l_arg_verbs_tl]~\ignorespaces #3
                }
              }
              {Tapr} {
                \bool_if:NT \g_debug_bool {C1.2}
                \seq_gput_left:Nn \g_arg_mru_this {Tapr}
                \seq_gremove_duplicates:N \g_arg_mru_this
                \enum[\lref{\g_label_tl}]{
                  This;
                  #1
                }[\l_arg_verbs_tl]\ignorespaces #3
              }
              {Ctapr} {
                \bool_if:NT \g_debug_bool {C1.3}
                \seq_gput_left:Nn \g_arg_mru_this {Ctapr}
                \seq_gremove_duplicates:N \g_arg_mru_this
                Combining~
                \enum[\lref{\g_label_tl}]{
                  this;
                  #1
                } \proves[\l_arg_verbs_tl]~\ignorespaces #3
              }
            } {}
          }
        } {
          \str_case:VnF {\l_tmpa_tl} {
             {Ahpr} {
              \bool_if:NT \g_debug_bool {C2.1}
              \seq_gput_left:Nn \g_arg_mru_this {Ahpr}
              \seq_gremove_duplicates:N \g_arg_mru_this
              \enum:nxnn {#1} {\lref{\g_label_tl}} {-} {\BooleanTrue}
              \hence~
              \prove[\l_arg_verbs_tl]~\ignorespaces #3
            }
            {Tapr} {
              \bool_if:NT \g_debug_bool {C2.2}
              \seq_gput_left:Nn \g_arg_mru_this {Tapr}
              \seq_gremove_duplicates:N \g_arg_mru_this
              \enum[\lref{\g_label_tl}]{
                This;
                #1
              }[\l_arg_verbs_tl]\ignorespaces #3
            }
            {Ctapr} {
              \int_case:nn {\int_rand:nn {0} {1}} {
                {0} {
                  \bool_if:NT \g_debug_bool {C2.3}
                  \seq_gput_left:Nn \g_arg_mru_this {Ctapr}
                  \seq_gremove_duplicates:N \g_arg_mru_this
                  Combining~
                  \enum[\lref{\g_label_tl}]{
                    this;
                    #1
                  } \proves[\l_arg_verbs_tl]~\ignorespaces #3
                }
                {1} {
                  \bool_if:NT \g_debug_bool {C2.4}
                  \seq_gput_left:Nn \g_arg_mru_this {Ctapr}
                  \seq_gremove_duplicates:N \g_arg_mru_this
                  Combining~
                  \enum:nxnn {#1} {\lref{\g_label_tl}} {-} {\BooleanFalse}
                  \hence~
                  \proves[\l_arg_verbs_tl]~\ignorespaces #3
                }
              }
            }
          } {}
        }
      } {
        \seq_set_eq:NN \l_tmpa_seq \g_arg_mru_this
        \seq_remove_all:Nn \l_tmpa_seq {H}
        \seq_remove_all:Nn \l_tmpa_seq {Ahpr}
        \seq_get_right:NN \l_tmpa_seq \l_tmpa_tl
        \int_case:nnF {\seq_count:N \l_arg_seq} {
          {1} {
            \str_case:VnF {\l_tmpa_tl} {
              {Tapr} {
                \bool_if:NT \g_debug_bool {C3.1}
                \seq_gput_left:Nn \g_arg_mru_this {Tapr}
                \seq_gremove_duplicates:N \g_arg_mru_this
                \enum[\lref{\g_label_tl}]{
                  This;
                  #1
                }[\l_arg_verbs_tl]\ignorespaces #3
              }
              {Ctapr} {
                \bool_if:NT \g_debug_bool {C3.2}
                \seq_gput_left:Nn \g_arg_mru_this {Ctapr}
                \seq_gremove_duplicates:N \g_arg_mru_this
                Combining~
                \enum[\lref{\g_label_tl}]{
                  this;
                  #1
                } \proves[\l_arg_verbs_tl]~\ignorespaces #3
              }
            } {}
          }
        } {
          \str_case:VnF {\l_tmpa_tl} {
            {Tapr} {
              \bool_if:NT \g_debug_bool {C4.1}
              \seq_gput_left:Nn \g_arg_mru_this {Tapr}
              \seq_gremove_duplicates:N \g_arg_mru_this
              \enum[\lref{\g_label_tl}]{
                This;
                #1
              }[\l_arg_verbs_tl]\ignorespaces #3		
            }
            {Ctapr} {
              \int_case:nn {\int_rand:nn {0} {1}} {
                {0} {
                  \bool_if:NT \g_debug_bool {C4.2}
                  \seq_gput_left:Nn \g_arg_mru_this {Ctapr}
                  \seq_gremove_duplicates:N \g_arg_mru_this
                  Combining~
                  \enum[\lref{\g_label_tl}]{
                    this;
                    #1
                  } \proves[\l_arg_verbs_tl]~\ignorespaces #3		
                }
                {1} {
                  \bool_if:NT \g_debug_bool {C4.3}
                  \seq_gput_left:Nn \g_arg_mru_this {Ctapr}
                  \seq_gremove_duplicates:N \g_arg_mru_this
                  Combining~
                  \enum:nxnn {#1} {\lref{\g_label_tl}} {-} {\BooleanFalse}
                  \hence~
                  \proves[\l_arg_verbs_tl]~\ignorespaces #3    
                }
              }
            }
          } {}
        }
      }
    } {
      \tl_if_head_eq_catcode:oNTF \l_cons_tl a {
        \seq_set_eq:NN \l_tmpa_seq \g_arg_mru_this
        \seq_remove_all:Nn \l_tmpa_seq {Ctapr}
        \seq_remove_all:Nn \l_tmpa_seq {Ahpr}
        \seq_get_right:NN \l_tmpa_seq \l_tmpa_tl
        \str_case:VnF {\l_tmpa_tl} {
          {H} {
            \bool_if:NT \g_debug_bool {C5.1}
            \seq_gput_left:Nn \g_arg_mru_this {H}
            \seq_gremove_duplicates:N \g_arg_mru_this
            Hence,~we~obtain~\ignorespaces #3
          }
          {Tapr} {
            \bool_if:NT \g_debug_bool {C5.2}
            \seq_gput_left:Nn \g_arg_mru_this {Tapr}
            \seq_gremove_duplicates:N \g_arg_mru_this
            This~\proves[\l_arg_verbs_tl]~\ignorespaces #3
          }
        } {}
      } {
        \bool_if:NT \g_debug_bool {C6.1}
        \seq_gput_left:Nn \g_arg_mru_this {Tapr}
        \seq_gremove_duplicates:N \g_arg_mru_this
        This~\proves[\l_arg_verbs_tl]~\ignorespaces #3
      }
    } 
  } {
    \int_compare:nNnTF {\seq_count:N \l_arg_seq} = {0} {
      \bool_if:NTF \g_arg_start_bool {
        \bool_if:NT \g_debug_bool {C7.1}
        \Nobs\unskip
        #3
      } {
        \bool_if:NT \g_debug_bool {C7.2}
        \Moreover~
        #3
      }
    } {
      \bool_if:NTF \g_arg_start_bool {
        \bool_if:NT \g_debug_bool {C8.1}
        \tl_log:N \l_arg_verbs_tl
        \Nobs~that~
        \enum{
          #1
        }[\l_arg_verbs_tl]\ignorespaces #3
      } {
        \int_compare:nNnTF {\seq_count:N \l_arg_seq} = {1} {
          \seq_set_eq:NN \l_tmpa_seq \g_arg_mru_nothis
          \seq_remove_all:Nn \l_tmpa_seq {Nwc}
          \seq_remove_all:Nn \l_tmpa_seq {Itnswc}
          \seq_get_right:NN \l_tmpa_seq \l_tmpa_tl
        } {
          \seq_get_right:NN \g_arg_mru_nothis \l_tmpa_tl
        }
        \str_case:VnF {\l_tmpa_tl} {
          {Mo} {
            \bool_if:NT \g_debug_bool {C9.1}
            \seq_gput_left:Nn \g_arg_mru_nothis {Mo}
            \seq_gremove_duplicates:N \g_arg_mru_nothis
            Moreover,~\nobs~that~
            \enum{
              #1
            }[\l_arg_verbs_tl]\ignorespaces #3		
          }
          {Fm} {
            \bool_if:NT \g_debug_bool {C9.2}
            \seq_gput_left:Nn \g_arg_mru_nothis {Fm}
            \seq_gremove_duplicates:N \g_arg_mru_nothis
            Furthermore,~\nobs~that~
            \enum{
              #1
            }[\l_arg_verbs_tl]\ignorespaces #3		
          }
          {Ia} {
            \bool_if:NT \g_debug_bool {C9.3}
            \seq_gput_left:Nn \g_arg_mru_nothis {Ia}
            \seq_gremove_duplicates:N \g_arg_mru_nothis
            In~addition,~\nobs~that~
            \enum{
              #1
            }[\l_arg_verbs_tl]\ignorespaces #3		
          }
          {N} {
            \bool_if:NT \g_debug_bool {C9.4}
            \seq_gput_left:Nn \g_arg_mru_nothis {N}
            \seq_gremove_duplicates:N \g_arg_mru_nothis
            Next,~\nobs~that~
            \enum{
              #1
            }[\l_arg_verbs_tl]\ignorespaces #3		
          }
          {Itns} {
            \bool_if:NT \g_debug_bool {C9.5}
            \seq_gput_left:Nn \g_arg_mru_nothis {Itnswc}
            \seq_gput_left:Nn \g_arg_mru_nothis {Itns}
            \seq_gremove_duplicates:N \g_arg_mru_nothis
            In~the~next~step~we~\nobs~that~
            \enum{
              #1
            }[\l_arg_verbs_tl]\ignorespaces #3		
          }
          {Nwc} {
            \bool_if:NT \g_debug_bool {C9.6}
            \seq_gput_left:Nn \g_arg_mru_nothis {Nwc}
            \seq_gremove_duplicates:N \g_arg_mru_nothis
            Next~we~combine~
            \enum{
              #1
            }to~obtain~\ignorespaces #3
          }
          {Itnswc} {
            \bool_if:NT \g_debug_bool {C9.7}
            \seq_gput_left:Nn \g_arg_mru_nothis {Itns}
            \seq_gput_left:Nn \g_arg_mru_nothis {Itnswc}
            \seq_gremove_duplicates:N \g_arg_mru_nothis
            In~the~next~step~we~combine~
            \enum{
              #1
            }to~obtain~\ignorespaces #3
          }
        } {}
      }
    }
  }
  \bool_gset_false:N \g_arg_start_bool
  \bool_gset_false:N \l_insidearg_bool
  \cfload[.]
  \color{black}
}

\tl_new:N \g_label_tl
\tl_gset:Nn \g_label_tl { }

\NewDocumentCommand{\savelabel}{m}{
  \bool_if:NTF \l_insidearg_bool {
    \tl_gset:Nn \g_label_tl {#1}
  } {
    \tl_gset:Nn \g_label_tl { }
  }
}

\ExplSyntaxOff

\ExplSyntaxOn

\NewDocumentEnvironment {athm} {m m o} {
\str_if_eq:noTF {example} {#1} {
  \bool_gset_true:N \g_example_bool
} {
  \bool_gset_false:N \g_example_bool
}
\cfclear
\IfNoValueTF{#3}{
\begin{#1}\label{#2}\global\def\loc{#2}
}{
\begin{#1}[#3]\label{#2}\global\def\loc{#2}
}
}{
\end{#1}
}

\NewDocumentEnvironment {adef} {m} {
\begin{definition}\label{#1}\global\def\loc{#1}
}{
\end{definition}
}

\NewDocumentEnvironment{aproof} {} {
\bool_if:NTF \g_example_bool {
  \bool_gset_true:N \g_arg_start_bool
  \begin{proof}[Proof~for~\cref{\loc}]
} {
  \bool_gset_true:N \g_arg_start_bool
  \begin{proof}[Proof~of~\cref{\loc}]
}
\bool_gset_false:N \g_finishproof_bool
}{
\bool_if:NTF \g_finishproof_bool {}
{\finishproofthus}
\end{proof}
}

\NewDocumentCommand{\finishproofthus} {} {
  \bool_gset_true:N \g_finishproof_bool 
  \bool_if:NTF \g_example_bool {
    The~proof~for~\cref{\loc}~is~thus~complete.
  } {
    The~proof~of~\cref{\loc}~is~thus~complete.
  }
}
\NewDocumentCommand{\finishproofthis} {} {
  \bool_gset_true:N \g_finishproof_bool 
  \bool_if:NTF \g_example_bool {
    This~completes~the~proof~for~\cref{\loc}.
  } {
    This~completes~the~proof~of~\cref{\loc}.
  }
}

\ExplSyntaxOff

\NewDocumentEnvironment{cproof}{m}
{\begin{proof}[Proof of \cref{#1}]}%
{\noindent The proof of \cref{#1} is thus complete.
\end{proof}}

\NewDocumentEnvironment{cproof2}{m}
{\begin{proof}[Proof of \cref{#1}]}%
{\noindent This completes the proof of \cref{#1}.
\end{proof}}

\hypersetup{
    colorlinks,
    linkcolor={blue!80!black},
    citecolor={green},
    urlcolor={blue!80!black}
}

\makeatletter
\ExplSyntaxOn
\seq_new:N \g_abbrs
\prop_new:N \g_abbr_counts
\tl_new:N \l_abbr_count_tl

\ExplSyntaxOn

\bool_new:N \g_forexample

\NewDocumentCommand{\eg}{ o }{
	\IfValueT{#1}{
		\str_if_eq:noTF {fe} {#1} {
			\bool_gset_true:N \g_forexample
		} {\bool_gset_false:N \g_forexample}
	}
	\bool_if:nTF { \g_forexample } {
		\bool_gset_false:N \g_forexample
		for~example
	}{
		\bool_gset_true:N \g_forexample
		for~instance
	}
}

\NewDocumentCommand{\abbr}{m m O{#1} m m O{#4} m}{
	\expandafter\newcommand\csname#3\endcsname[1][]{
		\seq_if_in:NnTF \g_abbrs {#1} {
			\prop_get:NnN \g_abbr_counts {#1} \l_abbr_count_tl
			\prop_gput:Nnx \g_abbr_counts {#1} {\int_eval:n {\l_abbr_count_tl + 1}}
			\hyperref[#1]{#7}
		} {
			\seq_gput_left:Nn \g_abbrs {#1}
			\prop_gput:Nnn \g_abbr_counts {#1} {1}
			\expandafter\gdef\csname#1@def\endcsname{#2}
			\phantomsection\label{#1}
			\str_if_eq:nnTF{##1}{}{\emph{#2}}{##1}~(\hyperref[#1]{#7})
		}
	}
	\expandafter\newcommand\csname#6\endcsname[1][]{
		\seq_if_in:NnTF \g_abbrs {#1} {
			\prop_get:NnN \g_abbr_counts {#1} \l_abbr_count_tl
			\prop_gput:Nnx \g_abbr_counts {#1} {\int_eval:n {\l_abbr_count_tl + 1}}
			\hyperref[#1]{#4}
		} {
			\expandafter\gdef\csname#1@def\endcsname{#5}
			\seq_gput_left:Nn \g_abbrs {#1}
			\prop_gput:Nnn \g_abbr_counts {#1} {1}
			\phantomsection\label{#1}
			\str_if_eq:nnTF{##1}{}{\emph{#5}}{##1}~(\hyperref[#1]{#4})
		}
	}
}

\ExplSyntaxOff
\makeatother
\abbr{GD}{gradient descent}{GDs}{gradient descents}{GD}
\abbr{ANN}{artificial neural network}{ANNs}{artificial neural networks}{ANN}
\abbr{DNN}{deep neural network}{DNNs}{deep neural networks}{DNN}
\abbr{SGD}{Stochastic gradient descent}{SGDs}{Stochastic gradient descent}{SGD}
\abbr{iid}{independent and identically distributed}{i.i.d}{independent and identically distributed}{i.i.d.}
\abbr{SOP}{stochastic optimization problem}{SOPs}{stochastic optimization problems}{SOP}
\abbr{DL}{Deep learning}{DLs}{Deep learning}{DL}
\abbr{Adam}{adaptive moment estimation \SGD}{Adams}{adaptive moment estimation \SGD}{Adam}
\abbr{AI}{artificial intelligence}{AIs}{artificial intelligence}{AI}
\abbr{LLM}{large language model}{LLMs}{large language models}{LLM}
\abbr{PDE}{partial differential equation}{PDEs}{partial differential equations}{PDE}
\abbr{as}{almost surely}{ass}{almost surely}{a.s.}
\abbr{pas}{$\P$-almost surely}{pass}{almost surely}{$\P$-a.s.}
\abbr{ReLUs}{rectified linear unit}{ReLU}{rectified linear unit}
\makeatother

\title{
Uniform a priori bounds and error analysis for the\\Adam stochastic gradient descent optimization method
}
\author{Steffen Dereich$^{1}$, Thang Do$^{2,3}$, and Arnulf Jentzen$^{4,5}$
	\bigskip
	\\
	\small{$^1$ Institute for Mathematical Stochastics, Faculty of Mathematics and Computer Science,}\vspace{-0.1cm}\\
\small{University of M\"unster, Germany, e-mail: \texttt{steffen.dereich@uni-muenster.de}}
\smallskip
\\
    	\small{$^2$ School of Data Science, The Chinese University of Hong Kong, Shenzhen}
	\vspace{-0.1cm}\\
	\small{ (CUHK-Shenzhen), China, e-mail: \texttt{minhthangdo@link.cuhk.edu.cn}}
 \smallskip
	\\
    \small{$^3$ Department of Probability and Statistic, Institute of Mathematics,}
	\vspace{-0.1cm}\\
	\small{Vietnam Academy of Science and Technology, Vietnam, e-mail: \texttt{dmthang@math.ac.vn}}
	\smallskip
	\\
	\small{$^4$ School of Data Science and School of Artificial Intelligence, The Chinese University}
	\vspace{-0.1cm}\\
	\small{of Hong Kong, Shenzhen (CUHK-Shenzhen), China, e-mail: \texttt{ajentzen@cuhk.edu.cn}}
	\smallskip
	\\
 \small{$^5$ Applied Mathematics: Institute for Analysis and Numerics, Faculty of Mathematics and}
	\vspace{-0.1cm}\\
	\small{Computer Science, University of M{\"u}nster, Germany, e-mail: \texttt{ajentzen@uni-muenster.de}}
	\smallskip
	\\
}

\date{\today}
\allowdisplaybreaks 
\begin{document}
\maketitle
\begin{abstract}
    The \emph{adaptive moment estimation (Adam)} optimizer proposed by Kingma \& Ba (2014) is presumably the most popular stochastic gradient descent (SGD) optimization method for the training of \emph{deep neural networks (DNNs)} in \emph{artificial intelligence (AI)} systems. Despite its groundbreaking success in the training of AI systems, it still remains an open research problem to provide a complete error analysis of Adam, not only for optimizing DNNs but even when applied to strongly convex \emph{stochastic optimization problems (SOPs)}. Previous error analysis results for strongly convex SOPs in the literature provide conditional convergence analyses that rely on the assumption that Adam does not diverge to infinity but remains uniformly bounded. It is the key contribution of this work to establish uniform a priori bounds for Adam and, thereby, to provide -- for the first time -- an unconditional error analysis for Adam for a large class of strongly convex SOPs.
\end{abstract}
\tableofcontents
\section{Introduction}\label{sec: introduction}

\SGD\ optimization schemes are the methods of choice to train \DNNs\ in \AI\ systems. Beyond the plain vanilla standard \SGD\ method \cite{MR42668}, there is a special interest in sophisticated \SGD\ methods involving adaptivity and acceleration techniques \cite{ArBePhi2024,Ruder2016AnOO}. The \Adam\ optimizer proposed in Kingma \& Ba \cite{KingmaBa2024_Adam} in 2014 -- a work that has gathered more than 240\,000 citations on {\sc Google Scholar} -- is presumably the most popular of such sophisticated \SGD\ optimization methods.

Despite the groundbreaking success of the \Adam\ optimization method in the training of \AI\ systems, it remains an open problem of research to provide a complete error and convergence analysis for \Adam, even in the situation of strongly convex \SOP. In particular, in previous works in the literature \cite{DeDoArPhi2025,DereichAdamconvergence2024} error analyses for \Adam\ are established under the assumption that the \Adam\ optimization process does not diverge to infinity but stays bounded. However, even in the situation of strongly convex \SOPs, it remained unclear whether the \Adam\ method satisfies this assumption.

In this work we partially solve this problem by establishing in \cref{cor: priori bound stochastic Adam 1 non explosion} in \cref{subsec: deterministic bound} below explicit \emph{pathwise uniform a priori bounds} for the \Adam\ optimization process applied to strongly convex \SOPs\ with the gradients of the objective functions satisfying suitable (local) Lipschitz continuity properties. Combining the pathwise uniform a priori bounds in \cref{cor: priori bound stochastic Adam 1 non explosion} in this work with existing conditional error analyses for \Adam\ that rely on a priori bounds for \Adam\ \cite{DeDoArPhi2025,DereichAdamconvergence2024} allows us then to establish in \cref{cor: convegence of Adam3} in \cref{subsec: convergence without L} an unconditional error analysis for \Adam\ that is also applicable to several concrete examples of \SOPs\ (see \cref{cor: priori bound stochastic Adam 2 explosion} and \cref{cor: priori bound stochastic Adam 2 non explosion} in this introductory section below). To illustrate the contribution of this work within this introductory section, we now present in the following \cref{main theorem} a special case of our general convergence analysis from \cref{cor: convegence of Adam3}. 
\subsection{Main result: Unconditional error analysis for Adam}

In \cref{main theorem} the triple $( \Omega, \cF, \P )$ is the underlying probability space for the considered \SOP, the natural number $d \in \N = \{ 1, 2, 3, \dots \}$ represents the dimensionality of the considered \SOP\ (the number of real parameters that are optimized), the natural number $\dimX \in \N$ represents the dimensionality of the data in the considered \SOP, the number $p \in \N$ is a sufficiently large constant, and the bounded \iid\ random variables $X_{ n, m } \colon \Omega \to [ - p, p ]^{ \dimX }$ represent the (input and output) data in the considered \SOP. We now present \cref{main theorem} with all mathematical details in a completely self-contained fashion and, thereafter, we provide more explanatory comments regarding the mathematical objects and the contribution of \cref{main theorem}.

\begin{savenotes}
\begin{samepage}
\begin{tcolorbox}[colback=white!95!gray,
                  colframe=black,
                  boxrule=0.5pt,
                  sharp corners,
                  enhanced,
                  breakable,
                 ]
\begin{athm}{theorem}{main theorem}
     Let $( \Omega, \cF,\P )$ be a probability space, let $d, \dimX \in \N$, $\varepsilon,p \in (0,\infty)$, $q,r\in (0,\nicefrac{1}{2})$, $\xi\in \R^d$, let $(\gamma_n)_{n\in \N}\subseteq (0,\infty)$ be non-increasing, let $U\subseteq \R^\dimX$ be compact and convex, let $\smalll=(\smalll(\theta,x))_{(\theta,x)\in \R^\fd\times \R^\dimX}\allowbreak\in C^2(\R^\fd\times \R^\dimX,\R)$ satisfy\footnote{Note that for all $n \in \N$, $v = ( v_1, \dots, v_n ) \in \R^n$ it holds that $\| v \| = (\sum_{ i = 1 }^n | v_i |^2)^{ 1 / 2 }$ (standard norm).} for all $\theta=(\theta_1,\dots,\theta_\fd)\in \R^\fd$, $x\in U$ that 
 \begin{equation}\llabel{eq1: main theorem}
\textcolor{magenta}{\textstyle\sum_{i=1}^d\bigl(\|(\nabla_{\theta_i}\nabla_\theta\smalll)(\theta,x)\|+(1+\|\theta\|)^{-r}\|(\nabla_{\theta_i}\nabla_x\smalll)(\theta,x)\|\bigr)\leq p},
 \end{equation}
 assume
       $\limsup_{n\to\infty}( ( \gamma_n )^{ - 2 } ( \gamma_n- \gamma_{ n + 1 } ) +\sum_{ m = n }^{ \infty } ( \gamma_m)^p ) =0$,
 let $X_{ n,m } \colon \Omega \to U$, $(n,m ) \in \N^2$, be \iid\ random variables, for every $M\in \N$, $\para=(\para_1,\para_2)\in (0,1)^2$ let 
    $
\mandV{k}{M}{\para}{n}=(\mandVcom{k}{M}{\para}{n}{1},\dots,\mandVcom{k}{M}{\para}{n}{\fd})\colon\allowbreak \Omega\to\R^\fd$, $(k,n)\in (\N_0)^2$, satisfy for all $k\in \{1,2\}$, $n\in \N$, $i\in \{1,2,\dots,\fd\}$ that
    \begin{gather}
      \textcolor{magenta}{ \mandV{k}{M}{\para}{0}=0,\quad \mandVcom{k}{M}{\para}{n}{i}= \beta_k \mandVcom{k}{M}{\para}{n-1}{i}+(1-\beta_k)\bigl[\textstyle \frac 1M \sum_{m=1}^M(\nabla_{\theta_i} \smalll)(\mandV{0}{M}{\beta}{n-1},X_{n,m})\bigr]^k}, \label{eq2: main theorem}\\
        \textcolor{magenta}{ \mandVcom{0}{M}{\para}{n}{i}=\mandVcom{0}{M}{\para}{n-1}{i} -\gamma_n [1-(\para_1)^n]^{-1}\bigl[\varepsilon+\bigl[[1-(\para_2)^n]^{-1}\mandVcom{2}{M}{\para}{n}{i}\bigr]^{\nicefrac{1}{2}}\bigr]^{-1}\mandVcom{1}{M}{\para}{n}{i}}\label{eq3: main theorem},
     \end{gather}
     and  $   \textcolor{magenta}{\mandV{0}{M}{\para}{0}=\xi}$, assume for all $x \in U$ that $\R^d \ni \theta \mapsto \smalll( \theta, x ) - q \| \theta \|^2 \in \R$ is convex, and let $\globalmin \in \R^\fd$ satisfy $
           \E[\smalll(\globalmin,X_{1,1})]=\inf_{\theta\in \R^\fd}\E[\smalll(\theta,X_{1,1})]$.
       Then there exists $(\scrc_{\para})_{\para\in \R^2}\subseteq \R$ such that for all $M,n \in \N$, $\para = ( \para_1, \para_2 ) \in   [ q, 1 )^2$ with $( \beta_1 )^2 + q \leq \beta_2 $ it holds that
\begin{equation} \label{eq4: main theorem}
   \textcolor{magenta}{  \bigl( \E\bigl[ \| \mandV{0}{M}{\para}{n} - \globalmin\|^p \bigr] \bigr)^{ 1/p } \leq  \scrc_0 M^{ - 1 } ( 1 - \beta_2 \mathbbm 1_{ [\scrc_0,\infty)} ( M ) ) + \scrc_{ \beta }  \sqrt{ \gamma_{ n  } }}.
\end{equation}
\end{athm}
\end{tcolorbox}
\end{samepage}
\end{savenotes}
\cref{main theorem} is an immediate consequence from \cref{cor: convegence of Adam3} in \cref{sec: convergence of Adam} below. In \cref{eq2: main theorem} and \cref{eq3: main theorem} in \cref{main theorem} the \Adam\ optimization process is specified. In \cref{main theorem} we assume there exists an arbitrarily small strictly positive real number $q \in (0,\nicefrac{1}{2})$ such that for every data point $x \in \R^{ \dimX }$ we have that the loss function $\smalll( \cdot, x )$ is $q$-strongly convex (cf., \eg, \cite[Definition 5.7.18]{ArBePhi2024}). This ensures that the objective function (the function we intend to minimize) $\R^d \ni \theta \mapsto \E[ \smalll( \theta, X_{ 1, 1 } ) ]\in \R$ in \cref{main theorem} is $ q $-strongly convex. This, in turn, implies that the objective function of the \SOP\ in \cref{main theorem} admits a unique global minimizer $\vartheta$ (see, \eg, \cite[Corollary 5.7.22]{ArBePhi2024}), which is denoted by $\vartheta \in  \R^d$ in \cref{main theorem}. 

In \cref{eq4: main theorem} in \cref{main theorem} we provide an error estimate for the strong $L^p$-distance between the \Adam\ optimization process $\Theta^{ 0, M, \beta }_n$ after $n \in \N$ steps and the minimizer $\vartheta \in \R^d$ of the \SOP. Given a prescribed approximation accuracy $\epsilon > 0$ we can first choose the second moment parameter $\beta_2 \in (0,1)$ sufficiently close to $1$ and/or the size of the mini-batches $M \in \N$ sufficiently large to ensure that the first summand on the right hand side of \cref{main theorem} is smaller than $\nicefrac{ \epsilon }{ 2 }$ and, thereafter, we can choose the number of \Adam\ steps $n \in \N$ sufficiently large so that the second summand on the right hand side of \cref{eq4: main theorem} is also smaller than $\nicefrac{ \epsilon }{ 2 }$ to thereby assure that the overall error between the \Adam\ method and the minimizer is smaller than $\varepsilon$. 

In {\sc Tensorflow} \cite{tensorflow2025adam} and {\sc Pytorch} \cite{pytorch2025adam} the default value for the second moment parameter $\beta_2 \in (0,1)$ of \Adam\ \cite[Algorithm 1]{KingmaBa2024_Adam} satisfies $\beta_2 = 1 - \frac{ 1 }{ 1000 } = 0.999$ and is thus very close to $1$. 
In the training of \LLMs\ \cite[Table 1]{DeDoArPhi2025} a quite popular choice for the second moment parameter $\beta_2 \in (0,1)$ of \Adam\ is the choice $\beta_2 = 0.95$, in which case the error term $1 - \beta_2 = 0.05 = \nicefrac{ 1 }{ 20 }$ (see \cref{eq4: main theorem}) is small but not so small anymore.

It should also be pointed out that the first summand $\scrc_0 M^{ - 1 } ( 1 - \beta_2 \mathbbm 1_{ [\scrc_0,\infty) }( M ) )$ on the right hand side of \cref{eq4: main theorem} can, in general, not be omitted as the conclusion of \cref{eq4: main theorem} without the first summand on the right hand side of \cref{eq4: main theorem} does, in general, not hold anymore: We refer to the characterization in the \emph{\Adam\ symmetry theorem} in \cite[Theorem 1.1]{DeDoArPhi2025} for details.

\subsection{Convergence rates for concrete example stochastic optimization problems}

A key feature of \cref{main theorem} is that it does not only contain abstract assumptions that can not be verified in examples but instead that there are several concrete examples of \SOPs\ to which \cref{main theorem} is applicable. For instance, \cref{main theorem} is applicable to this class of quadratic \SOPs\ with (or without) regularization
\begin{equation}
\label{**}
  \textstyle \operatorname{argmin}_{ \theta \in \R^d } 
  \E\bigl[ 
    \| A_0 \theta - f_0( X_{ 1, 1 } ) \|^2 
    + 
    \sum_{ i = 1 }^\numofsum  ( \| A_i \theta \|^2 + \mu_i )^{ r_i } f_i( X_{ 1, 1 } )
  \bigr] 
\end{equation}
where $\numofsum, \dimofsum \in \N$ are natural numbers,  
where $A_0 \in \R^{ d \times d }$ is an invertible matrix, 
where $A_1, A_2, \dots, A_d \in \R^{ \dimofsum \times d }$ are matrices, 
where $\mu_1, \mu_2, \dots, \mu_\numofsum \in (0,\infty)$, $r_1, r_2, \dots, r_\numofsum \in [ \nicefrac{1}{2}, \nicefrac{3}{4} )$ are reals, 
and where $f_0 \in C( \R^\dimX, \R^d )$, $f_1\in C( \R^\dimX, [0,\infty) )$, $\dots$, $f_\numofsum\in C( \R^\dimX, [0,\infty) )$ are continuous functions. For the detailed application of \cref{main theorem} to the \SOPs\ in \cref{**} above we refer to \cref{cor: convegence of Adam4} in \cref{subsec: assume without L smooth} below.

To the best of our knowledge, even in the special case in \cref{**} in which it holds for all $x \in \R^\dimX$ that $\sum_{ i = 1 }^v \| f_i( x ) \| = 0$ in \cref{**} (quadratic \SOP\ without regularization), we have that \cref{main theorem} is the first result in the scientific literature that provides an unconditional (without assuming boundedness of \Adam) upper bound for the strong optimization error $\E\bigl[ \| \Theta^{ 0, M, \beta }_n - \vartheta \| \bigr]$ of \Adam\ that converges to zero as $n$ increases as well as $M$ increases and $1-\beta$ decreases, respectively (see also the brief literature review in the next subsection).

\subsection{Literature review}
There is also a large number of further works in the scientific literature that study \Adam\ 
and related \SGD\ optimization methods analytically. 
For works that study upper bounds for \Adam\ applied to \SOPs\ in terms of the strong/weak optimization error, 
the regret, or averages or minima (along the gradient steps $n \in \{0, 1, \dots , N \}$ of \Adam) of the norm of the gradient of the objective function we mention, \eg, \cite{Barakattopological2021,DereichAdamconvergence2024,Alexconvergence2025,ar2112.03459,ar2104.14840,ar2304.13972,ReddiKale2019,ar1808.03408,ar2403.15146,ar2208.09900,ar2404.01436,ZhangChen2022,ZouShen2019}. 
We also mention the work \cite[Lemma 2.9]{DeRoArAd2025} in which a priori bounds for \Adam\ applied to the stochastic \SOP\
\begin{equation}
\label{***}
\R^d \ni \E\bigl[ \| \theta - X_{ 1, 1 } \|^2\bigr] \in \R
\end{equation}
are established in the case $d = \dimX$ that explode
as the \Adam\ parameter $\beta_2$ approaches $1$. 
In this work we substantially extend these findings by establishing  
a priori bounds for \Adam\ for the whole class of \SOPs\ in \cref{main theorem} (including \cref{**} and \cref{***} 
above, respectively, 
as special cases) that are also uniform in the \Adam\ parameters $\beta_1$ and $\beta_2$ (see \cref{cor: priori bound stochastic Adam 1 non explosion} in \cref{subsec: deterministic bound} for details).
Moreover, for works that study \Adam\ and related gradient descent based optimizers in the context of deterministic 
optimization problems we mention, \eg, \cite{ar1911.07596}, \cite{Sebastianimprovement2022}, \cite{Bock2019APO}, \cite[Theorem 3.4]{ar1807.06766}, \cite{DeArAdsharp2025}, \cite{Polyak1964SomeMO}, \cite[Theorem 4.2]{Shi2021RMSpropCW}, and the references therein. In addition, for works that study 
limitations and lower bounds in the form of non-convergence results for \Adam\ and related methods 
we mention, \eg, \cite{Sebastiannonconvergence2025,CheriditoJentzenRossmanek2021,DeRoAr2024nonconvergence, HannibalJentzenThang2024,DoArAd2025,gallon2022blowphenomenagradientdescent,ArAd2024,LSSK2020,ReddiKale2019,ZhangChen2022}. A more detailed review of many of the above mentioned works and further articles 
on \Adam\ can be found in \cite[Section 1.3]{DeDoArPhi2025}. 
To the best of our knowledge, this work establishes for the first time unconditional 
(without including the boundedness of \Adam\ as an assumption of the error analysis) 
upper convergence bounds for the optimization error of \Adam; see \cref{eq4: main theorem} in \cref{main theorem}. Another conditional result for \Adam\ applied to rather general \SOPs\ is also given in \cite[Corollary 1.10]{DereichJentzenKassing2025} (cf., \eg, \cite[Theorem 3.2]{DeDoArPhi2025}) which shows that if \Adam\ does converge to a random variable, then this variable must be a limit of the \Adam\ vector field \cite[Definition 2.4]{DereichAdamconvergence2024}.

\subsection{Structure of this article}

The remainder of this work is organized as follows. In \cref{sec: non-explosion bound} we establish uniform a priori bounds for the \Adam\ optimizer applied to strongly convex \SOPs, which are also of independent interest. In \cref{sec: convergence of Adam} we combine the uniform a priori bounds from \cref{sec: non-explosion bound} with the error analysis for \Adam\ in our previous work \cite{DeDoArPhi2025} to establish an unconditional error analysis for \Adam\ (without including uniform a priori bounds as an assumption of the convergence rate analysis) and, thereby, we in particular prove \cref{main theorem} in this introductory section.

  \section{A priori bounds for strongly convex stochastic optimization problems (SOPs) }\label{sec: non-explosion bound}
  In this section we establish in \cref{lem: priori bound stochastic Adam non explosion} from \cref{subsec: explicit quantitative} and \cref{cor: priori bound stochastic Adam 1 non explosion} from \cref{subsec: deterministic bound} (which are the main results of this section) explicit pathwise a priori estimates for the \Adam\ optimizer when applied to general classes of strongly convex \SOPs. We prove \cref{cor: priori bound stochastic Adam 1 non explosion} through an application of the more general a priori bounds in \cref{lem: priori bound stochastic Adam non explosion}.

  In \cref{subsec: non-uniform priori} (see \cref{cor: priori bound stochastic Adam 2 explosion} in \cref{subsec: non-uniform priori}) we illustrate the qualitative a priori bounds for \Adam\ for general strongly convex \SOPs\ in \cref{lem: priori bound stochastic Adam non explosion} in the case of a simple example class of quadratic \SOPs\ with and without regularization but without uniformly controlling the first- and second-moment parameters of \Adam\ $\beta_1$ and $\beta_2$ (see \cref{def: V: cor: priori bound stochastic Adam 2 explosion} in \cref{cor: priori bound stochastic Adam 2 explosion}). 
  
In \cref{subsec: Example 2} (see \cref{cor: priori bound stochastic Adam 2 non explosion} and \cref{cor: priori bound stochastic Adam 2 non explosion copy} in \cref{subsec: Example 2}) we illustrate the quantitative a priori bounds for \Adam\ for general strongly convex \SOPs\ in \cref{cor: priori bound stochastic Adam 1 non explosion} in the case of simple example classes of quadratic \SOPs\ with and without regularization and with uniformly controlling, both, the first- and second-moment parameters of \Adam\ $\beta_1$ and $\beta_2$ (see \cref{def: bbV: cor: priori bound stochastic Adam 2 non explosion} in \cref{cor: priori bound stochastic Adam 2 non explosion} and \cref{def: bbV: cor: priori bound stochastic Adam 2 non explosion copy} in \cref{cor: priori bound stochastic Adam 2 non explosion copy}).
   \subsection{Quantitative a priori bounds for Adam for general strongly convex SOPs}\label{subsec: explicit quantitative}
    \begin{tcolorbox}[colback=white!95!gray,
                  colframe=black,
                  boxrule=0.5pt,
                  sharp corners,
                  enhanced,
                  breakable,
                 ]
  \begin{athm}{lemma}{V estimate}
    Let $\varepsilon\in (0,\infty)$, $\alpha\in [0,1)$, $\beta\in (\alpha^2,1)$. Then for all $n\in \N\cap(\log_\beta(\frac{\beta-\alpha^2}{1-\alpha^2}),\infty)$, $v\in [0,\infty)$ it holds that
    \begin{equation}\llabel{conclude}
    \textstyle  \alpha \Bigl[\varepsilon+\Bigl[\frac{\beta v}{1-\beta^n}\Bigr]^{\nicefrac{1}{2}}\Bigr]^{-1} \Bigl[\frac{1}{1-\alpha^n}\Bigr]\leq \textstyle \Bigl[\varepsilon+\Bigl[\frac{v }{1-\beta^{n-1}}\Bigr]^{\nicefrac{1}{2}}\Bigr]^{-1} \Bigl[\frac{1}{1-\alpha^{n-1}}\Bigr].
    \end{equation}
\end{athm}
\end{tcolorbox}
\begin{aproof}
Throughout this proof assume without loss of generality that $\alpha>0$.
    \argument{the assumption that $0<\alpha<1$;}{that for all $n\in \N\backslash\{1\}$ it holds that
    \begin{equation}\llabel{evidence 1}
        \frac{1}{1-\alpha^n}\leq \frac{1}{1-\alpha^{n-1}}.
    \end{equation}}
     \argument{the assumption that $1>\beta>\alpha^2>0$;}{that for all $n\in \N\cap(\log_\beta(\frac{\beta-\alpha^2}{1-\alpha^2}),\infty)$ it holds that
    \begin{equation}\llabel{eq1.1}
        \beta^n\leq \frac{\beta-\alpha^2}{1-\alpha^2}.
    \end{equation}}
    \argument{\lref{eq1.1};the fact that $\beta>0$}{that for all $n\in \N\cap[\log_\beta(\frac{\beta-\alpha^2}{1-\alpha^2}),\infty)$ it holds that
    \begin{equation}\llabel{eq1.2}
       1-\frac{\alpha^2}{\beta}= \frac{\beta-\alpha^2}{\beta}\geq \beta^{n-1}(1-\alpha^2)=\beta^{n-1}-\alpha^2\beta^{n-1}.
    \end{equation}}
    \argument{\lref{eq1.2};}{for all $n\in \N\cap(\log_\beta(\frac{\beta-\alpha^2}{1-\alpha^2}),\infty)$ that
    \begin{equation}\llabel{eq1.3}
        1-\beta^{n-1}\geq \frac{\alpha^2}{\beta}-\alpha^2\beta^{n-1}=\frac{\alpha^2(1-\beta^n)}{\beta}.
    \end{equation}}
    \argument{\lref{eq1.3}; the fact that $0<\beta<1$; the fact that $\alpha>0$}{that for all $n\in \N\cap(\log_\beta(\frac{\beta-\alpha^2}{1-\alpha^2}),\infty)$ it holds that
    \begin{equation}\llabel{eq1}
        \frac{1-\beta^n}{1-\beta^{n-1}}\leq \frac{\beta}{\alpha^2}.
    \end{equation}}
    \argument{\lref{eq1};}{for all $n\in \N\cap(\log_\beta(\frac{\beta-\alpha^2}{1-\alpha^2}),\infty)$ that
    \begin{equation}\llabel{eq2}
        \frac{\beta}{1-\beta^n}\geq \frac{\alpha^2}{1-\beta^{n-1}}.
    \end{equation}}
    \argument{\lref{eq2};the assumption that $0<\alpha<1$;the assumption that $\varepsilon>0$}{that for all $n\in \N\cap(\log_\beta(\frac{\beta-\alpha^2}{1-\alpha^2}),\infty)$, $v\in [0,\infty)$ it holds that
    \begin{equation}\llabel{eq3}
        \varepsilon+\sqrt{\frac{\beta v}{1-\beta^n}}\geq \varepsilon\alpha+\sqrt{\frac{\alpha^2 v}{1-\beta^{n-1}}}=\alpha\biggl(\varepsilon+\sqrt{\frac{v}{1-\beta^{n-1}}}\biggr).
    \end{equation}}
    \argument{\lref{eq3}}{for all $n\in \N\cap(\log_\beta(\frac{\beta-\alpha^2}{1-\alpha^2}),\infty)$, $v\in [0,\infty)$ that
    \begin{equation}\llabel{eq4}
        \textstyle  \Bigl[\varepsilon+\Bigl[\frac{\beta v}{1-\beta^n}\Bigr]^{\nicefrac{1}{2}}\Bigr]^{-1}\leq \displaystyle\alpha^{-1}\Bigl[\textstyle\varepsilon+\Bigl[\frac{v }{1-\beta^{n-1}}\Bigr]^{\nicefrac{1}{2}}\Bigr]^{-1}.
    \end{equation}}
    \argument{\lref{eq4};\lref{evidence 1}}{\lref{conclude}\dott}
\end{aproof}
 \begin{tcolorbox}[colback=white!95!gray,
                  colframe=black,
                  boxrule=0.5pt,
                  sharp corners,
                  enhanced,
                  breakable,
                 ]
\begin{athm}{lemma}{lem: bounded increament Adam}
  Let $\alpha\in [0,1)$, $\beta\in (\alpha^2,1)$, $\varepsilon\in (0,\infty)$, for every $n\in \N_0$ let $\bfm_n,\bbV_n\in \R$, assume $\bbV_0\geq 0$, and for every $n\in \N$ let $x_n\in\R$ satisfy 
    \begin{equation}\llabel{def: bfm}
        \bfm_n=\alpha\bfm_{n-1}+(1-\alpha)x_n\qqandqq \bbV_n=\beta\bbV_{n-1}+(1-\beta)|x_n|^2.
    \end{equation}
    Then it holds for all $n\in \N$ that
    \begin{equation}\llabel{conclude}
          \Bigl[\frac{|\bfm_{n}|}{1-\alpha^n}\Bigr] \Bigl[\varepsilon+\Bigl[\frac{\democrator_{n}}{1-\beta^n}\Bigr]^{\nicefrac{1}{2}}\Bigr]^{-1}\leq \frac{\bigl[\sum_{k=0}^{n-1}\beta^k\bigr]^{1/2}}{\sqrt{1-\alpha^2\beta^{-1}}}+\frac{|\bfm_{0}|}{\varepsilon(1-\alpha)}.
    \end{equation}
\end{athm}
\end{tcolorbox}
\begin{aproof}
      \argument{\lref{def: bfm};\unskip, \eg, \cite[Lemma 2.7]{DeRoArAd2025};}{that for all $n\in \N_0$ it holds that
\begin{equation}\llabel{need to prove 0}
     \bfm_{n}=(1-\alpha)\biggl[\textstyle\sum\limits_{k=1}^n\alpha^{n-k} x_k\biggr]+\alpha^n \bfm_{0}
\end{equation}
\begin{equation}\llabel{need to prove}
     \text{and}\qquad \bbV_{n}=(1-\beta)\biggl[\textstyle\sum\limits_{k=1}^n\beta^{n-k} |x_k|^2\biggr]+\beta^n \bbV_{0}.
\end{equation}}
            \argument{\lref{def: bfm};\lref{need to prove}; the assumption that $\bbV_{0}\geq 0$; Cauchy-Schwarz's inequality;the fact that $1>\beta>\alpha^2\geq 0$}{that for all $n\in \N$ it holds that
            \begin{equation}\llabel{eq2'}
            \begin{split}
           & |\bfm_{n}|\leq |\bfm_{0}|+(1-\alpha)\textstyle\biggl|\sum\limits_{k=1}^n \alpha^{n-k}
               x_k\biggr|\\
                &= |\bfm_{0}|+\textstyle(1-\alpha)\biggl|\sum\limits_{k=1}^n \alpha^{n-k}\beta^{-(n-k)/2}\beta^{
                (n-k)/2} x_k\biggr|\\
                &\leq |\bfm_{0}|+ (1-\alpha)\textstyle \biggl(\sum\limits_{k=1}^{n}\alpha^{2n-2k}\beta^{-(n-k)}\biggr)^{1/2}\biggl(\sum\limits_{k=1}^n\beta^{n-k}| x_k|^2\biggr)^{1/2}
                \leq |\bfm_{0}|+\frac{1-\alpha}{\sqrt{1-\alpha^2\beta^{-1}}} \bigl[\frac{\democrator_{n}}{1-\beta}\bigr]^{1/2}\\
                &\textstyle=|\bfm_{0}|+\frac{(1-\alpha)\bigl[\sum_{k=0}^{n-1}\beta^k\bigr]^{1/2}}{\sqrt{1-\alpha^2\beta^{-1}}} \bigl[\frac{\democrator_{n}}{1-\beta^n}\bigr]^{1/2}\leq \Bigl(\frac{(1-\alpha)\bigl[\sum_{k=0}^{n-1}\beta^k\bigr]^{1/2}}{\sqrt{1-\alpha^2\beta^{-1}}} +\frac{|\bfm_{0}|}{\varepsilon}\Bigr)\Bigl[\varepsilon+\bigl[\frac{\democrator_{n}}{1-\beta^n}\bigr]^{\nicefrac{1}{2}}\Bigr].
                \end{split}
            \end{equation}}
            \argument{\lref{eq2'};\lref{def: bfm};the assumption that $0\leq \alpha<1$;the assumption that $0<\beta<1$}{that for all $n\in \N$ it holds that
            \begin{equation}\llabel{eq3'}
            \begin{split}
              \Bigl[\frac{|\bfm_{n}|}{1-\alpha^n}\Bigr] \Bigl[\varepsilon+\Bigl[\frac{\democrator_{n}}{1-\beta^n}\Bigr]^{\nicefrac{1}{2}}\Bigr]^{-1}&\leq \frac{\bigl[\sum_{k=0}^{n-1}\beta^k\bigr]^{1/2}}{\sqrt{1-\alpha^2\beta^{-1}}}+\frac{|\bfm_{0}|}{\varepsilon(1-\alpha)}.
                \end{split}
            \end{equation}}
            \argument{\lref{eq3'}}{\lref{conclude}\dott}
\end{aproof}
 \begin{tcolorbox}[colback=white!95!gray,
                  colframe=black,
                  boxrule=0.5pt,
                  sharp corners,
                  enhanced,
                  breakable,
                 ]
\begin{athm}{prop}{lem: priori bound stochastic Adam non explosion}
    Let $\alpha\in [0,1)$, $\beta\in (\alpha^2,1)$, $M,\fd,\dimX\in \N$, let $O\subseteq \R^\dimX$ be open, let $\setX\subseteq O$ be compact and convex, let
     $\smalll=(\smalll(\theta,x))_{(\theta,x)\in \R^\fd\times O}\allowbreak\in C^{1}(\R^\fd\times O,\R)$, for every $n,m\in\N$ let $X_{n,m}\in \setX$, let  $\cK,\kappa,\varepsilon,\tau_2\in (0,\infty)$, $\bfM,\cV,\tau_1\in [0,\infty)$ satisfy $3\tau_1+4\tau_2+(\tau_1)^2+2\tau_1\tau_2<2$ and $\tau_1+2\tau_2<1$, let $(\gamma_n)_{n\in \N}\subseteq [0,\infty)$ be non-increasing,
     for every $n\in \N_0$ let $\Theta_n=(\Theta_{n}^{1},\dots,\Theta_{n}^{\fd})$, $\bfm_n=(\bfm_{n,1},\dots,\bfm_{n,\fd})$,  $\democrator_n=(\democrator_{n,1},\dots,\democrator_{n,\fd})\in\R^\fd$, assume for all $n\in \N$, $i\in \{1,2,\dots,\fd\}$ that
     \begin{equation} \llabel{def: bfm}
    |\bfm_{0,i}| \leq \bfM,\qquad \bfm_{n,i}=\alpha \bfm_{n-1,i}+(1-\alpha)\bigl[\textstyle \frac 1M \sum_{j=1}^M(\frac{\partial}{\partial\theta_i} \smalll)(\Theta_{n-1},X_{n,j})\bigr],
    \end{equation}
    \begin{equation} \llabel{def: V}
    0\leq \democrator_{0,i}\leq \cV^2,\qquad \democrator_{n,i}=\beta \democrator_{n-1,i}+(1-\beta)\bigl[\textstyle \frac 1M \sum_{j=1}^M(\frac{\partial}{\partial\theta_i} \smalll)(\Theta_{n-1},X_{n,j})\bigr]^2,
    \end{equation}
    \begin{equation}\llabel{def: Theta}
    \text{and}\qquad\textstyle\Theta_{n}^i=\Theta_{n-1}^i -\gamma_n\bigl[\frac{\bfm_{n,i}}{1-\alpha^n}\bigr] \Bigl[\varepsilon+\bigl[\frac{\democrator_{n,i}}{1-\beta^n}\bigr]^{\nicefrac{1}{2}}\Bigr]^{-1}, 
    \end{equation}
    let $\scrc\in [1,\infty)$ satisfy $\setX\subseteq\{x\in \R^\dimX\colon \|x\|\leq \scrc\}$,
   assume for all $\theta,\vartheta\in \R^\fd$, $x\in \setX$ that $ \|(\nabla_\theta\smalll)(\theta,x)- (\nabla_\theta\smalll)(\vartheta,x)\|\leq  \cK\|\theta-\vartheta\|(1+\|\theta\|^{\tau_1}+\|\vartheta\|^{\tau_1})$, $ \kappa\|\theta-\vartheta\|^2\leq \spro{ \theta-\vartheta, (\nabla_\theta\smalll)(\theta,x)- (\nabla_\theta\smalll)(\vartheta,x)} $, and
 \begin{equation}\llabel{assumption}
 \|(\nabla_x\smalll)(\theta,x)-(\nabla_x\smalll)(\vartheta,x)\|\leq \cK\|\theta-\vartheta\|(1+\|\theta\|^{\tau_2}+\|\vartheta\|^{\tau_2}),
 \end{equation}
  let $\eta,\Gamma\in \R$ satisfy
 \begin{equation}\llabel{def: eta}
     \eta=\frac{d}{\sqrt{(\beta-\alpha^2)(\beta^{-1}-1)}}+\frac{\bfM d}{\varepsilon(1-\alpha)} \qqandqq \Gamma=\sum_{n=1}^\infty\tau_1\eta^2(\gamma_n)^2,
 \end{equation}
 let $N,\fn\in \N$ satisfy \begin{equation}\llabel{def: N}
\textstyle\sup_{n\in \N\cap[N,\infty)}\max\{\cK\varepsilon^{-1}\gamma_n\bigl(3+(\gamma_1\eta )^{\tau_1}\bigr),2\cK\scrc d^{\tau_2}(1+\tau_2)(2\varepsilon^{-1}\gamma_n)^{\tau_2}\}\leq \frac 18
 \end{equation}
  \begin{equation}\llabel{def: fn}
     \text{and}\qquad \fn=\bigl\lceil\max\{\textstyle N,-[\log_{2}(\alpha+2\mathbbm 1_{\{0\}}(\alpha))]^{-1},\log_\beta(\frac{\beta-\alpha^2}{1-\alpha^2})+1\}\bigr\rceil\ifnocf,
  \end{equation}
 let $\vartheta\in \R^\fd$, $\bfx\in \setX$, $\rho,\delta,\chi\in \R$ satisfy
    \begin{equation}\llabel{def: vartheta}
      \textstyle \smalll(\vartheta,\bfx)=\inf_{\theta\in \R^\fd}\smalll(\theta,\bfx),\qquad\rho=1+\sup_{x\in \setX}\|\nabla_\theta\smalll(\vartheta,x)\|,
    \end{equation}
    \begin{equation}\llabel{def: delta}
       \delta=\frac{ \fn^{1/2} d}{\sqrt{1-\alpha^2\beta^{-1}}}+\frac{\bfM d}{\varepsilon(1-\alpha)},\qqandqq \chi=\max\biggl\{\frac{\alpha}{1-\alpha},4\biggr\},
    \end{equation}
  and let $(\cA_n)_{n\in \N}\subseteq\R$ satisfy
  \begin{equation}\llabel{def: A1}
  \begin{split}
      \cA_1&=\frac{\cK}{2}( \fn\gamma_1\delta+\|\Theta_0-\vartheta\|)^2(1+(\fn\gamma_1\delta+\|\Theta_0-\vartheta\|)^{\tau_1}+2\|\vartheta\|^{\tau_1})\\
        &+\frac{d}{2(1-\alpha)} \bigl[ \cK(\fn\gamma_1\delta+\|\Theta_0-\vartheta\|)(1+(\fn\gamma_1\delta+\|\Theta_0-\vartheta\|)^{\tau_1}+2\|\vartheta\|^{\tau_1})+\rho+\bfM\bigr]\gamma_1\delta,
        \end{split}
  \end{equation}
  \begin{equation}\llabel{def: A2}
     \textstyle \cA_2=\max\bigl\{\frac{8(1+\Gamma)\cK^2}{\kappa d^2},\frac{\kappa}{2},\frac{\kappa\|\vartheta\|^2}{2},\frac{\rho^2\kappa}{2\cK^2},\kappa\bigl(\frac{\cV^2}{\cK^2(1-\beta)}\bigr)^{1/(1+\tau_1)},\kappa\bigl(\frac{\varepsilon}{\cK}\bigr)^{2/(1+\tau_1)},\bigl(\frac{1+32\cK^2\scrc^2}{\cK^2\scrc^2}\bigr)^{1/\tau_2}\kappa\bigr\},
  \end{equation}
  \begin{equation}\llabel{def: A3}
      \cA_3=\frac{\rho^{4/(1+\tau_1+2\tau_2)}\kappa}{(\chi d\varepsilon^{-1}\scrc^2\cK^3)^{2/(1+\tau_1+2\tau_2)}(1+\Gamma)},
  \end{equation}
  \begin{equation}\llabel{def: A4}
  \begin{split}
      \cA_4&=\max\bigl\{1610(1+\gamma_1)\cK^{3/2}\kappa^{-(5+\tau_1+2\tau_2)/4}(1+\kappa)\scrc\varepsilon^{-3/2}(1+\varepsilon)(1+\chi^{1/2}) \\
      &\quad \cdot d(1+\Gamma)^{(1+\tau_1+2\tau_2)/4},1\bigr\},
      \end{split}
  \end{equation}
  \begin{equation}\llabel{def: A5}
  \cA_5=\bigl(2\cK(1+\|\vartheta\|^{\tau_1})(\cA_4)^{2+\tau_1}\bigr)^{4/(2-3\tau_1-4\tau_2-(\tau_1)^2-2\tau_1\tau_2)},
  \end{equation}
  \begin{equation}\llabel{def: A6}
     \textstyle \text{and}\qquad \cA_6=\bigl(328000(1-\alpha)^{-1}\gamma_1\cK^3\kappa^{-(1+\tau_1+2\tau_2)/2}\scrc^2d\varepsilon^{-2}(1+\Gamma)^{(1+\tau_1+2\tau_2)/2}\bigr)^{2/(1-\tau_1-2\tau_2)}.
  \end{equation}
 Then 
   \begin{equation}\llabel{conclude}
   \begin{split}
\textstyle\sup_{n\in\N}\|\Theta_n\|\leq \|\vartheta\|+\sup_{n\in \N}\|\Theta_n-\vartheta\|\leq  \|\vartheta\|+\bigl[2\kappa^{-1}\textstyle\max\{1,\sup_{n\in \N}\cA_n\}(1+\Gamma)\bigr]^{1/2}.
\end{split}
   \end{equation}
\end{athm}
\end{tcolorbox}
\begin{aproof}
      Throughout this proof let $\bfA\in \R$ satisfy
      \begin{equation}\llabel{def: bfA}
          \bfA=\max\{1,\cA_1,\cA_2,\cA_3,\cA_5,\cA_6\},
      \end{equation}
      let $ \cG_1=(\cG_1^1,\dots,\cG_1^{d+\dimX})  \colon \R^d\times O \to \R^d $, $ \cG_2=(\cG_2^1,\dots,\cG_2^{d+\dimX})  \colon \R^d\times O \to \R^\dimX $ satisfy for all $\theta \in \R^\fd$, $x \in O$ that
\begin{equation}\llabel{def: cG}
    \cG_1( \theta, x ) = ( \nabla_{ \theta } \smalll )( \theta, x ) \qqandqq \cG_2( \theta, x ) = (\nabla_x \smalll )( \theta,x ),
\end{equation}
     and for every $n\in \N$ let $\bfV_n=(\bfV_{n,1},\dots,\bfV_{n,\fd})\in \R^d$ satisfy for all $i\in\{1,2,\dots,\fd\}$ that
      \begin{equation}\llabel{def: bfV}
\textstyle\bfV_{n,i}=\frac{\gamma_n}{1-\alpha^n}\Bigl[\varepsilon+\bigl[\frac{\democrator_{n,i}}{1-\beta^n}\bigr]^{\nicefrac{1}{2}}\Bigr]^{-1}
      \end{equation}
  and let $H_n\in \R$ satisfy 
  \begin{equation}\llabel{def: H}
    H_n=\smalll(\Theta_n,\bfx)+\frac{1}{2(1-\alpha)}\textstyle\biggl[\sum\limits_{i=1}^d(\bfm_{n,i})^2\bfV_{n,i}\biggr].
  \end{equation}
  \argument{the fundamental theorem of calculus}{that for all $\cL\in C^1(\R^{\fd},\R)$, $\theta_1,\theta_2\in \R^\fd$ it holds that
\begin{equation}\llabel{eqp1}
\begin{split}
    &\cL(\theta_1)-\cL(\theta_2)
    =\int_{0}^1\spro{(\nabla\cL)(\theta_2+t(\theta_1-\theta_2)),\theta_1-\theta_2}\,\d t\\
    &=\spro{(\nabla\cL)(\theta_2),\theta_1-\theta_2}+\int_{0}^1\spro{(\nabla\cL)(\theta_2+t(\theta_1-\theta_2))-(\nabla\cL)(\theta_2),\theta_1-\theta_2}\,\d t.
    \end{split}
\end{equation}}
\argument{\lref{assumption};\lref{def: cG}}{that for all $m\in \N$, $\theta_1,\theta_2\in \R^\fd$, $x_1,x_2,\dots,x_m\in \setX$ it holds that
\begin{equation}\llabel{eqp1.1}
    \kappa\|\theta_1-\theta_2\|^2\leq \bigl\langle \theta_1-\theta_2, \textstyle\bigl(\frac 1m \textstyle\sum_{j=1}^m \cG_1(\theta_1,x_j)\bigr)- \bigl(\textstyle\frac 1m\textstyle\sum_{j=1}^m \cG_1(\theta_2,x_j)\bigr)\bigr\rangle.
\end{equation}}
\argument{\lref{assumption};\lref{eqp1.1};}{that for all $m,n\in \N$, $\theta_1,\theta_2\in \R^\fd$, $x_1,x_2,\dots,x_m\in \setX$ it holds that
\begin{equation}\llabel{assumption 2}
\begin{split}
    &\kappa\|\theta_1-\theta_2\|^2\leq \bigl\langle \theta_1-\theta_2, \bigl(\textstyle\frac 1m \textstyle\sum_{j=1}^m \cG_1(\theta_1,x_j)\bigr)-\bigl(\textstyle\frac 1m \textstyle\sum_{j=1}^m \cG_1(\theta_2,x_j)\bigr)\bigr\rangle\\
    &\leq \cK\|\theta_1-\theta_2\|^2(1+\|\theta_1\|^{\tau_1}+\|\theta_2\|^{\tau_1}).
    \end{split}
\end{equation}}
\argument{\lref{assumption 2};\lref{eqp1};the fact that for all $x,y\in \R^d$ it holds that $\|x+y\|^{\tau_1}\leq \|x\|^{\tau_1}+\|y\|^{\tau_1}$}{that for all $m\in \N$, $\theta_1,\theta_2\in \R^\fd$, $x_1,x_2,\dots,x_m\in \setX$ it holds that
\begin{align}
   &\bigl[\textstyle \frac 1m \sum_{j=1}^m\smalll(\theta_1,x_{j})\bigr]-\bigl[\textstyle \frac 1m \sum_{j=1}^m \smalll(\theta_2,x_{j})\bigr]\notag\\
   &\leq \bigl\langle\bigl(\textstyle \frac 1m \sum_{j=1}^m \cG_1(\theta_2,x_{j})\bigr),\theta_1-\theta_2\bigr\rangle+ \displaystyle\int_{0}^1 t\cK\|\theta_1-\theta_2\|^2(1+\|\theta_2+t(\theta_1-\theta_2)\|^{\tau_1}+\|\theta_2\|^{\tau_1})\,\d t\notag\\
   &\leq \bigl\langle\bigl(\textstyle \frac 1m \sum_{j=1}^m \cG_1(\theta_2,x_{j})\bigr),\theta_1-\theta_2\bigr\rangle+ \displaystyle\int_{0}^1 t\cK\|\theta_1-\theta_2\|^2(1+\|\theta_2\|^{\tau_1}+\|t(\theta_1-\theta_2)\|^{\tau_1}+\|\theta_2\|^{\tau_1})\,\d t\notag\\
    &\leq \bigl\langle\bigl(\textstyle \frac 1m \sum_{j=1}^m \cG_1(\theta_2,x_{j})\bigr),\theta_1-\theta_2\bigr\rangle+\frac{\cK}{2}\|\theta_1-\theta_2\|^2(1+\|\theta_1-\theta_2\|^{\tau_1}+2\|\theta_2\|^{\tau_1})\llabel{eqp2'}.
\end{align}}
 \argument{\lref{def: Theta};\lref{def: bfV};the fact that for all $x=(x_1,\dots,x_\fd)\in \R^\fd$ it holds that $\|x\|\leq \sum_{i=1}^\fd |x_i|$}{that for all $n\in \N$ it holds that
\begin{equation}\llabel{eq: x2.5}
\begin{split}
    &\textstyle\|\Theta_n-\Theta_{n-1}\|\leq  \sum\limits_{i=1}^d|\Theta_n^i-\Theta_{n-1}^i|= \sum\limits_{i=1}^d\gamma_n\bigl[\frac{|\bfm_{n,i}|}{1-\alpha^n}\bigr] \Bigl[\varepsilon+\bigl[\frac{\democrator_{n,i}}{1-\beta^n}\bigr]^{\nicefrac{1}{2}}\Bigr]^{-1}=\textstyle\sum\limits_{i=1}^d|\bfm_{n,i}|\bfV_{n,i}.
    \end{split}
\end{equation}}
\argument{\lref{def: bfV};\lref{eqp2'}; \lref{eq: x2.5}; the fact that for all $m,n\in \N$ it holds that $X_{m,n}\in \setX$; the assumption that for all $n\in\N$ it holds that $\gamma_n>0$;}{that for all $n\in \N$ it holds that
    \begin{align}
       &\bigl[\textstyle \frac 1M \sum_{j=1}^M\smalll(\Theta_n,X_{n,j})\bigr]-\bigl[\textstyle \frac 1M \sum_{j=1}^M \smalll(\Theta_{n-1},X_{n,j})\bigr]\notag\\
        &\leq \bigl\langle\bigl(\textstyle \frac 1M \sum_{j=1}^M \cG_1(\Theta_{n-1},X_{n,j})\bigr),\Theta_n-\Theta_{n-1}\bigr\rangle
        +\frac{\cK}{2}\|\Theta_n-\Theta_{n-1}\|^2(1+\|\Theta_n-\Theta_{n-1}\|^{\tau_1}+2\|\Theta_{n-1}\|^{\tau_1})\notag\\
        &\leq -\textstyle\bigl[\sum_{i=1}^\fd \bigl[\textstyle \frac 1M \sum_{j=1}^M\cG_1^i(\Theta_{n-1},X_{n,j})\bigr]\bfm_{n,i}\bfV_{n,i} \bigr]\notag\\
        &\textstyle+\tfrac{\cK}{2}\bigl[\textstyle\sum_{i=1}^\fd (\bfm_{n,i})^2(\bfV_{n,i})^2\bigr]\bigl(1+\bigl[\sum_{i=1}^d|\bfm_{n,i}|\bfV_{n,i}\bigr]^{\tau_1}+2\|\Theta_{n-1}\|^{\tau_1}\bigr)\llabel{eq4'}.
    \end{align}}
     \argument{\lref{def: cG};the fundamental theorem of calculus; the assumption that $\smalll\in C^1(\R^d\times O,\R)$; the asssumption that $\setX$ is convex}{that for all $\theta_1,\theta_2\in \R^d$, $x\in \setX$ it holds that
    \begin{equation}\llabel{eq: x}
    \begin{split}
       & \smalll(\theta_1,x)-\smalll(\theta_1,\bfx)-\smalll(\theta_2,x)+\smalll(\theta_2,\bfx)\\
      &=\int_0^1\spro{\cG_2(\theta_1,\bfx+t(x-\bfx)),x-\bfx}\,\d t-\int_0^1\spro{\cG_2(\theta_2,\bfx+t(x-\bfx)),x-\bfx}\,\d t.
        \end{split}
    \end{equation}}
    \argument{\lref{eq: x};\lref{assumption}; the Cauchy-Schwarz inequality; the assumption that $\setX$ is convex; the fact for all $x\in \setX$ it holds that $\|x\|\leq \scrc$}{that for all  $\theta_1,\theta_2\in \R^d$, $x\in \setX$ it holds that
    \begin{equation}\llabel{eq: x2}
         \begin{split}
       & |\smalll(\theta_1,x)-\smalll(\theta_1,\bfx)-\smalll(\theta_2,x)+\smalll(\theta_2,\bfx)|\\
      &\leq \int_0^1|\spro{\cG_2(\theta_1,\bfx+t(x-\bfx)),x-\bfx}-\spro{\cG_2(\theta_2,\bfx+t(x-\bfx)),x-\bfx}|\,\d t\\
      &\leq \int_0^1\|\cG_2(\theta_1,\bfx+t(x-\bfx))-\cG_2(\theta_2,\bfx+t(x-\bfx))\|\|x-\bfx\|\,\d t\\
      &\leq \int_0^1\cK\|x-\bfx\| \|\theta_1-\theta_2\|(1+\|\theta_1\|^{\tau_2}+\|\theta_2\|^{\tau_2})\,\d t\\
      &\leq  2\cK\scrc \|\theta_1-\theta_2\|(1+\|\theta_1\|^{\tau_2}+\|\theta_2\|^{\tau_2}).
        \end{split}
    \end{equation}}
    \argument{\lref{eq: x2};\lref{eq: x2.5};the fact that for all $i,j\in \N$ it holds that $X_{i,j}\in \setX$; the fact that for all $x,y\in (0,\infty)$ it holds that $|x+y|^{\tau_2}\leq |x|^{\tau_2}+|y|^{\tau_2}$;the assumption that $0<\beta<1$}{that for all $n\in \N$ it holds that
    \begin{equation}\llabel{eq: x3}
    \begin{split}
     &\biggl|\biggl[\displaystyle\frac 1M \textstyle\sum\limits_{j=1}^M\smalll(\Theta_n,X_{n,j})\biggr]-\smalll(\Theta_n,\bfx)-\biggl[\displaystyle\frac 1M \textstyle\sum\limits_{j=1}^M\smalll(\Theta_{n-1},X_{n,j})\biggr]+\smalll(\Theta_{n-1},\bfx)\biggr|\notag\\
     &\leq \sup_{j\in \{1,2,\dots,M\}} |\smalll(\Theta_n,X_{n,j})-\smalll(\Theta_n,\bfx)-\smalll(\Theta_{n-1},X_{n,j})+\smalll(\Theta_{n-1},\bfx)|\\
     &\leq 2\cK\scrc\|\Theta_n-\Theta_{n-1}\|(1+\|\Theta_n\|^{\tau_2}+\|\Theta_{n-1}\|^{\tau_2})\\
     &
     \leq 2\cK\scrc\textstyle\biggl[\sum\limits_{i=1}^d|\bfm_{n,i}|\bfV_{n,i}\biggr](1+\|\Theta_n\|^{\tau_2}+\|\Theta_{n-1}\|^{\tau_2})\\
     &\leq 2\cK\scrc\textstyle\biggl[\sum\limits_{i=1}^d|\bfm_{n,i}|\bfV_{n,i}\biggr]\biggl(1+\|\Theta_{n-1}\|^{\tau_2}+\biggl[\sum\limits_{i=1}^d|\bfm_{n,i}|\bfV_{n,i}\biggr]^{\tau_2}+\|\Theta_{n-1}\|^{\tau_2}\biggr)\\
     &\leq 2\cK\scrc\textstyle\biggl[\sum\limits_{i=1}^d|\bfm_{n,i}|\bfV_{n,i}\biggr]^{1+\tau_2}+2\cK\scrc\textstyle\biggl[\sum\limits_{i=1}^d|\bfm_{n,i}|\bfV_{n,i}\biggr](1+2\|\Theta_{n-1}\|^{\tau_2})
     .
     \end{split}
    \end{equation}}
        \argument{\lref{eq: x3};\lref{eq4'}}{for all $n\in \N$ that
    \begin{equation}\llabel{eq4}
    \begin{split}
        &\smalll(\Theta_n,\bfx)-\smalll(\Theta_{n-1},\bfx)\leq
-\textstyle\bigl[\sum_{i=1}^\fd \bigl[\textstyle \frac 1M \sum_{j=1}^M\cG_1^i(\Theta_{n-1},X_{n,j})\bigr]\bfm_{n,i}\bfV_{n,i} \bigr]\\
&+\tfrac{\cK}{2}\bigl[\textstyle\sum_{i=1}^\fd (\bfm_{n,i})^2(\bfV_{n,i})^2\bigr]\bigl(1+\bigl[\sum_{i=1}^d|\bfm_{n,i}|\bfV_{n,i}\bigr]^{\tau_1}+2\|\Theta_{n-1}\|^{\tau_1}\bigr)\\
&+ 2\cK\scrc\textstyle\bigl[\sum_{i=1}^d|\bfm_{n,i}|\bfV_{n,i}\bigr]^{1+\tau_2}+2\cK\scrc\textstyle\bigl[\sum_{i=1}^d|\bfm_{n,i}|\bfV_{n,i}\bigr](1+2\|\Theta_{n-1}\|^{\tau_2}).
        \end{split}
    \end{equation}}
      \argument{\lref{def: bfm};\lref{def: cG}}{that for all $n\in \N$, $i\in \{1,2,\dots,d\}$ it holds that
  \begin{equation}\llabel{eq1}
      \bfm_{n-1,i}=\bfm_{n,i}-(1-\alpha)\bigl[\textstyle\textstyle \frac 1M \bigl[\sum_{j=1}^M\cG_1^i(\Theta_{n-1},X_{n,j})\bigr]-\bfm_{n-1,i}\bigr].
  \end{equation}}
  \argument{\lref{eq1};}{for all $n\in \N$, $i\in \{1,2,\dots,d\}$ that
  \begin{equation}\llabel{eq2}
  \begin{split}
      (\bfm_{n,i})^2-(\bfm_{n-1,i})^2&=2(1-\alpha)\bfm_{n,i}\bigl[\textstyle\frac 1M \bigl[\sum_{j=1}^M\cG_1^i(\Theta_{n-1},X_{n,j})\bigr]-\bfm_{n-1,i}\bigr]\\
      &-(1-\alpha)^2\bigl[\textstyle\frac 1M \bigl[\sum_{j=1}^M\cG_1^i(\Theta_{n-1},X_{n,j})\bigr]-\bfm_{n-1,i}\bigr]^2.
        \end{split}
  \end{equation}}
  \argument{\lref{eq2};}{that for all $n\in \N\backslash\{1\}$ it holds that
  \begin{align}
    &\frac{1}{2(1-\alpha)}\biggl[\textstyle\sum\limits_{i=1}^d(\bfm_{n,i})^2\bfV_{n,i} \biggr]\notag\\
    &=\displaystyle\frac{1}{2(1-\alpha)}\biggl[\textstyle\sum\limits_{i=1}^d(\bfm_{n-1,i})^2\bfV_{n-1,i} \biggr]+\displaystyle\frac{1}{2(1-\alpha)}\biggl[\textstyle\sum\limits_{i=1}^d\bigl((\bfm_{n,i})^2-(\bfm_{n-1,i})^2\bigr)\bfV_{n,i} \biggr]\notag
    \\
    &+\displaystyle\frac{1}{2(1-\alpha)}\biggl[\textstyle\sum\limits_{i=1}^d(\bfm_{n-1,i})^2(\bfV_{n,i}-\bfV_{n-1,i}) \biggr]\notag\\
     &=\displaystyle\frac{1}{2(1-\alpha)}\biggl[\textstyle\sum\limits_{i=1}^d(\bfm_{n-1,i})^2\bfV_{n-1,i} \biggr]+\biggl[\textstyle\sum\limits_{i=1}^d\bfm_{n,i}\bigl[\textstyle\textstyle \frac 1M \bigl[\sum_{j=1}^M\cG_1^i(\Theta_{n-1},X_{n,j})\bigr]-\bfm_{n-1,i}\bigr]\bfV_{n,i} \biggr]\notag
    \\
    &-\displaystyle\frac{1-\alpha}{2}\biggl[\textstyle\sum\limits_{i=1}^d\bigl[\textstyle\textstyle \frac 1M \bigl[\sum_{j=1}^M\cG_1^i(\Theta_{n-1},X_{n,j})\bigr]-\bfm_{n-1,i}\bigr]^2\bfV_{n,i} \biggr]\notag\\
    &+\displaystyle\frac{1}{2(1-\alpha)}\biggl[\textstyle\sum\limits_{i=1}^d(\bfm_{n-1,i})^2(\bfV_{n,i}-\bfV_{n-1,i}) \biggr]\llabel{eq5}.
  \end{align}}
  \argument{\lref{eq5};\lref{def: H};\lref{eq4}}{for all $n\in \N\backslash\{1\}$ that
  \begin{equation}\llabel{eq6}
  \begin{split}
      H_n
      &\leq H_{n-1}+\displaystyle\frac{\cK}{2}\biggl[\textstyle\sum\limits_{i=1}^d (\bfm_{n,i})^2(\bfV_{n,i})^2\biggr]\biggl(1+\biggl[\sum\limits_{i=1}^d|\bfm_{n,i}|\bfV_{n,i}\biggr]^{\tau_1}+2\|\Theta_{n-1}\|^{\tau_1}\biggr)\\
      &-\biggl[\textstyle\sum\limits_{i=1}^d\bfm_{n,i}\bfm_{n-1,i}\bfV_{n,i} \biggr]-\displaystyle\frac{1-\alpha}{2}\biggl[\textstyle\sum\limits_{i=1}^d\bigl[\frac 1M \bigl[\sum_{j=1}^M\cG_1^i(\Theta_{n-1},X_{n,j})\bigr]-\bfm_{n-1,i}\bigr]^2\bfV_{n,i} \biggr]\\
      &+\displaystyle\frac{1}{2(1-\alpha)}\biggl[\textstyle\sum\limits_{i=1}^d(\bfm_{n-1,i})^2(\bfV_{n,i}-\bfV_{n-1,i}) \biggr]+ 2\cK\scrc\textstyle\biggl[\sum\limits_{i=1}^d|\bfm_{n,i}|\bfV_{n,i}\biggr]^{1+\tau_2}\\
&+2\cK\scrc\textstyle\biggl[\sum\limits_{i=1}^d|\bfm_{n,i}|\bfV_{n,i}\biggr](1+2\|\Theta_{n-1}\|^{\tau_2}).
      \end{split}
  \end{equation}}
  \argument{\lref{eq6};\lref{eq1}}{for all $n\in \N\backslash\{1\}$ that
  \begin{equation}\llabel{eq7}
      \begin{split}
      H_n
      &\leq H_{n-1}+\displaystyle\frac{\cK}{2}\biggl[\textstyle\sum\limits_{i=1}^d (\bfm_{n,i})^2(\bfV_{n,i})^2\biggr]\biggl(1+\biggl[\sum\limits_{i=1}^d|\bfm_{n,i}|\bfV_{n,i}\biggr]^{\tau_1}+2\|\Theta_{n-1}\|^{\tau_1}\biggr)\\
      &-\biggl[\textstyle\sum\limits_{i=1}^d(\bfm_{n,i})^2\bfV_{n,i} \biggr]
      +(1-\alpha)\biggl[\textstyle\sum\limits_{i=1}^d\bfm_{n,i}\bigl[\textstyle\frac 1M \bigl[\sum_{j=1}^M\cG_1^i(\Theta_{n-1},X_{n,j})\bigr]-\bfm_{n-1,i}\bigr]\bfV_{n,i} \biggr]\\
      &-\displaystyle\frac{1-\alpha}{2}\biggl[\textstyle\sum\limits_{i=1}^d\bigl[\textstyle\frac 1M \bigl[\sum_{j=1}^M\cG_1^i(\Theta_{n-1},X_{n,j})\bigr]-\bfm_{n-1,i}\bigr]^2\bfV_{n,i} \biggr]\\
      &+\displaystyle\frac{1}{2(1-\alpha)}\biggl[\textstyle\sum\limits_{i=1}^d(\bfm_{n-1,i})^2(\bfV_{n,i}-\bfV_{n-1,i}) \biggr]+ 2\cK\scrc\textstyle\biggl[\sum\limits_{i=1}^d|\bfm_{n,i}|\bfV_{n,i}\biggr]^{1+\tau_2}\\
&+2\cK\scrc\textstyle\biggl[\sum\limits_{i=1}^d|\bfm_{n,i}|\bfV_{n,i}\biggr](1+2\|\Theta_{n-1}\|^{\tau_2}).
      \end{split}
  \end{equation}}
    \argument{\lref{def: V};}{that for all $n\in \N\backslash\{1\}$, $i\in \{1,2,\dots,d\}$ it holds that
  \begin{equation}\llabel{eq8'}
      \bbV_{n,i}\geq \beta\bbV_{n-1,i}\geq 0.
  \end{equation}}
  \argument{\lref{def: V};\lref{eq8'};\lref{def: bfV};\cref{V estimate}; the fact that for all $n\in \N\backslash\{1\}$ it holds that $\gamma_{n}\leq \gamma_{n-1}$}{that for all $n\in \N\cap(\log_\beta(\frac{\beta-\alpha^2}{1-\alpha^2}),\infty)$,  $i\in \{1,2,\dots,\fd\}$ it holds that
  \begin{equation}\llabel{eq9'}
  \begin{split}
\alpha\bfV_{n,i}&=\textstyle\alpha\gamma_n\Bigl[\varepsilon+\Bigl[\frac{\democrator_{n,i}}{1-\beta^n}\Bigr]^{\nicefrac{1}{2}}\Bigr]^{-1} \Bigl[\frac{1}{1-\alpha^n}\Bigr]\leq \alpha \gamma_{n-1}\Bigl[\varepsilon+\Bigl[\frac{\beta\democrator_{n-1,i}}{1-\beta^n}\Bigr]^{\nicefrac{1}{2}}\Bigr]^{-1} \Bigl[\frac{1}{1-\alpha^n}\Bigr]\\
&\leq \gamma_{n-1}\Bigl[\varepsilon+\Bigl[\textstyle\frac{\democrator_{n-1,i}}{1-\beta^{n-1}}\Bigr]^{\nicefrac{1}{2}}\Bigr]^{-1} \Bigl[\frac{1}{1-\alpha^{n-1}}\Bigr]=\bfV_{n-1,i}.
\end{split}
  \end{equation}}
  \argument{\lref{eq9'};\lref{def: fn};\lref{eq7}}{for all $n\in \N\cap[\fn,\infty)$ that
  \begin{equation}\llabel{eq10}
      \begin{split}
      H_n
      &\leq H_{n-1}+\displaystyle\frac{\cK}{2}\biggl[\textstyle\sum\limits_{i=1}^d (\bfm_{n,i})^2(\bfV_{n,i})^2\biggr]\biggl(1+\biggl[\sum\limits_{i=1}^d|\bfm_{n,i}|\bfV_{n,i}\biggr]^{\tau_1}+2\|\Theta_{n-1}\|^{\tau_1}\biggr)\\
      &-\biggl[\textstyle\sum\limits_{i=1}^d(\bfm_{n,i})^2\bfV_{n,i} \biggr]+(1-\alpha)\biggl[\textstyle\sum\limits_{i=1}^d\bfm_{n,i}\bigl[\textstyle\frac 1M \bigl[\sum_{j=1}^M\cG_1^i(\Theta_{n-1},X_{n,j})\bigr]-\bfm_{n-1,i}\bigr]\bfV_{n,i} \biggr]\\
      &-\displaystyle\frac{1-\alpha}{2}\biggl[\textstyle\sum\limits_{i=1}^d\bigl[\textstyle\frac 1M \bigl[\sum_{j=1}^M\cG_1^i(\Theta_{n-1},X_{n,j})\bigr]-\bfm_{n-1,i}\bigr]^2\bfV_{n,i} \biggr]+\displaystyle\frac{1}{2}\biggl[\textstyle\sum\limits_{i=1}^d(\bfm_{n-1,i})^2\bfV_{n,i} \biggr]\\
      &+ 2\cK\scrc\textstyle\biggl[\sum\limits_{i=1}^d|\bfm_{n,i}|\bfV_{n,i}\biggr]^{1+\tau_2}+2\cK\scrc\textstyle\biggl[\sum\limits_{i=1}^d|\bfm_{n,i}|\bfV_{n,i}\biggr](1+2\|\Theta_{n-1}\|^{\tau_2}).
      \end{split}
  \end{equation}}
  \argument{\lref{eq10};\lref{eq2}}{for all $n\in \N\cap[\fn,\infty)$ that
  \begin{equation}\llabel{eq11}
      \begin{split}
          H_n&\leq H_{n-1}+\displaystyle\frac{\cK}{2}\biggl[\textstyle\sum\limits_{i=1}^d (\bfm_{n,i})^2(\bfV_{n,i})^2\biggr]\biggl(1+\biggl[\sum\limits_{i=1}^d|\bfm_{n,i}|\bfV_{n,i}\biggr]^{\tau_1}+2\|\Theta_{n-1}\|^{\tau_1}\biggr)\\
          &-\biggl[\textstyle\sum\limits_{i=1}^d(\bfm_{n,i})^2\bfV_{n,i} \biggr]
      +(1-\alpha)\biggl[\textstyle\sum\limits_{i=1}^d\bfm_{n,i}\bigl[\textstyle\frac 1M \bigl[\sum_{j=1}^M\cG_1^i(\Theta_{n-1},X_{n,j})\bigr]-\bfm_{n-1,i}\bigr]\bfV_{n,i}\biggr]\\
      &-\displaystyle\frac{1-\alpha}{2}\biggl[\textstyle\sum\limits_{i=1}^d\bigl[\textstyle\frac 1M \bigl[\sum_{j=1}^M\cG_1^i(\Theta_{n-1},X_{n,j})\bigr]-\bfm_{n-1,i}\bigr]^2\bfV_{n,i} \biggr]+\displaystyle\frac{1}{2}\biggl[\textstyle\sum\limits_{i=1}^d(\bfm_{n,i})^2\bfV_{n,i} \biggr]\\
      &-\displaystyle(1-\alpha)\biggl[\textstyle\sum\limits_{i=1}^d\bfm_{n,i}\bigl[\textstyle\frac 1M \bigl[\sum_{j=1}^M\cG_1^i(\Theta_{n-1},X_{n,j})\bigr]-\bfm_{n-1,i}\bigr]\bfV_{n,i} \biggr]
      \\
      &+\displaystyle\frac{(1-\alpha)^2}{2}\biggl[\textstyle\sum\limits_{i=1}^d\bigl[\textstyle\frac 1M \bigl[\sum_{j=1}^M\cG_1^i(\Theta_{n-1},X_{n,j})\bigr]-\bfm_{n-1,i}\bigr]^2\bfV_{n,i}\biggr]\\
      &+ 2\cK\scrc\textstyle\biggl[\sum\limits_{i=1}^d|\bfm_{n,i}|\bfV_{n,i}\biggr]^{1+\tau_2}
+2\cK\scrc\textstyle\biggl[\sum\limits_{i=1}^d|\bfm_{n,i}|\bfV_{n,i}\biggr](1+2\|\Theta_{n-1}\|^{\tau_2}).
      \end{split}
  \end{equation}}
  \argument{\lref{eq11};}{for all $n\in \N\cap[\fn,\infty)$ that
  \begin{equation}\llabel{eq12}
      \begin{split}
          H_n&\leq H_{n-1}+\displaystyle\frac{\cK}{2}\biggl[\textstyle\sum\limits_{i=1}^d (\bfm_{n,i})^2(\bfV_{n,i})^2\biggr]\biggl(1+\biggl[\sum\limits_{i=1}^d|\bfm_{n,i}|\bfV_{n,i}\biggr]^{\tau_1}+2\|\Theta_{n-1}\|^{\tau_1}\biggr)\\
          &-\displaystyle\frac{1}{2}\biggl[\textstyle\sum\limits_{i=1}^d(\bfm_{n,i})^2\bfV_{n,i} \biggr]-\displaystyle\frac{\alpha(1-\alpha)}{2}\biggl[\textstyle\sum\limits_{i=1}^d\bigl[\textstyle\frac 1M \bigl[\sum_{j=1}^M\cG_1^i(\Theta_{n-1},X_{n,j})\bigl]-\bfm_{n-1,i}\bigr]^2\bfV_{n,i} \biggr]\\&+2\cK\scrc\textstyle\bigl[\sum_{i=1}^d|\bfm_{n,i}|\bfV_{n,i}\bigr]^{1+\tau_2}+2\cK\scrc\textstyle\bigl[\sum_{i=1}^d|\bfm_{n,i}|\bfV_{n,i}\bigr](1+2\|\Theta_{n-1}\|^{\tau_2}).
      \end{split}
  \end{equation}}
  \argument{\lref{eq12};\lref{def: V}; the assumption that $0\leq \alpha<1$}{that for all $n\in \N\cap[\fn,\infty)$ it holds that
  \begin{equation}\llabel{eq13}
  \begin{split}
    H_n
      &\leq H_{n-1}+\displaystyle\frac{\cK}{2}\biggl[\textstyle\sum\limits_{i=1}^d (\bfm_{n,i})^2(\bfV_{n,i})^2\biggr]\biggl(1+\biggl[\sum\limits_{i=1}^d|\bfm_{n,i}|\bfV_{n,i}\biggr]^{\tau_1}+2\|\Theta_{n-1}\|^{\tau_1}\biggr)\\
      &-\displaystyle\frac{1}{2}\biggl[\textstyle\sum\limits_{i=1}^d(\bfm_{n,i})^2\bfV_{n,i} \biggr]-\displaystyle\frac{\alpha(1-\alpha)}{2}\biggl[\textstyle\sum\limits_{i=1}^d\bigl[\textstyle\frac 1M \bigl[\sum_{j=1}^M\cG_1^i(\Theta_{n-1},X_{n,j})\bigl]-\bfm_{n-1,i}\bigr]^2\bfV_{n,i} \biggr]\\
&+2\cK\scrc\textstyle\biggl[\sum\limits_{i=1}^d|\bfm_{n,i}|\bfV_{n,i}\biggr]^{1+\tau_2}+2\cK\scrc\textstyle\biggl[\sum\limits_{i=1}^d|\bfm_{n,i}|\bfV_{n,i}\biggr](1+2\|\Theta_{n-1}\|^{\tau_2}).
      \end{split}
  \end{equation}}
   \argument{\lref{def: bfV};the assumption that $0\leq \alpha<1$; the fact that $0<\beta<1$; the assumption that for all $n\in \N$ it holds that $\gamma_n\geq 0$}{that for all $n\in \N$, $i\in \{1,2,\dots,\fd\}$ it holds that 
  \begin{equation}\llabel{arg1}
      \bfV_{n,i}\geq 0.
  \end{equation}}
   \argument{\lref{def: bfV};the assumption that $0\leq \alpha<1$;}{that for all $n\in \N\cap[\max\{N,-[\log_{2}(\alpha+2\mathbbm 1_{\{0\}}(\alpha))]^{-1}\},\infty)$, $i\in \{1,2,\dots,\fd\}$ it holds that
  \begin{equation}\llabel{eq14}
\bfV_{n,i}=\gamma_n\Bigl[\textstyle\varepsilon+\Bigl[\frac{\democrator_{n,i}}{1-\beta^n}\Bigr]^{\nicefrac{1}{2}}\Bigr]^{-1} \Bigl[\frac{1}{1-\alpha^n}\Bigr]\leq 2\gamma_n\varepsilon^{-1}.
  \end{equation}}
  \argument{\lref{def: N};\lref{def: fn};\lref{eq14};\lref{eq13};Cauchy-Schwarz inequality;Young inequality; the fact that $\tau_2\leq \frac 12<1$}{that for all $n\in \N\cap[\fn,\infty)$ it holds that
  \begin{align}
    &H_n
      \leq H_{n-1}+\frac{\cK}{2}\biggl[\textstyle\sum\limits_{i=1}^d(\bfm_{n,i})^2(\bfV_{n,i})^2 \biggr]\biggl(1+\biggl[\sum\limits_{i=1}^d|\bfm_{n,i}|\bfV_{n,i}\biggr]^{\tau_1}+2\|\Theta_{n-1}\|^{\tau_1}\biggr)\notag\\
      &-\displaystyle\frac{1}{2}\biggl[\textstyle\sum\limits_{i=1}^d(\bfm_{n,i})^2\bfV_{n,i} \biggr]-\displaystyle\frac{\alpha(1-\alpha)}{2}\biggl[\textstyle\sum\limits_{i=1}^d\bigl[\textstyle\frac 1M \bigl[\sum_{j=1}^M\cG_1^i(\Theta_{n-1},X_{n,j})\bigl]-\bfm_{n-1,i}\bigr]^2\bfV_{n,i} \biggr]\notag\\
&+2\cK\scrc d^{\tau_2}\textstyle\biggl[\sum\limits_{i=1}^d|\bfm_{n,i}|^{1+\tau_2}(\bfV_{n,i})^{1+\tau_2}\biggr]+\frac{1}{16}\textstyle\biggl[\sum\limits_{i=1}^d|\bfm_{n,i}|^2\bfV_{n,i}\biggr]+\textstyle16\cK^2\scrc^2\biggl[\sum\limits_{i=1}^d(1+2\|\Theta_{n-1}\|^{\tau_2})^2\bfV_{n,i}\biggr]\notag\\
&\leq H_{n-1}+\frac{\cK}{2}\biggl[\textstyle\sum\limits_{i=1}^d(\bfm_{n,i})^2(\bfV_{n,i})^2 \biggr]\biggl(1+\biggl[\sum\limits_{i=1}^d|\bfm_{n,i}|\bfV_{n,i}\biggr]^{\tau_1}+2\|\Theta_{n-1}\|^{\tau_1}\biggr)\notag\\
&-\displaystyle\frac{1}{2}\biggl[\textstyle\sum\limits_{i=1}^d(\bfm_{n,i})^2\bfV_{n,i} \biggr]-\displaystyle\frac{\alpha(1-\alpha)}{2}\biggl[\textstyle\sum\limits_{i=1}^d\bigl[\textstyle\frac 1M \bigl[\sum_{j=1}^M\cG_1^i(\Theta_{n-1},X_{n,j})\bigl]-\bfm_{n-1,i}\bigr]^2\bfV_{n,i} \biggr]\notag\\
&+\cK\scrc d^{\tau_2}(1+\tau_2)\textstyle\biggl[\sum\limits_{i=1}^d|\bfm_{n,i}|^{2}(\bfV_{n,i})^{1+\tau_2}\biggr]+\cK\scrc d^{\tau_2}(1-\tau_2)\textstyle\biggl[\sum\limits_{i=1}^d(\bfV_{n,i})^{1+\tau_2}\biggr]\notag\\
&+\frac{1}{16}\textstyle\biggl[\sum\limits_{i=1}^d|\bfm_{n,i}|^2\bfV_{n,i}\biggr]+\textstyle32\cK^2\scrc^2\biggl[\sum\limits_{i=1}^d(1+4\|\Theta_{n-1}\|^{2\tau_2})\bfV_{n,i}\biggr]\notag\\
&\leq H_{n-1}+\frac{\cK}{2}\biggl[\textstyle\sum\limits_{i=1}^d(\bfm_{n,i})^2(\bfV_{n,i} )^2\biggr]\biggl(1+\biggl[\sum\limits_{i=1}^d|\bfm_{n,i}|\bfV_{n,i}\biggr]^{\tau_1}+2\|\Theta_{n-1}\|^{\tau_1}\biggr)\notag\\
&-\displaystyle\frac{1}{2}\biggl[\textstyle\sum\limits_{i=1}^d(\bfm_{n,i})^2\bfV_{n,i} \biggr]-\displaystyle\frac{\alpha(1-\alpha)}{2}\biggl[\textstyle\sum\limits_{i=1}^d\bigl[\textstyle\frac 1M \bigl[\sum_{j=1}^M\cG_1^i(\Theta_{n-1},X_{n,j})\bigl]-\bfm_{n-1,i}\bigr]^2\bfV_{n,i} \biggr]\notag\\
&+\frac{1}{16}\textstyle\biggl[\sum\limits_{i=1}^d|\bfm_{n,i}|^{2}\bfV_{n,i}\biggr]+\textstyle\biggl[\sum\limits_{i=1}^d\bfV_{n,i}\biggr]+\displaystyle\frac{1}{16}\textstyle\biggl[\sum\limits_{i=1}^d|\bfm_{n,i}|^2\bfV_{n,i}\biggr]\notag\\
&+\textstyle32\cK^2\scrc^2\biggl[\sum\limits_{i=1}^d(1+4\|\Theta_{n-1}\|^{2\tau_2})\bfV_{n,i}\biggr]\llabel{eq14'}.
  \end{align}}
  \argument{\lref{eq14'};}{for all $n\in \N\cap[\fn,\infty)$ that
  \begin{equation}\llabel{eqmain'}
  \begin{split}
     H_n-H_{n-1}
      &\leq \frac{\cK}{2}\biggl[\textstyle\sum\limits_{i=1}^d(\bfm_{n,i})^2(\bfV_{n,i})^2\biggr]\biggl(1+\biggl[\sum\limits_{i=1}^d|\bfm_{n,i}|\bfV_{n,i}\biggr]^{\tau_1}+2\|\Theta_{n-1}\|^{\tau_1}\biggr)\\
      & -\displaystyle\frac{3}{8}\biggl[\textstyle\sum\limits_{i=1}^d(\bfm_{n,i})^2\bfV_{n,i} \biggr]+(1+32\cK^2\scrc^2+128\cK^2\scrc^2\|\Theta_{n-1}\|^{2\tau_2})\textstyle\biggl[\sum\limits_{i=1}^d\bfV_{n,i}\biggr]\\
      &-\displaystyle\frac{\alpha(1-\alpha)}{2}\biggl[\textstyle\sum\limits_{i=1}^d\bigl[\textstyle\frac 1M \bigl[\sum_{j=1}^M\cG_1^i(\Theta_{n-1},X_{n,j})\bigl]-\bfm_{n-1,i}\bigr]^2\bfV_{n,i} \biggr].
      \end{split}
  \end{equation}}
  \argument{\lref{eqmain'};}{for all $n\in \N\cap[\fn,\infty)$ that
  \begin{equation}\llabel{eqmain''.1}
       \begin{split}
     H_n-H_{n-1}
     &\leq \frac{\cK}{2}\biggl[\textstyle\sum\limits_{i=1}^d(\bfm_{n,i})^2(\bfV_{n,i})^2 \biggr]\biggl(1+\biggl[\sum\limits_{i=1}^d|\bfm_{n,i}|\bfV_{n,i}\biggr]^{\tau_1}+2\|\Theta_{n-1}\|^{\tau_1}\biggr)\\
      &-\displaystyle\frac{3}{8}\biggl[\textstyle\sum\limits_{i=1}^d(\bfm_{n,i})^2\bfV_{n,i} \biggr]+ (1+32\cK^2\scrc^2+128\cK^2\scrc^2\|\Theta_{n-1}\|^{2\tau_2})\textstyle\biggl[\sum\limits_{i=1}^d\bfV_{n,i}\biggr] \\
      &-\displaystyle\frac{\alpha(1-\alpha)}{2}\biggl[\textstyle\sum\limits_{i=1}^d\bigl[\textstyle\frac 1M \bigl[\sum_{j=1}^M\cG_1^i(\Theta_{n-1},X_{n,j})\bigl]-\bfm_{n-1,i}\bigr]^2\bfV_{n,i} \biggr].
      \end{split}
  \end{equation}}
  \argument{\lref{eqmain''.1};}{for all $n\in \N\cap[\fn,\infty)$ that
  \begin{equation}\llabel{eqmain''}
  \begin{split}
    H_n-H_{n-1}&\leq \frac{\cK}{2}\biggl[\textstyle\sum\limits_{i=1}^d(\bfm_{n,i})^2(\bfV_{n,i})^2\biggr]\biggl(1+\biggl[\sum\limits_{i=1}^d|\bfm_{n,i}|\bfV_{n,i}\biggr]^{\tau_1}+2\|\Theta_{n-1}\|^{\tau_1}\biggr)\\
    &-\displaystyle\frac{1}{4}\biggl[\textstyle\sum\limits_{i=1}^d(\bfm_{n,i})^2\bfV_{n,i} \biggr]
    + \textstyle  (1+32\cK^2\scrc^2+128\cK^2\scrc^2\|\Theta_{n-1}\|^{2\tau_2})\textstyle\biggl[\sum\limits_{i=1}^d\bfV_{n,i}\biggr] \\
      &-\textstyle\bigl[\max\{\frac{2\alpha}{1-\alpha},8\}\bigr]^{-1}\biggl[\textstyle\sum\limits_{i=1}^d(\bfm_{n,i})^2\bfV_{n,i} \biggr]\\
      &-\textstyle \bigl[\max\{\frac{2\alpha}{1-\alpha},8\}\bigr]^{-1}\biggl[\textstyle\sum\limits_{i=1}^d\bigl[\textstyle\frac {\alpha}{M} \bigl[\sum_{j=1}^M\cG_1^i(\Theta_{n-1},X_{n,j})\bigl]-\alpha\bfm_{n-1,i}\bigr]^2\bfV_{n,i} \biggr]\textstyle.
      \end{split}
  \end{equation}}
  \argument{\lref{eqmain''};\lref{def: bfm};\lref{def: delta}; the fact that for all $x,y\in \R$ it holds that $x^2+y^2\geq \frac 12 (x+y)^2$}{that for all $n\in \N\cap[\fn,\infty)$ it holds that
  \begin{equation}\llabel{eqmain'''}
  \begin{split}
     H_n-H_{n-1}
     &\leq \frac{\cK}{2}\biggl[\textstyle\sum\limits_{i=1}^d(\bfm_{n,i})^2(\bfV_{n,i})^2\biggr]\biggl(1+\biggl[\sum\limits_{i=1}^d|\bfm_{n,i}|\bfV_{n,i}\biggr]^{\tau_1}+2\|\Theta_{n-1}\|^{\tau_1}\biggr)\\
    &-\displaystyle\frac{1}{4}\biggl[\textstyle\sum\limits_{i=1}^d(\bfm_{n,i})^2\bfV_{n,i} \biggr]+(1+32\cK^2\scrc^2+128\cK^2\scrc^2\|\Theta_{n-1}\|^{2\tau_2})\textstyle\biggl[\sum\limits_{i=1}^d\bfV_{n,i}\biggr] \\
      &\textstyle -\textstyle \frac 12\bigl[\max\{\frac{2\alpha}{1-\alpha},8\}\bigr]^{-1}\biggl[\textstyle\sum\limits_{i=1}^d\bigl[\textstyle\frac {\alpha}{M} \bigl[\sum_{j=1}^M\cG_1^i(\Theta_{n-1},X_{n,j})\bigl]+\bfm_{n,i}-\alpha\bfm_{n-1,i}\bigr]^2\bfV_{n,i} \biggr]\\
&=\frac{\cK}{2}\biggl[\textstyle\sum\limits_{i=1}^d(\bfm_{n,i})^2(\bfV_{n,i})^2\biggr]\biggl(1+\biggl[\sum\limits_{i=1}^d|\bfm_{n,i}|\bfV_{n,i}\biggr]^{\tau_1}+2\|\Theta_{n-1}\|^{\tau_1}\biggr)\\
    &-\displaystyle\frac{1}{4}\biggl[\textstyle\sum\limits_{i=1}^d(\bfm_{n,i})^2\bfV_{n,i} \biggr]+(1+32\cK^2\scrc^2+128\cK^2\scrc^2\|\Theta_{n-1}\|^{2\tau_2})\textstyle\biggl[\sum\limits_{i=1}^d\bfV_{n,i}\biggr] \\
      &-\textstyle \frac 12\bigl[\max\{\frac{2\alpha}{1-\alpha},8\}\bigr]^{-1}\biggl[\textstyle\sum\limits_{i=1}^d\bigl[\textstyle\frac {1}{M} \bigl[\sum_{j=1}^M\cG_1^i(\Theta_{n-1},X_{n,j})\bigl]\bigr]^2\bfV_{n,i} \biggr]\\ &=\frac{\cK}{2}\biggl[\textstyle\sum\limits_{i=1}^d(\bfm_{n,i})^2(\bfV_{n,i})^2\biggr]\biggl(1+\biggl[\sum\limits_{i=1}^d|\bfm_{n,i}|\bfV_{n,i}\biggr]^{\tau_1}+2\|\Theta_{n-1}\|^{\tau_1}\biggr)\\
    &-\displaystyle\frac{1}{4}\biggl[\textstyle\sum\limits_{i=1}^d(\bfm_{n,i})^2\bfV_{n,i} \biggr]+(1+32\cK^2\scrc^2+128\cK^2\scrc^2\|\Theta_{n-1}\|^{2\tau_2})\textstyle\biggl[\sum\limits_{i=1}^d\bfV_{n,i}\biggr]\\
      &-\displaystyle \frac {1}{4\chi}\biggl[\textstyle\sum\limits_{i=1}^d\bigl[\textstyle\frac {1}{M} \bigl[\sum_{j=1}^M\cG_1^i(\Theta_{n-1},X_{n,j})\bigl]\bigr]^2\bfV_{n,i} \biggr].
      \end{split}
  \end{equation}}
  \argument{\lref{eqmain'''};\lref{def: H};\lref{eq14}}{for all $n\in \N\cap[\fn,\infty)$ that
  \begin{equation}\llabel{eqmain''''}
      \begin{split}
      &\smalll(\Theta_n,\bfx)-\smalll(\vartheta,\bfx)+\frac{1}{2(1-\alpha)}\textstyle\biggl[\sum\limits_{i=1}^d(\bfm_{n,i})^2\bfV_{n,i}\biggr]\\
      &-\biggl[\smalll(\Theta_{n-1},\bfx)-\smalll(\vartheta,\bfx)+\frac{1}{2(1-\alpha)}\textstyle\biggl[\sum\limits_{i=1}^d(\bfm_{n-1,i})^2\bfV_{n-1,i}\biggr]\biggr]\\
     &\leq \frac{\cK}{2}\biggl[\textstyle\sum\limits_{i=1}^d(\bfm_{n,i})^2(\bfV_{n,i})^2\biggr]\biggl(1+\biggl[\sum\limits_{i=1}^d|\bfm_{n,i}|\bfV_{n,i}\biggr]^{\tau_1}+2\|\Theta_{n-1}\|^{\tau_1}\biggr)\\
    &-\displaystyle\frac{1}{4}\biggl[\textstyle\sum\limits_{i=1}^d(\bfm_{n,i})^2\bfV_{n,i} \biggr]+(1+32\cK^2\scrc^2+128\cK^2\scrc^2\|\Theta_{n-1}\|^{2\tau_2})\textstyle\biggl[\sum\limits_{i=1}^d\bfV_{n,i}\biggr] \\
     &-\displaystyle \frac {1}{4\chi}\biggl[\textstyle\sum\limits_{i=1}^d\bigl[\textstyle\frac {1}{M} \bigl[\sum_{j=1}^M\cG_1^i(\Theta_{n-1},X_{n,j})\bigl]\bigr]^2\bfV_{n,i} \biggr]\\
     &\leq \frac{\cK}{2}\biggl[\textstyle\sum\limits_{i=1}^d(\bfm_{n,i})^2(\bfV_{n,i})^2\biggr]\biggl(1+\biggl[\sum\limits_{i=1}^d|\bfm_{n,i}|\bfV_{n,i}\biggr]^{\tau_1}+2\|\Theta_{n-1}\|^{\tau_1}\biggr)\\
    &-\displaystyle\frac{1}{4}\biggl[\textstyle\sum\limits_{i=1}^d(\bfm_{n,i})^2\bfV_{n,i} \biggr]+2(1+32\cK^2\scrc^2+128\cK^2\scrc^2\|\Theta_{n-1}\|^{2\tau_2})d\gamma_n\varepsilon^{-1}\\
     &-\displaystyle \frac {1}{4\chi}\biggl[\textstyle\sum\limits_{i=1}^d\bigl[\textstyle\frac {1}{M} \bigl[\sum_{j=1}^M\cG_1^i(\Theta_{n-1},X_{n,j})\bigl]\bigr]^2\bfV_{n,i} \biggr].
      \end{split}
  \end{equation}}
       \argument{\lref{def: bfm};\lref{def: V};\lref{def: delta};\cref{lem: bounded increament Adam}}{that for all $n\in \N$, $i\in \{1,2,\dots,d\}$ it holds that
\begin{equation}\llabel{eqtt1}
\begin{split}
    &\textstyle|\bfm_{n,i}|\bfV_{n,i}=\frac{ \gamma_n|\bfm_{n,i}|}{1-\alpha^n} \Bigl[\varepsilon+\bigl[\frac{\democrator_{n,i}}{1-\beta^n}\bigr]^{\nicefrac{1}{2}}\Bigr]^{-1}\leq \frac{\gamma_n }{\sqrt{(1-\alpha^2\beta^{-1})(1-\beta)}}+\frac{\gamma_n \bfM}{\varepsilon(1-\alpha)}=\gamma_n\eta d^{-1}.
    \end{split}
\end{equation}}
\argument{\lref{eqtt1};\lref{def: N};\lref{eq14};Young inequality; the fact that for all $n\in \N$ it holds that $\gamma_n\leq \gamma_1$}{that for all $n\in \N\cap[\fn,\infty)$ it holds that
\begin{equation}\llabel{eqtt2}
    \begin{split}
        &\frac{\cK}{2}\biggl[\textstyle\sum\limits_{i=1}^d(\bfm_{n,i})^2(\bfV_{n,i})^2\biggr]\biggl(1+\biggl[\sum\limits_{i=1}^d|\bfm_{n,i}|\bfV_{n,i}\biggr]^{\tau_1}+2\|\Theta_{n-1}\|^{\tau_1}\biggr)\\
        &\leq \frac{\cK}{2}\biggl[\textstyle\sum\limits_{i=1}^d(\bfm_{n,i})^2(\bfV_{n,i})^2\biggr]\biggl(1+\biggl[\sum\limits_{i=1}^d|\bfm_{n,i}|\bfV_{n,i}\biggr]^{\tau_1}+2(1-\tau_1)+2\tau_1\|\Theta_{n-1}\|\biggr)\\
        &\leq  \frac{\cK}{2}\biggl[\textstyle\sum\limits_{i=1}^d(\bfm_{n,i})^2(\bfV_{n,i})^2\biggr]\bigl(1+(\gamma_n\eta )^{\tau_1}+2(1-\tau_1)+2\tau_1\|\Theta_{n-1}\|\bigr)\\
        &\textstyle\leq  \frac{\cK}{2}\biggl[\textstyle\sum\limits_{i=1}^d(\bfm_{n,i})^2(\bfV_{n,i})^2\biggr]\bigl(3+(\gamma_1\eta )^{\tau_1}\bigr)+\tau_1\cK\biggl[\textstyle\sum\limits_{i=1}^d(\bfm_{n,i})^2(\bfV_{n,i})^2\biggr]\|\Theta_{n-1}\|\\
        &\leq  \biggl[\textstyle\sum\limits_{i=1}^d(\bfm_{n,i})^2\bfV_{n,i}\biggr]2\gamma_n\varepsilon^{-1}\bigl(\frac{\cK}{2}\bigr)\bigl(3+(\gamma_1\eta )^{\tau_1}\bigr)+\tau_1\cK d(\gamma_n\eta d^{-1})^2\|\Theta_{n-1}\|\\
        &\leq \frac{1}{8}\biggl[\textstyle\sum\limits_{i=1}^d(\bfm_{n,i})^2\bfV_{n,i}\biggr]+\tau_1\cK d^{-1}\eta^2(\gamma_n)^2\|\Theta_{n-1}\|.
        \end{split}
\end{equation}}
\argument{\lref{eqtt2};\lref{eqmain''''}}{for all $n\in \N\cap[\fn,\infty)$ that
\begin{equation}\llabel{eqmain'''''}
      \begin{split}
      &\smalll(\Theta_n,\bfx)-\smalll(\vartheta,\bfx)+\frac{1}{2(1-\alpha)}\textstyle\biggl[\sum\limits_{i=1}^d(\bfm_{n,i})^2\bfV_{n,i}\biggr]\\
      &-\biggl[\smalll(\Theta_{n-1},\bfx)-\smalll(\vartheta,\bfx)+\frac{1}{2(1-\alpha)}\textstyle\biggl[\sum\limits_{i=1}^d(\bfm_{n-1,i})^2\bfV_{n-1,i}\biggr]\biggr]\\
&\leq\tau_1\cK d^{-1}\eta^2(\gamma_n)^2\|\Theta_{n-1}\|-\displaystyle \frac {1}{4\chi}\biggl[\textstyle\sum\limits_{i=1}^d\bigl[\textstyle\frac {1}{M} \bigl[\sum_{j=1}^M\cG_1^i(\Theta_{n-1},X_{n,j})\bigl]\bigr]^2\bfV_{n,i} \biggr]\\
&+2(1+32\cK^2\scrc^2+128\cK^2\scrc^2\|\Theta_{n-1}\|^{2\tau_2})d\gamma_n\varepsilon^{-1}-\displaystyle\frac{1}{8}\biggl[\textstyle\sum\limits_{i=1}^d(\bfm_{n,i})^2\bfV_{n,i} \biggr].
      \end{split}
\end{equation}}
       \argument{\lref{eqp2'}; the fact that $\bfx\in \setX$}{that for all $\theta_1,\theta_2\in \R^d$ it holds that
    \begin{equation}\llabel{eq15'}
    \begin{split}
       &\spro{\cG_1(\theta_2,\bfx),\theta_1-\theta_2}+\frac{\cK}{2}\|\theta_1-\theta_2\|^2(1+\|\theta_1-\theta_2\|^{\tau_1}+2\|\theta_2\|^{\tau_1})\geq  \smalll(\theta_1,\bfx)-\smalll(\theta_2,\bfx)\\
       &\geq \spro{\cG_1(\theta_2,\bfx),\theta_1-\theta_2}+\frac{\kappa}{2}\|\theta_1-\theta_2\|^2.
       \end{split}
    \end{equation}}
    \argument{\lref{eq15'};\lref{def: vartheta}}{that for all $\theta\in \R^\fd$ it holds that
    \begin{equation}\llabel{material 1}
       \frac{\cK}{2}\|\theta-\vartheta\|^2(1+\|\theta-\vartheta\|^{\tau_1}+2\|\vartheta\|^{\tau_1}) \geq \smalll(\theta,\bfx)-\smalll(\vartheta,\bfx)\geq \frac{\kappa}{2}\|\theta-\vartheta\|^2.
    \end{equation}}
   \argument{\lref{def: vartheta};\lref{eqp1.1};the Cauchy-Schwarz inequality}{that for all $\theta\in \R^\fd$, $m\in \N$, $x_1,x_2,\dots,x_m\in K$ it holds that
    \begin{equation}\llabel{eq15.1}
    \begin{split}
        \kappa\|\theta-\vartheta\|^2&\leq \bigl\langle\theta-\vartheta, \nabla_\theta\bigl(\textstyle\frac 1m \textstyle\sum_{j=1}^m \smalll(\theta,x_j)\bigr)- \nabla_\theta\bigl(\textstyle\frac 1m\textstyle\sum_{j=1}^m \smalll(\vartheta,x_j)\bigr)\bigr\rangle\\
        &\leq \|\theta-\vartheta\|\bigl\|\nabla_\theta\bigl(\textstyle\frac 1m\textstyle\sum_{j=1}^m \smalll(\theta,x_j)-\nabla_\theta\bigl(\textstyle\frac 1m\textstyle\sum_{j=1}^m \smalll(\vartheta,x_j)\bigr)\bigr\|.
        \end{split}
    \end{equation}}
      \argument{\lref{eq15.1};\lref{assumption}}{for all $\theta\in \R^\fd$, $m\in \N$, $x_1,x_2,\dots,x_m\in \setX$ that
    \begin{equation}\llabel{eq15.2}
    \begin{split}
        &\kappa\|\theta-\vartheta\|\leq \bigl\|\bigl(\textstyle\frac 1m\textstyle\sum_{j=1}^m (\nabla_\theta\smalll)(\theta,x_j)\bigr)-\bigl(\textstyle\frac 1m\textstyle\sum_{j=1}^m (\nabla_\theta\smalll)(\vartheta,x_j)\bigr)\bigr\|\\
        &\textstyle\leq \frac 1m \bigl[\sum_{j=1}^m\|(\nabla_\theta\smalll)(\theta,x_j)-(\nabla_\theta\smalll)(\vartheta,x_j)\|\bigr]\leq \frac 1m\bigl[\sum_{j=1}^m\cK\|\theta-\vartheta\|(1+\|\theta\|^{\tau_1}+\|\vartheta\|^{\tau_1})\bigr]\\
        &= \cK\|\theta-\vartheta\|(1+\|\theta\|^{\tau_1}+\|\vartheta\|^{\tau_1}).
        \end{split}
    \end{equation}}
    \argument{\lref{eq15.2};the triangle inequality}{that for all $\theta\in \R^\fd$, $m\in \N$, $x_1,x_2,\dots,x_m\in K$ it holds that
    \begin{equation}\llabel{material 2'}
    \begin{split}
        &\textstyle\kappa\|\theta-\vartheta\|-\sup_{y\in \setX}\|(\nabla_\theta\smalll)(\vartheta,y)\|\leq \kappa\|\theta-\vartheta\|-\frac 1m \bigl\|\sum_{j=1}^m (\nabla_\theta\smalll)(\vartheta,x_j)\bigr\|\\
        &\textstyle\leq  \frac 1m \bigl\|\sum_{j=1}^m(\nabla_\theta\smalll)(\theta,x_j)\bigr\|\leq \cK\|\theta-\vartheta\|(1+\|\theta\|^{\tau_1}+\|\vartheta\|^{\tau_1})+\frac 1m \bigl\|\sum_{j=1}^m(\nabla_\theta\smalll)(\vartheta,x_j)\bigr\|\\
        &\textstyle\leq \cK\|\theta-\vartheta\|(1+\|\theta\|^{\tau_1}+\|\vartheta\|^{\tau_1})+\sup_{y\in \setX}\|(\nabla_\theta\smalll)(\vartheta,y)\|.
        \end{split}
    \end{equation}}
    \argument{\lref{material 2'};\lref{def: vartheta};the fact that for all $x,y\in \R^d$ it holds that $\|x+y\|^{\tau_1}\leq \|x\|^{\tau_1}+\|y\|^{\tau_1}$}{that for all $\theta\in \R^\fd$, $m\in \N$, $x_1,x_2,\dots,x_m\in K$ it holds that
    \begin{equation}\llabel{material 2}
    \begin{split}
        &\textstyle\kappa\|\theta-\vartheta\|-\rho\leq   \frac 1m \bigl\|\sum_{j=1}^m(\nabla_\theta\smalll)(\theta,x_j)\bigr\|\leq \cK\|\theta-\vartheta\|(1+\|\theta\|^{\tau_1}+\|\vartheta\|^{\tau_1})+\rho\\
        &\leq \cK\|\theta-\vartheta\|(1+\|\theta-\vartheta\|^{\tau_1}+2\|\vartheta\|^{\tau_1})+\rho.
        \end{split}
    \end{equation}}
     \argument{\lref{def: bfm};\lref{def: V};\lref{def: cG};\unskip, \eg, \cite[Lemma 2.7]{DeRoArAd2025}}{that for all $n\in \N$, $i\in \{1,2,\dots,d\}$ it holds that
    \begin{equation}\llabel{eq18.25}
        \bfm_{n,i}=\textstyle(1-\alpha)\biggl[\sum\limits_{k=1}^n\alpha^{n-k}\textstyle \frac 1M \bigl[\textstyle\sum_{j=1}^M\cG_1^i(\Theta_{k-1},X_{k,j})\bigr]\biggr]+\alpha^n\bfm_{0,i}
        \end{equation}
        \begin{equation}\llabel{eq18.5}
            \text{and}\qquad  \bbV_{n,i}=(1-\beta)\bigl[\textstyle\sum_{k=1}^n\beta^{n-k}\bigl[ \frac 1M \bigl(\textstyle\sum_{j=1}^M\cG_1^i(\Theta_{k-1},X_{k,j})\bigr)\bigr]^2 \textstyle \bigr]+\beta^n\bbV_{0,i}.
    \end{equation}}
    \argument{\lref{eq18.5};\lref{def: bfm};\lref{def: delta};\lref{eq: x2.5};the fact that for all $n\in \N$ it holds that $\gamma_n\leq \gamma_1$;the fact that $0<\beta<1$;\cref{lem: bounded increament Adam}}{that for all $n\in \N\cap[1,\fn]$ it holds that
    \begin{equation}\llabel{eq16.1}
    \begin{split}
        &\textstyle\|\Theta_n-\Theta_{n-1}\|\leq  \sum\limits_{i=1}^d\gamma_n\bigl[\frac{|\bfm_{n,i}|}{1-\alpha^n}\bigr] \Bigl[\varepsilon+\bigl[\frac{\democrator_{n,i}}{1-\beta^n}\bigr]^{\nicefrac{1}{2}}\Bigr]^{-1}\\
        &\textstyle \leq \sum\limits_{i=1}^d\biggl[\frac{\gamma_n \bigl[\sum_{k=0}^{n-1}\beta^k\bigr]^{1/2}}{\sqrt{1-\alpha^2\beta^{-1}}}+\frac{\gamma_n |\bfm_{0,i}|}{\varepsilon(1-\alpha)}\biggr]\leq \gamma_n d \biggl[\frac{ \bigl[\sum_{k=0}^{n-1}\beta^k\bigr]^{1/2}}{\sqrt{1-\alpha^2\beta^{-1}}}+\frac{\bfM}{\varepsilon(1-\alpha)}\biggr]\\
        &\leq \gamma_n d \biggl[\frac{ n^{1/2}}{\sqrt{1-\alpha^2\beta^{-1}}}+\frac{\bfM}{\varepsilon(1-\alpha)}\biggr]\leq \gamma_n d \biggl[\frac{ \fn^{1/2}}{\sqrt{1-\alpha^2\beta^{-1}}}+\frac{\bfM}{\varepsilon(1-\alpha)}\biggr]=\gamma_n \delta\leq \gamma_1 \delta.
        \end{split}
    \end{equation}}
    \argument{\lref{eq16.1};}
    {that for all $n\in \N\cap[0,\fn]$ it holds that
    \begin{equation}\llabel{eq16}
        \|\Theta_n-\vartheta\|\leq \|\Theta_n-\Theta_0\|+\|\Theta_0-\vartheta\|\leq \fn\gamma_1\delta+\|\Theta_0-\vartheta\|.
    \end{equation}}
    \argument{\lref{eq16};\lref{material 1};}{for all $n\in \N_0\cap[0,\fn]$ that
    \begin{equation}\llabel{eq17}
    \begin{split}
        &\smalll(\Theta_n,\bfx)-\smalll(\vartheta,\bfx)\leq    \frac{\cK}{2}\|\Theta_n-\vartheta\|^2(1+\|\Theta_n-\vartheta\|^{\tau_1}+2\|\vartheta\|^{\tau_1}) \\
        &\leq \frac{\cK}{2}( \fn\gamma_1\delta+\|\Theta_0-\vartheta\|)^2(1+(\fn\gamma_1\delta+\|\Theta_0-\vartheta\|)^{\tau_1}+2\|\vartheta\|^{\tau_1}).
        \end{split}
    \end{equation}}
    \argument{\lref{def: cG};\lref{material 2};\lref{eq16};}{that for all $n\in \N_0\cap[0,\fn]$, $i\in \{1,2,\dots,d\}$, $x\in \setX$ it holds that
    \begin{equation}\llabel{eq18}
    \begin{split}
      & \textstyle|\cG_1^i(\Theta_n,x)|\leq \|\cG_1(\Theta_n,x)\|
      \leq \cK\|\Theta_n-\vartheta\|(1+\|\Theta_n-\vartheta\|^{\tau_1}+2\|\vartheta\|^{\tau_1})+\rho\\
      &\leq \cK(\fn\gamma_1\delta+\|\Theta_0-\vartheta\|)(1+(\fn\gamma_1\delta+\|\Theta_0-\vartheta\|)^{\tau_1}+2\|\vartheta\|^{\tau_1})+\rho.
       \end{split}
    \end{equation}}
    \argument{\lref{eq18};\lref{def: bfm};\lref{eq18.25};the assumption that for all $n,m\in \N$ it holds that $X_{n,m}\in \setX$}{that for all $n\in \N_0\cap[0,\fn]$, $i\in \{1,2,\dots,d\}$ it holds that
    \begin{equation}\llabel{eq19}
        |\bfm_{n,i}|\leq\cK(\fn\gamma_1\delta+\|\Theta_0-\vartheta\|)(1+(\fn\gamma_1\delta+\|\Theta_0-\vartheta\|)^{\tau_1}+2\|\vartheta\|^{\tau_1})+\rho+\bfM.
    \end{equation}}
    \argument{\lref{def: Theta};\lref{def: bfV};\lref{eq16.1};}{that for all $n\in \N\cap[1,\fn]$, $i\in \{1,2,\dots,d\}$ it holds that
    \begin{equation}\llabel{eq20}
        |\bfm_{n,i}\bfV_{n,i}|=|\Theta_n^i-\Theta_{n-1}^i|\leq \|\Theta_n-\Theta_{n-1}\|\leq \gamma_n\delta\leq  \gamma_1\delta.
    \end{equation}}
    \argument{\lref{eq20};\lref{def: A1};\lref{def: bfA};\lref{eq17};\lref{eq19};}{for all $n\in \N_0\cap[1,\fn]$ that
    \begin{equation}\llabel{eqq1}
    \begin{split}
        &\smalll(\Theta_n,\bfx)-\smalll(\vartheta,\bfx)+\frac{1}{2(1-\alpha)}\textstyle\biggl[\sum\limits_{i=1}^d(\bfm_{n,i})^2\bfV_{n,i}\biggr]\\&\leq \frac{\cK}{2}( \fn\gamma_1\delta+\|\Theta_0-\vartheta\|)^2(1+(\fn\gamma_1\delta+\|\Theta_0-\vartheta\|)^{\tau_1}+2\|\vartheta\|^{\tau_1})\\
        &+\frac{d}{2(1-\alpha)} \bigl[ \cK(\fn\gamma_1\delta+\|\Theta_0-\vartheta\|)(1+(\fn\gamma_1\delta+\|\Theta_0-\vartheta\|)^{\tau_1}+2\|\vartheta\|^{\tau_1})+\rho+\bfM\bigr]\gamma_1\delta\\
        &=\cA_1\leq\bfA.
        \end{split}
    \end{equation}}
    In the following we prove that for all $n\in \N\cap[\fn,\infty)$ it holds that
    \begin{equation}\llabel{need to prove}
        \smalll(\Theta_n,\bfx)-\smalll(\vartheta,\bfx)+\frac{1}{2(1-\alpha)}\textstyle\biggl[\sum\limits_{i=1}^d(\bfm_{n,i})^2\bfV_{n,i}\biggr]\leq \bfA+\bfA\sum_{k=1}^n\tau_1\eta^2(\gamma_k)^2.
    \end{equation}
    We prove \lref{need to prove} by induction on $n\in \N_0\cap[\fn,\infty)$. Note that \lref{eqq1} establishes \lref{need to prove} in the base case $n=\fn$. For the induction step we assume that there exists $n\in \N_0\cap[\fn+1,\infty)$ which satisfies that for all $k\in \{\fn,\fn+1,,\dots,n-1\}$ it holds that
\begin{equation}\llabel{induction: assume}
    \smalll(\Theta_k,\bfx)-\smalll(\vartheta,\bfx)+\frac{1}{2(1-\alpha)}\textstyle\biggl[\sum\limits_{i=1}^d(\bfm_{k,i})^2\bfV_{k,i}\biggr]\leq \bfA+\bfA\sum_{h=1}^k\tau_1\eta^2(\gamma_h)^2.
\end{equation}
\startnewargseq
\argument{\lref{def: eta};\lref{arg1};\lref{eqq1};\lref{induction: assume}}{that for all $k\in \{1,2,\dots,n-1\}$ it holds that
\begin{equation}\llabel{induction: eq1.1}
    \smalll(\Theta_k,\bfx)-\smalll(\vartheta,\bfx)\leq \smalll(\Theta_k,\bfx)-\smalll(\vartheta,\bfx)+\frac{1}{2(1-\alpha)}\textstyle\biggl[\sum\limits_{i=1}^d(\bfm_{k,i})^2\bfV_{k,i}\biggr]\leq \bfA+\bfA\Gamma.
\end{equation}}
\argument{\lref{def: A1};\lref{def: bfA};\lref{eq17};}{that
    \begin{equation}\llabel{induction: eq1.2}
    \begin{split}
        \smalll(\Theta_0,\bfx)-\smalll(\vartheta,\bfx)&\leq \frac{\cK}{2}( \fn\gamma_1\delta+\|\Theta_0-\vartheta\|)^2(1+(\fn\gamma_1\delta+\|\Theta_0-\vartheta\|)^{\tau_1}+2\|\vartheta\|^{\tau_1})\\
        &\leq \cA_1\leq\bfA.
        \end{split}
    \end{equation}}
    \argument{\lref{induction: eq1.2};\lref{induction: eq1.1};}{for all $k\in \{0,1,\dots,n-1\}$ that
\begin{equation}\llabel{induction: eq1}
    \smalll(\Theta_k,\bfx)-\smalll(\vartheta,\bfx)\leq \bfA+\bfA\Gamma.
\end{equation}}
\argument{\lref{induction: eq1};\lref{material 1}}{for all $k\in \{0,1,\dots,n-1\}$ that
\begin{equation}\llabel{induction: eq2}
    \|\Theta_k-\vartheta\|^2\leq \frac{2}{\kappa}[\smalll(\Theta_k,\bfx)-\smalll(\vartheta,\bfx)]\leq \frac{2\bfA(1+\Gamma)}{\kappa}.
\end{equation}}
\argument{\lref{induction: eq2};\lref{def: A2};the triangle inequality}{that for all $k\in \{0,1,\dots,n-1\}$ it holds that
\begin{equation}\llabel{eqtt3}
\begin{split}
  &  
\textstyle\|\Theta_k\|\leq \|\Theta_k-\vartheta\|+\|\vartheta\|\leq \sqrt{2\bfA(1+\Gamma)\kappa^{-1}}+\|\vartheta\|\\
&\leq \sqrt{2\bfA(1+\Gamma)\kappa^{-1}}+\sqrt{2\bfA(1+\Gamma)\kappa^{-1}}
    =\textstyle 2\sqrt{2\bfA(1+\Gamma)\kappa^{-1}}\leq \cK^{-1}d\bfA.
    \end{split}
\end{equation}}
\argument{\lref{eqtt3};\lref{def: A2};\lref{eqmain'''''};the fact that $(2\sqrt{2})^{2\tau_2}\leq 8$}{for all $n\in \N\cap[\fn,\infty)$ that
\begin{equation}\llabel{eqmain'6}
      \begin{split}
      &\smalll(\Theta_n,\bfx)-\smalll(\vartheta,\bfx)+\frac{1}{2(1-\alpha)}\textstyle\biggl[\sum\limits_{i=1}^d(\bfm_{n,i})^2\bfV_{n,i}\biggr]\\
      &-\biggl[\smalll(\Theta_{n-1},\bfx)-\smalll(\vartheta,\bfx)+\frac{1}{2(1-\alpha)}\textstyle\biggl[\sum\limits_{i=1}^d(\bfm_{n-1,i})^2\bfV_{n-1,i}\biggr]\biggr]\\
&\leq\bfA\tau_1\eta^2(\gamma_n)^2-\displaystyle \frac {1}{4\chi}\biggl[\textstyle\sum\limits_{i=1}^d\bigl[\textstyle\frac {1}{M} \bigl[\sum_{j=1}^M\cG_1^i(\Theta_{n-1},X_{n,j})\bigl]\bigr]^2\bfV_{n,i} \biggr]\\
&+2(1+32\cK^2\scrc^2+1024\cK^2\scrc^2(\bfA\kappa^{-1}(1+\Gamma))^{\tau_2})\textstyle d\gamma_n\varepsilon^{-1}-\displaystyle\frac{1}{8}\biggl[\textstyle\sum\limits_{i=1}^d(\bfm_{n,i})^2\bfV_{n,i} \biggr]\\
&\leq\bfA\tau_1\eta^2(\gamma_n)^2-\displaystyle \frac {1}{4\chi}\biggl[\textstyle\sum\limits_{i=1}^d\bigl[\textstyle\frac {1}{M} \bigl[\sum_{j=1}^M\cG_1^i(\Theta_{n-1},X_{n,j})\bigl]\bigr]^2\bfV_{n,i} \biggr]\\
&+2(\cK^2\scrc^2(\kappa^{-1}\cA_2)^{\tau_2}+1024\cK^2\scrc^2(\bfA\kappa^{-1}(1+\Gamma))^{\tau_2})\textstyle d\gamma_n\varepsilon^{-1}-\displaystyle\frac{1}{8}\biggl[\textstyle\sum\limits_{i=1}^d(\bfm_{n,i})^2\bfV_{n,i} \biggr].
      \end{split}
\end{equation}}
\argument{\lref{def: bfA};\lref{eqmain'6}}{for all $n\in \N\cap[\fn,\infty)$ that
\begin{equation}\llabel{eqmain}
      \begin{split}
      &\smalll(\Theta_n,\bfx)-\smalll(\vartheta,\bfx)+\frac{1}{2(1-\alpha)}\textstyle\biggl[\sum\limits_{i=1}^d(\bfm_{n,i})^2\bfV_{n,i}\biggr]\\
      &-\biggl[\smalll(\Theta_{n-1},\bfx)-\smalll(\vartheta,\bfx)+\frac{1}{2(1-\alpha)}\textstyle\biggl[\sum\limits_{i=1}^d(\bfm_{n-1,i})^2\bfV_{n-1,i}\biggr]\biggr]\\
&\leq\bfA\tau_1\eta^2(\gamma_n)^2-\displaystyle \frac {1}{4\chi}\biggl[\textstyle\sum\limits_{i=1}^d\bigl[\textstyle\frac {1}{M} \bigl[\sum_{j=1}^M\cG_1^i(\Theta_{n-1},X_{n,j})\bigl]\bigr]^2\bfV_{n,i} \biggr]\\
&+2050\cK^2\scrc^2(\bfA\kappa^{-1}(1+\Gamma))^{\tau_2}\textstyle d\gamma_n\varepsilon^{-1}-\displaystyle\frac{1}{8}\biggl[\textstyle\sum\limits_{i=1}^d(\bfm_{n,i})^2\bfV_{n,i} \biggr].
      \end{split}
\end{equation}}
\argument{\lref{def: A2};\lref{def: bfA};\lref{material 2};\lref{induction: eq2};the assumption that for all $n,m\in \N$ it holds that $X_{n,m}\in \setX$}{that for all $i\in \{1,2,\dots,\fd\}$, $j\in \{1,2,\dots,M\}$, $k\in \{1,2,\dots,n\}$ it holds that
\begin{equation}\llabel{induction: eq3}
\begin{split}
  & |\textstyle \cG_1^i(\Theta_{k-1},X_{k,j})|\leq \|\cG_1(\Theta_{k-1},X_{k,j})\|\leq \cK\|\Theta_{k-1}-\vartheta\|(1+\|\Theta_{k-1}-\vartheta\|^{\tau_1}+2\|\vartheta\|^{\tau_1})+\rho
   \\
   &\leq \textstyle\cK(\frac{2\bfA(1+\Gamma)}{\kappa})^{1/2}\bigl(1+(\frac{2\bfA(1+\Gamma)}{\kappa})^{\tau_1/2}+2\|\vartheta\|^{\tau_1}\bigr)+\rho\\
   &\textstyle\leq \cK(\frac{2\bfA(1+\Gamma)}{\kappa})^{1/2}\bigl((\frac{2\cA_2}{\kappa})^{\tau_1/2}+(\frac{2\bfA(1+\Gamma)}{\kappa})^{\tau_1/2}+2(\frac{2\cA_2}{\kappa})^{\tau_1/2}\bigr)+\cK(\frac{2\cA_2}{\kappa})^{(1+\tau_1)/2}\\
   &\textstyle\leq 5\cK(\frac{2\bfA(1+\Gamma)}{\kappa})^{(1+\tau_1)/2}.
   \end{split}
\end{equation}}
\argument{\lref{induction: eq3};}{that for all $i\in \{1,2,\dots,\fd\}$, $k\in \{1,2,\dots,n\}$ it holds that
\begin{equation}\llabel{induction: eq4}
    \bigl| \textstyle\frac 1M \bigl(\textstyle\sum_{j=1}^M\cG_1^i(\Theta_{k-1},X_{k,j})\bigr)\bigr|\leq \sqrt{80}\cK(\bfA\kappa^{-1}(1+\Gamma))^{(1+\tau_1)/2}.
\end{equation}}
\argument{\lref{def: V};\lref{def: A2};\lref{def: bfA};\lref{induction: eq4};\lref{eq18.5}}{that for all $i\in \{1,2,\dots,\fd\}$ it holds that
\begin{equation}\llabel{induction: eq5}
\begin{split}
    \frac{\bbV_{n,i}}{1-\beta^n}&\leq \frac{1-\beta}{1-\beta^n}\biggl[\textstyle\sum\limits_{k=1}^n\beta^{n-k}\textstyle \bigl[\sqrt{80}\cK(\bfA\kappa^{-1}(1+\Gamma))^{(1+\tau_1)/2}\bigr]^2\biggr]+\displaystyle\frac{\cV^2}{1-\beta^n}\\
   &= \frac{1}{\sum_{k=1}^n\beta^{n-k}}\biggl[\textstyle\sum\limits_{k=1}^n\beta^{n-k}\textstyle \bigl[\sqrt{80}\cK(\bfA\kappa^{-1}(1+\Gamma))^{(1+\tau_1)/2}\bigr]^2\biggr]+
   \displaystyle\frac{\cV^2}{1-\beta^n}\\
   &\leq 80\cK^2(\bfA\kappa^{-1}(1+\Gamma))^{1+\tau_1}+\frac{\cV^2}{1-\beta}\\
   &= 80\cK^2(\bfA\kappa^{-1}(1+\Gamma))^{1+\tau_1}+\cK^2\kappa^{-1-\tau_1}\biggl(\frac{\cK^{-2}\kappa^{1+\tau_1}\cV^2}{1-\beta}\biggr)\\
   &\leq 80\cK^2(\bfA\kappa^{-1}(1+\Gamma))^{1+\tau_1}+\cK^2\kappa^{-1-\tau_1}(\cA_2)^{1+\tau_1}\\
   &\leq 80\cK^2(\bfA\kappa^{-1}(1+\Gamma))^{1+\tau_1}+\cK^2\kappa^{-1-\tau_1}\bfA^{1+\tau_1}\leq 81\cK^2(\bfA\kappa^{-1}(1+\Gamma))^{1+\tau_1}.
    \end{split}
\end{equation}}
\argument{\lref{induction: eq5};\lref{def: bfV};}{for all $i\in\{1,2,\dots,\fd\}$ that
\begin{equation}\llabel{induction: eq6'}
\begin{split}
  \textstyle\bfV_{n,i}=\frac{\gamma_n}{1-\alpha^n}\Bigl[\varepsilon+\bigl[\frac{\democrator_{n,i}}{1-\beta^n}\bigr]^{\nicefrac{1}{2}}\Bigr]^{-1}\geq \gamma_n\bigl[\varepsilon+9\cK(\bfA\kappa^{-1}(1+\Gamma))^{(1+\tau_1)/2}\bigr]^{-1}.
  \end{split}
\end{equation}}
\argument{\lref{induction: eq6'};\lref{def: A2};\lref{def: bfA}}{for all $i\in\{1,2,\dots,\fd\}$ that
\begin{equation}\llabel{induction: eq6}
\begin{split}
\textstyle\bfV_{n,i}&\geq \gamma_n\bigl[\cK\kappa^{-(1+\tau_1)/2}(\cA_2)^{(1+\tau_1)/2}+9\cK(\bfA\kappa^{-1}(1+\Gamma))^{(1+\tau_1)/2}\bigr]^{-1}\\
  &\geq  \gamma_n\bigl[\cK\kappa^{-(1+\tau_1)/2}\bfA^{(1+\tau_1)/2}+9\cK(\bfA\kappa^{-1}(1+\Gamma))^{(1+\tau_1)/2}\bigr]^{-1}\\
  &\geq\gamma_n\bigl[10\cK(\bfA\kappa^{-1}(1+\Gamma))^{(1+\tau_1)/2}\bigr]^{-1}.
  \end{split}
\end{equation}}
In our proof of \lref{need to prove} we distinguish between the case $\|\Theta_{n-1}-\vartheta\|\geq700\cK^{3/2}\kappa^{-(5+\tau_1+2\tau_2)/4}\scrc\varepsilon^{-1/2}(\chi d)^{1/2}\allowbreak(\bfA(1+\Gamma))^{(1+\tau_1+2\tau_2)/4}$, the case $\sum_{i=1}^d (\bfm_{n,i})^2\geq 164000\cK^3\scrc^2d\varepsilon^{-1}\allowbreak(\bfA\kappa^{-1}(1+\Gamma))^{(1+\tau_1+2\tau_2)/2}$, and the case $\|\Theta_{n-1}-\vartheta\|\leq 700\cK^{3/2}\kappa^{-(5+\tau_1+2\tau_2)/4}\scrc\varepsilon^{-1/2}(\chi d)^{1/2}(\bfA(1+\Gamma))^{(1+\tau_1+2\tau_2)/4}$ and $\sum_{i=1}^d (\bfm_{n,i})^2\leq 164000\cK^3\scrc^2d\varepsilon^{-1}(\bfA\kappa^{-1}(1+\Gamma))^{(1+\tau_1+2\tau_2)/2}$. We first prove \lref{need to prove} in the case
\begin{equation}\llabel{induction: case 1}
    \|\Theta_{n-1}-\vartheta\|\geq 700\cK^{3/2}\kappa^{-(5+\tau_1+2\tau_2)/4}\scrc\varepsilon^{-1/2}(\chi d)^{1/2}(\bfA(1+\Gamma))^{(1+\tau_1+2\tau_2)/4}.
\end{equation}
\startnewargseq
\argument{\lref{material 2};}{that for all $x_1,x_2,\dots,x_M\in \setX$ it holds that
\begin{equation}\llabel{eq15t}
\begin{split}
      \bigl\|\textstyle\frac 1M\textstyle\sum_{j=1}^M \cG_1(\Theta_{n-1},x_j)\bigr\|
      \geq \kappa\|\Theta_{n-1}-\vartheta\|-\rho.
      \end{split}
\end{equation}}
\argument{\lref{def: A3};\lref{def: bfA}}{that
\begin{equation}\llabel{eqphu1}
\begin{split}
&\cK^{3/2}\kappa^{-(1+\tau_1+2\tau_2)/4}\scrc(1+\Gamma)^{(1+\tau_1+2\tau_2)/4}\varepsilon^{-1/2}(\chi d)^{1/2}\bfA^{(1+\tau_1+2\tau_2)/4}\\&\textstyle \geq \cK^{3/2}\kappa^{-(1+\tau_1+2\tau_2)/4}\scrc(1+\Gamma)^{(1+\tau_1+2\tau_2)/4}\varepsilon^{-1/2}(\chi d)^{1/2}\\
&\quad\textstyle\cdot\Bigl(\frac{\rho^{4/(1+\tau_1+2\tau_2)}\kappa}{(\chi d\varepsilon^{-1}\scrc^2\cK^3)^{2/(1+\tau_1+2\tau_2)}(1+\Gamma)}\Bigr)^{(1+\tau_1+2\tau_2)/4}\\
&= \rho.
   \end{split}
\end{equation}}
\argument{\lref{induction: case 1};\lref{eqphu1};\lref{eq15t};}{that for all $x_1,x_2,\dots,x_M\in \setX$ it holds that
\begin{equation}\llabel{induction: eq7}
\begin{split}
    &\bigl\|\textstyle\frac 1M\textstyle\sum_{j=1}^M \cG_1(\Theta_{n-1},x_j)\bigr\|\geq \kappa\|\Theta_{n-1}-\vartheta\|-\rho\\
    &\geq 700\kappa\cK^{3/2}\kappa^{-(5+\tau_1+2\tau_2)/4}\scrc\varepsilon^{-1/2}(\chi d)^{1/2}(\bfA(1+\Gamma))^{(1+\tau_1+2\tau_2)/4}-\rho\\
    &\geq 700\kappa\cK^{3/2}\kappa^{-(5+\tau_1+2\tau_2)/4}\scrc\varepsilon^{-1/2}(\chi d)^{1/2}(\bfA(1+\Gamma))^{(1+\tau_1+2\tau_2)/4}\\
    &-\cK^{3/2}\kappa^{-(1+\tau_1+2\tau_2)/4}\scrc(1+\Gamma)^{(1+\tau_1+2\tau_2)/4}\varepsilon^{-1/2}(\chi d)^{1/2}\bfA^{(1+\tau_1+2\tau_2)/4}\\
&=699\cK^{3/2}\scrc\varepsilon^{-1/2}(\chi d)^{1/2}(\bfA\kappa^{-1}(1+\Gamma))^{(1+\tau_1+2\tau_2)/4}.
    \end{split}
\end{equation}}
\argument{\lref{induction: eq7};the fact that for all $n,m\in \N$ it holds that $X_{n,m}\in \setX$}{that
\begin{equation}\llabel{induction: eq8}
\begin{split}
&\textstyle\sum\limits_{i=1}^\fd \bigl[\textstyle \frac 1M \sum_{j=1}^M\cG_1^i(\Theta_{n-1},X_{n,j})\bigr]^2 \textstyle=
     \bigl\|\textstyle\frac 1M\textstyle\sum_{j=1}^M \cG_1(\Theta_{n-1},X_{n,j})\bigr\|^2\\
     &\geq699^2\cK^{3}\scrc^2\varepsilon^{-1}\chi d(\bfA\kappa^{-1}(1+\Gamma))^{(1+\tau_1+2\tau_2)/2}\\
     &\geq 82000\cK^{3}\scrc^2\varepsilon^{-1}\chi d(\bfA\kappa^{-1}(1+\Gamma))^{(1+\tau_1+2\tau_2)/2}.
     \end{split}
\end{equation}}
\argument{\lref{induction: eq8};\lref{eqmain};\lref{induction: eq6}}{that
\begin{equation}\llabel{induction: eq9}
\begin{split}
       &\smalll(\Theta_n,\bfx)-\smalll(\vartheta,\bfx)+\frac{1}{2(1-\alpha)}\textstyle\biggl[\sum\limits_{i=1}^d(\bfm_{n,i})^2\bfV_{n,i}\biggr]\\
      &-\biggl[\smalll(\Theta_{n-1},\bfx)-\smalll(\vartheta,\bfx)+\frac{1}{2(1-\alpha)}\textstyle\biggl[\sum\limits_{i=1}^d(\bfm_{n-1,i})^2\bfV_{n-1,i}\biggr]\biggr]\\
&\leq\bfA\tau_1\eta^2(\gamma_n)^2-\displaystyle \frac {1}{4\chi}\biggl[\textstyle\sum\limits_{i=1}^d\bigl[\textstyle\frac {1}{M} \bigl[\sum_{j=1}^M\cG_1^i(\Theta_{n-1},X_{n,j})\bigl]\bigr]^2\bfV_{n,i} \biggr]\\
&+2050\cK^2\scrc^2(\bfA\kappa^{-1}(1+\Gamma))^{\tau_2}\textstyle d\gamma_n\varepsilon^{-1}\\
&\leq\bfA\tau_1\eta^2(\gamma_n)^2 \\
&-\displaystyle \frac {1}{4\chi}\gamma_n\bigl[10\cK(\bfA\kappa^{-1}(1+\Gamma))^{(1+\tau_1)/2}\bigr]^{-1}82000\cK^{3}\scrc^2\varepsilon^{-1}\chi d(\bfA\kappa^{-1}(1+\Gamma))^{(1+\tau_1+2\tau_2)/2}\\
&+2050\cK^2\scrc^2(\bfA\kappa^{-1}(1+\Gamma))^{\tau_2}d\gamma_n\varepsilon^{-1}\\
&= \bfA\tau_1\eta^2(\gamma_n)^2.
\end{split}
\end{equation}}
\argument{\lref{induction: eq9};\lref{induction: assume}}{that 
\begin{equation}\llabel{evidence 1}
\begin{split}
 & \smalll(\Theta_n,\bfx)-\smalll(\vartheta,\bfx)+\frac{1}{2(1-\alpha)}\textstyle\biggl[\sum\limits_{i=1}^d(\bfm_{n,i})^2\bfV_{n,i}\biggr]\\
 &\leq \smalll(\Theta_{n-1},\bfx)-\smalll(\vartheta,\bfx)+\frac{1}{2(1-\alpha)}\textstyle\biggl[\sum\limits_{i=1}^d(\bfm_{n-1,i})^2\bfV_{n-1,i}\biggr]+ \bfA\tau_1\eta^2(\gamma_n)^2\\
 &\leq\textstyle \bfA+\bfA\biggl[\sum\limits_{h=1}^{n-1}\tau_1\eta^2(\gamma_h)^2\biggr]+\bfA\tau_1\eta^2(\gamma_n)^2=\bfA+\bfA\sum\limits_{h=1}^{n}\tau_1\eta^2(\gamma_h)^2.
    \end{split}
\end{equation}}
This proves \lref{need to prove} in the case $ \|\Theta_{n-1}-\vartheta\|\geq 700\cK^{3/2}\kappa^{-(5+\tau_1+2\tau_2)/4}\scrc(1+\Gamma)^{(1+\tau_1+2\tau_2)/4}\varepsilon^{-1/2}(\chi d)^{1/2}\allowbreak\bfA^{(1+\tau_1+2\tau_2)/4}$. Next we prove \lref{need to prove} in the case 
\begin{equation}\llabel{induction: case 3}
    \sum_{i=1}^d (\bfm_{n,i})^2\geq 164000\cK^3\scrc^2d\varepsilon^{-1}(\bfA\kappa^{-1}(1+\Gamma))^{(1+\tau_1+2\tau_2)/2}.
\end{equation}
\startnewargseq
\argument{\lref{eqmain};\lref{induction: eq6};\lref{induction: case 3}}{that
 \begin{equation}\llabel{case 3: eq1}
      \begin{split}
      &\smalll(\Theta_n,\bfx)-\smalll(\vartheta,\bfx)+\frac{1}{2(1-\alpha)}\textstyle\biggl[\sum\limits_{i=1}^d(\bfm_{n,i})^2\bfV_{n,i}\biggr]\\
      &-\biggl[\smalll(\Theta_{n-1},\bfx)-\smalll(\vartheta,\bfx)+\frac{1}{2(1-\alpha)}\textstyle\biggl[\sum\limits_{i=1}^d(\bfm_{n-1,i})^2\bfV_{n-1,i}\biggr]\biggr]\\
&\leq\bfA\tau_1\eta^2(\gamma_n)^2
+2050\cK^2\scrc^2(\bfA\kappa^{-1}(1+\Gamma))^{\tau_2}\textstyle d\gamma_n\varepsilon^{-1}-\displaystyle\frac{1}{8}\biggl[\textstyle\sum\limits_{i=1}^d(\bfm_{n,i})^2\bfV_{n,i} \biggr]\\
&\leq \bfA\tau_1\eta^2(\gamma_n)^2
+2050\cK^2\scrc^2(\bfA\kappa^{-1}(1+\Gamma))^{\tau_2}\textstyle d\gamma_n\varepsilon^{-1}\\
&-\displaystyle\frac{1}{8}\biggl[\textstyle\sum\limits_{i=1}^d(\bfm_{n,i})^2 \biggr]\gamma_n\bigl[10\cK(\bfA\kappa^{-1}(1+\Gamma))^{(1+\tau_1)/2}\bigr]^{-1} \\
&\leq \bfA\tau_1\eta^2(\gamma_n)^2
+2050\cK^2\scrc^2(\bfA\kappa^{-1}(1+\Gamma))^{\tau_2}\textstyle d\gamma_n\varepsilon^{-1}\\
&-\displaystyle\biggl(\frac{1}{8}\biggr)164000\cK^3\scrc^2d\varepsilon^{-1}(\bfA\kappa^{-1}(1+\Gamma))^{(1+\tau_1+2\tau_2)/2}\gamma_n\bigl[10\cK(\bfA\kappa^{-1}(1+\Gamma))^{(1+\tau_1)/2}\bigr]^{-1}\\
&=\bfA\tau_1\eta^2(\gamma_n)^2.
      \end{split}
  \end{equation}}
  \argument{\lref{case 3: eq1};\lref{induction: assume}}{that
  \begin{equation}
      \begin{split}
          &\smalll(\Theta_n,\bfx)-\smalll(\vartheta,\bfx)+\frac{1}{2(1-\alpha)}\textstyle\biggl[\sum\limits_{i=1}^d(\bfm_{n,i})^2\bfV_{n,i}\biggr]\\
      &\leq \smalll(\Theta_{n-1},\bfx)-\smalll(\vartheta,\bfx)+\frac{1}{2(1-\alpha)}\textstyle\biggl[\sum\limits_{i=1}^d(\bfm_{n-1,i})^2\bfV_{n-1,i}\biggr]+\bfA\tau_1\eta^2(\gamma_n)^2\\
      &\leq \textstyle \bfA+\bfA\biggl[\sum\limits_{h=1}^{n-1}\tau_1\eta^2(\gamma_h)^2\biggr]+\bfA\tau_1\eta^2(\gamma_n)^2=\bfA+\bfA\sum\limits_{h=1}^{n}\tau_1\eta^2(\gamma_h)^2.
      \end{split}
  \end{equation}}
This establishes \lref{need to prove} in the case $\sum_{i=1}^d (\bfm_{n,i})^2\geq 164000\cK^3\scrc^2d\varepsilon^{-1}(\bfA\kappa^{-1}(1+\Gamma))^{(1+\tau_1+2\tau_2)/2}$.
We now prove \lref{need to prove} in the case
\begin{equation}\llabel{induction: case 2}
     \|\Theta_{n-1}-\vartheta\|\leq 700\cK^{3/2}\kappa^{-(5+\tau_1+2\tau_2)/4}\scrc\varepsilon^{-1/2}(\chi d)^{1/2}(\bfA(1+\Gamma))^{(1+\tau_1+2\tau_2)/4}
\end{equation}
\begin{equation}\llabel{induction: case 2'}
     \text{and}\qquad \textstyle\sum\limits_{i=1}^d (\bfm_{n,i})^2\leq 164000\cK^3\scrc^2d\varepsilon^{-1}(\bfA\kappa^{-1}(1+\Gamma))^{(1+\tau_1+2\tau_2)/2}.
\end{equation}
\startnewargseq
\argument{\lref{induction: case 2'};Cauchy Schwarz inequality}{that
\begin{equation}\llabel{case 2: eq1}
\begin{split}
     &\textstyle \sum\limits_{i=1}^d |\bfm_{n,i}|\leq \sqrt{d}\biggl[\sum\limits_{i=1}^d (\bfm_{n,i})^2\biggr]^{1/2}\leq 405\cK^{3/2}\scrc d\varepsilon^{-1/2}(\bfA\kappa^{-1}(1+\Gamma))^{(1+\tau_1+2\tau_2)/4}.
\end{split}
\end{equation}}
\argument{\lref{def: A4};\lref{eq: x2.5};\lref{eq14};\lref{induction: case 2};\lref{induction: case 2'};\lref{case 2: eq1}; the fact that $\gamma_n\leq \gamma_1$}{that
\begin{equation}\llabel{induction: eq10}
\begin{split}
   & \|\Theta_n-\vartheta\|\leq  \|\Theta_{n-1}-\vartheta\|+ \|\Theta_n-\Theta_{n-1}\|\\
    &\leq \textstyle\|\Theta_{n-1}-\vartheta\|+\biggl[\sum\limits_{i=1}^d|\bfm_{n,i}|\bfV_{n,i}\biggr]\leq \|\Theta_{n-1}-\vartheta\|+2\gamma_n\varepsilon^{-1}\biggl[\sum\limits_{i=1}^d|\bfm_{n,i}|\biggr]\\
    &\leq 700\cK^{3/2}\kappa^{-(5+\tau_1+2\tau_2)/4}\scrc\varepsilon^{-1/2}(\chi d)^{1/2}(\bfA(1+\Gamma))^{(1+\tau_1+2\tau_2)/4}\\
    &+910\gamma_1\varepsilon^{-1}\cK^{3/2}\scrc d\varepsilon^{-1/2}(\bfA\kappa^{-1}(1+\Gamma))^{(1+\tau_1+2\tau_2)/4}\\
    &\leq 1610(1+\gamma_1)\cK^{3/2}\kappa^{-(5+\tau_1+2\tau_2)/4}(1+\kappa)\scrc\varepsilon^{-3/2}(1+\varepsilon)(1+\chi^{1/2}) d(\bfA(1+\Gamma))^{(1+\tau_1+2\tau_2)/4}\\
    &\leq \cA_4\bfA^{(1+\tau_1+2\tau_2)/4}.
    \end{split}
\end{equation}}
\argument{\lref{induction: eq10};\lref{def: A5};\lref{def: bfA};\lref{material 1}; the fact that $3\tau_1+4\tau_2+(\tau_1)^2+2\tau_1\tau_2<2$; the fact that $\tau_1+2\tau_2<1$; the fact that $\cA_4\geq 1$;the fact that $\bfA\geq 1$}{that 
\begin{equation}\llabel{induction: eq11}
\begin{split}
   &\smalll(\Theta_{n},\bfx)-\smalll(\vartheta,\bfx)\leq \frac{\cK}{2}\|\Theta_n-\vartheta\|^2(1+\|\Theta_n-\vartheta\|^{\tau_1}+2\|\vartheta\|^{\tau_1})\\
  &\leq \frac{\cK}{2}\bigl[\cA_4\bfA^{(1+\tau_1+2\tau_2)/4}\bigr]^2[1+2\|\vartheta\|^{\tau_1}+(\cA_4)^{\tau_1}\bfA^{\tau_1(1+\tau_1+2\tau_2)/4}]\\ & \leq \frac{\cK}{2}\bigl[\cA_4\bfA^{(1+\tau_1+2\tau_2)/4}\bigr]^2[(1+2\|\vartheta\|^{\tau_1})(\cA_4)^{\tau_1}\bfA^{\tau_1(1+\tau_1+2\tau_2)/4}+(\cA_4)^{\tau_1}\bfA^{\tau_1(1+\tau_1+2\tau_2)/4}]
  \\
  &=\cK(1+\|\vartheta\|^{\tau_1})(\cA_4)^{2+\tau_1}\bfA^{(2+3\tau_1+4\tau_2+(\tau_1)^2+2\tau_1\tau_2)/4}\\
  &\leq\frac 12 (\cA_5)^{(2-3\tau_1-4\tau_2-(\tau_1)^2-2\tau_1\tau_2)/4}\bfA^{(2+3\tau_1+4\tau_2+(\tau_1)^2+2\tau_1\tau_2)/4}\\
  &\leq \frac 12 \bfA^{(2-3\tau_1-4\tau_2-(\tau_1)^2-2\tau_1\tau_2)/4}\bfA^{(2+3\tau_1+4\tau_2+(\tau_1)^2+2\tau_1\tau_2)/4}=\frac{\bfA}{2}.
    \end{split}
\end{equation}}
\argument{\lref{def: A6};\lref{def: bfA};\lref{eq14};\lref{induction: case 2'};the fact that $\tau_1+2\tau_2<1$ ;the fact that $\gamma_n\leq \gamma_1$}{that
\begin{equation}\llabel{induction: eq13}
\begin{split}
    &\frac{1}{2(1-\alpha)}\textstyle\biggl[\sum\limits_{i=1}^d(\bfm_{n,i})^2\bfV_{n,i}\biggr]\leq\displaystyle \frac{\gamma_1}{(1-\alpha)\varepsilon}\textstyle\biggl[\sum\limits_{i=1}^d(\bfm_{n,i})^2\biggr]\\
    &\leq  \frac{\gamma_1}{(1-\alpha)\varepsilon}\bigl[164000\cK^3\scrc^2d\varepsilon^{-1}(\bfA\kappa^{-1}(1+\Gamma))^{(1+\tau_1+2\tau_2)/2}\bigr]\\
    &=164000(1-\alpha)^{-1}\gamma_1\cK^3\kappa^{-(1+\tau_1+2\tau_2)/2}\scrc^2d\varepsilon^{-2}(1+\Gamma)^{(1+\tau_1+2\tau_2)/2}\bfA^{(1+\tau_1+2\tau_2)/2}\\
    &\leq \frac{1}{2}(\cA_6)^{(1-\tau_1-2\tau_2)/2}\bfA^{(1+\tau_1+2\tau_2)/2}\leq \frac{\bfA}{2}.
    \end{split}
\end{equation}}
\argument{\lref{induction: eq13};\lref{induction: eq11}}{that 
\begin{equation}\llabel{induction: eq14}
         \smalll(\Theta_{n},\bfx)-\smalll(\vartheta,\bfx)+\frac{1}{2(1-\alpha)}\textstyle\biggl[\sum\limits_{i=1}^d(\bfm_{n,i})^2\bfV_{n,i}\biggr] \leq\displaystyle\frac 12\bfA+\frac{1}{2}\bfA=\bfA\leq \bfA+\bfA\sum_{k=1}^n\tau_1\eta^2(\gamma_k)^2.
\end{equation}}
This proves \lref{need to prove} in the case $\|\Theta_{n-1}-\vartheta\|\leq  700\cK^{3/2}\kappa^{-(5+\tau_1+2\tau_2)/4}\scrc\varepsilon^{-1/2}(\chi d)^{1/2}(\bfA(1+\Gamma))^{(1+\tau_1+2\tau_2)/4}$ and $\sum_{i=1}^d (\bfm_{n,i})^2\leq 164000\cK^3\scrc^2d\varepsilon^{-1}(\bfA\kappa^{-1}(1+\Gamma))^{(1+\tau_1+2\tau_2)/2}$. Induction thus establishes \lref{need to prove}.
\startnewargseq
\argument{\lref{def: eta};\lref{arg1};\lref{eqq1};\lref{need to prove};}{that for all $n\in \N$ it holds that
\begin{equation}\llabel{eqm1}
\begin{split}
    \smalll(\Theta_n,\bfx)-\smalll(\vartheta,\bfx)&\leq \smalll(\Theta_{n},\bfx)-\smalll(\vartheta,\bfx)+\frac{1}{2(1-\alpha)}\textstyle\biggl[\sum\limits_{i=1}^d(\bfm_{n,i})^2\bfV_{n,i}\biggr]\\&\leq \bfA+\bfA\sum_{k=1}^n\tau_1\eta^2(\gamma_k)^2\leq 
   \bfA+\bfA\Gamma.
    \end{split}
\end{equation}}
\argument{\lref{eqm1};\lref{material 1}}{for all $n\in \N$ that
\begin{equation}\llabel{eqm2}
    \|\Theta_n-\vartheta\|^2\leq \frac{2}{\kappa}(\smalll(\Theta_n,\bfx)-\smalll(\vartheta,\bfx))\leq \frac{2\bfA(1+\Gamma)}{\kappa}.
\end{equation}}
\argument{\lref{eqm2};}{that for all $n\in \N$ it holds that
\begin{equation}\llabel{eqm3}
    \|\Theta_n\|\leq \|\Theta_n-\vartheta\|+\|\vartheta\|\leq [2\bfA\kappa^{-1}(1+\Gamma)]^{1/2}+\|\vartheta\|.
\end{equation}}
\argument{\lref{eqm3};\lref{def: bfA}}{\lref{conclude}\dott}
\end{aproof}
\subsection{Non-uniform a priori bounds for Adam for quadratic example SOPs}\label{subsec: non-uniform priori}
 \begin{tcolorbox}[colback=white!95!gray,
                  colframe=black,
                  boxrule=0.5pt,
                  sharp corners,
                  enhanced,
                  breakable,
                 ]
\begin{athm}{lemma}{lem: verify 1}
    Let $d\in \N$, $r\in [2,\infty)$, $\delta\in [0,\infty)$, let $A\in \R^{d\times d}$ be invertible, and let $\smalll\colon\R^\fd\times\R^\fd\to\R$ satisfy for all $\theta,x\in \R^\fd$ that
    \begin{equation}\llabel{def: loss}
        \smalll(\theta,x)=\|A\theta-x\|^2+\delta\|\theta\|^{r}.
    \end{equation}
 Then 
 \begin{enumerate}[label=(\roman*)]
 \item \label{item 1: verify 1} it holds that $\smalll\in C^2(\R^d\times \R^d,\R)$ and
     \item \label{item 2: verify 1} there exist $\kappa,\cK\in (0,\infty)$ which satisfy for all $\theta,\vartheta,x\in \R^\fd$ that 
     \begin{equation}
         \|(\nabla_x\smalll)(\theta,x)-(\nabla_x\smalll)(\vartheta,x)\|\leq \cK\|\theta-\vartheta\|,\qquad \spro{ \theta-\vartheta, (\nabla_\theta\smalll)(\theta,x)- (\nabla_\theta\smalll)(\vartheta,x)}\geq \kappa\|\theta-\vartheta\|^2,
     \end{equation}
     \begin{equation}
        \text{and}\qquad \|(\nabla_\theta\smalll)(\theta,x)-(\nabla_\theta\smalll)(\vartheta,x)\|\leq \cK\|\theta-\vartheta\|(1+\|\theta\|^{r-2}+\|\vartheta\|^{r-2}).
     \end{equation}
 \end{enumerate}
\end{athm}
\end{tcolorbox}
\begin{aproof}
    \argument{\lref{def: loss};}{that for all $\theta,x\in \R^\fd$ it holds that
    \begin{equation}\llabel{eq1}
        (\nabla_\theta\smalll)(\theta,x)=2A^{*}(A\theta-x)+\delta r\theta\|\theta\|^{r-2+2\mathbbm 1_{\{0\}}(\theta)}
    \end{equation}
    and
    \begin{equation}\llabel{eq2}
        (\nabla_x\smalll)(\theta,x)=-2(A\theta-x).
    \end{equation}}
    \argument{\lref{eq2};the fact that $r\geq 2$}{\cref{item 1: verify 1}\dott}
    \startnewargseq
    \argument{the fundamental theorem of calculus;}{that for all $x,y\in (0,\infty)$ it holds that 
    \begin{equation}\llabel{eq2.1}
        |x^{r-2}-y^{r-2}|\leq |r-2|\max\{x^{r-3},y^{r-3}\} |x-y|.
    \end{equation}}
    \argument{\lref{eq1};\lref{eq2.1};the Cauchy-Schwarz inequality;}{that for all $\theta,\vartheta\in \R^{\fd}\backslash\{0\}$, $x\in \R^\fd$ with $\|\theta\|^{r-3}\geq \|\vartheta\|^{r-3}$ it holds that
    \begin{equation}\llabel{evidence 1.1}
    \begin{split}
       & \|(\nabla_\theta\smalll)(\theta,x)-(\nabla_\theta)\smalll(\vartheta,x)\|=\|2A^{*}A(\theta-\vartheta)+\delta r\theta\|\theta\|^{r-2}-\delta r\vartheta\|\vartheta\|^{r-2}\|\\
        &\leq 2\|A^{*}A(\theta-\vartheta)\|+\|\delta r\theta\|\theta\|^{r-2}-\delta r\vartheta\|\vartheta\|^{r-2}\|\\
        &\leq 2\|A^{*}A\|\|\theta-\vartheta\|+\|\delta r\theta\|\theta\|^{r-2}-\delta r\theta\|\vartheta\|^{r-2}\|+\|\delta r\theta\|\vartheta\|^{r-2}-\delta r\vartheta\|\vartheta\|^{r-2}\|\\
        &\leq 2\|A^{*}A\|\|\theta-\vartheta\|+\delta r|r-2|\|\theta\|^{r-2}\|\theta-\vartheta\|+\delta r\|\theta-\vartheta\|\|\vartheta\|^{r-2}\\
        &\leq \|\theta-\vartheta\|(2\|A^{*}A\|+\delta r|r-2|\|\theta\|^{r-2}+\delta r\|\vartheta\|^{r-2}).
        \end{split}
    \end{equation}}
    \argument{\lref{evidence 1.1};}{that for all $\theta,\vartheta\in \R^d\backslash\{0\}$, $x\in \R^\fd$ it holds that
    \begin{equation}\llabel{evidence 1.2}
    \begin{split}
       &\|(\nabla_\theta\smalll)(\theta,x)-(\nabla_\theta\smalll)(\vartheta,x)\|
        \\
        &\leq \|\theta-\vartheta\|(2\|A^{*}A\|+\delta \max\{r,r|r-2|\}\|\theta\|^{r-2}+\delta \max\{r,r|r-2|\}\|\vartheta\|^{r-2}).
        \end{split}
    \end{equation}}
    \argument{\lref{eq1};}{for all $\theta\in \R^d\backslash\{0\}$, $x\in \R^d$ that
    \begin{equation}\llabel{evidence 1.3}
        \begin{split}
           & \|(\nabla_\theta\smalll)(\theta,x)-(\nabla_\theta\smalll)(0,x)\|=\|2A^{*}A\theta+\delta r\theta\|\theta\|^{r-2}\|\\
            &\leq 2\|A^{*}A\theta\|+\|\delta r\theta\|\theta\|^{r-2}\|\leq \|\theta\|(2\|A^{*}A\|+\delta r\|\theta\|^{r-2}).
        \end{split}
    \end{equation}}
    \argument{\lref{evidence 1.3};\lref{evidence 1.2}}{for all $\theta,\vartheta,x\in \R^\fd$ that
    \begin{equation}\llabel{evidence 1}
    \begin{split}
      & \|(\nabla_\theta\smalll)(\theta,x)-(\nabla_\theta\smalll)(\vartheta,x)\|
        \\
        &\leq \|\theta-\vartheta\|(2\|A^{*}A\|+\delta \max\{r,r|r-2|\}\|\theta\|^{r-2}+\delta \max\{r,r|r-2|\}\|\vartheta\|^{r-2}).
        \end{split}
    \end{equation}}
    \argument{\lref{eq2};the Cauchy-Schwarz inequality}{for all $\theta,\vartheta,x\in \R^\fd$ that
    \begin{equation}\llabel{evidence 2}
                \|(\nabla_x\smalll)(\theta,x)-(\nabla_x\smalll)(\vartheta,x)\|=\|2A(\theta-\vartheta)\|\leq 2\|A\|\|\theta-\vartheta\|.
    \end{equation}}
    \argument{the fact that for all $\R^d\ni \theta\mapsto \|\theta\|^{r}\in \R$ is convex;the fact that $(\R^d\ni \theta\mapsto \|\theta\|^{r}\in \R)\in C^1(\R^d,\R)$;\cite[Lemma 5.7.3]{ArBePhi2024}}{that for all $\theta,\vartheta\in \R^d\backslash\{0\}$ it holds that
    \begin{equation}\llabel{evd3.0}
        \spro{\theta-\vartheta,\theta\|\theta\|^{r-2}-\vartheta\|\vartheta\|^{r-2}}\geq 0.
    \end{equation}}
    \argument{\lref{evd3.0};the fact that for all $\theta\in \R^d\backslash\{0\}$ it holds that $\spro{\theta,\theta\|\theta\|^{r-2}}\geq 0$;}{that for all $\theta,\vartheta\in \R^d$ it holds that
    \begin{equation}\llabel{evd3.1}
        \spro{\theta-\vartheta,\theta\|\theta\|^{r-2+2\mathbbm 1_{\{0\}}(\theta)}-\vartheta\|\vartheta\|^{r-2+2\mathbbm 1_{\{0\}}(\vartheta)}}\geq 0.
    \end{equation}}
    \argument{\lref{evd3.1};\lref{eq1};the fact that $\operatorname{det}(A)\neq 0$;the fact that $\delta\geq 0$; the Cauchy-Schwarz inequality;}{that for all $\theta,\vartheta,x\in \R^\fd$ it holds that
    \begin{equation}\llabel{evidence 3}
    \begin{split}
       & \spro{ \theta-\vartheta, (\nabla_\theta\smalll)(\theta,x)- (\nabla_\vartheta\smalll)(\vartheta,x)}\\
       &=\spro{\theta-\vartheta,2A^{*}A(\theta-\vartheta)}+ \spro{\theta-\vartheta,\delta r \theta\|\theta\|^{r-2+2\mathbbm 1_{\{0\}}(\theta)}-\delta r\vartheta\|\vartheta\|^{r-2+2\mathbbm 1_{\{0\}}(\vartheta)}}\\
       &\geq 2\|A(\theta-\vartheta)\|^2\geq 2(\|A^{-1}\|)^{-2}\|\theta-\vartheta\|^2.
        \end{split}
    \end{equation}}
    \argument{\lref{evidence 1};\lref{evidence 2};\lref{evidence 3}}{\cref{item 2: verify 1}\dott}
\end{aproof}
 \begin{tcolorbox}[colback=white!95!gray,
                  colframe=black,
                  boxrule=0.5pt,
                  sharp corners,
                  enhanced,
                  breakable,
                 ]
\begin{athm}{cor}{cor: priori bound stochastic Adam 2 explosion}
      Let $\fd\in \N$, $\varepsilon\in (0,\infty)$, $\beta_1\in (0,1)$, $\beta_2\in ((\beta_1)^2,1)$, $\delta\in [0,\infty)$, $r\in [2,\frac{1+\sqrt{17}}{2})$, let $A\in \R^{d\times d}$ be invertible, let $\setX\subseteq \R^d$ be compact and convex, let $\smalll=(\smalll(\theta,x))_{(\theta,x)\in \R^\fd\times \R^d}\allowbreak\in C^{1}(\R^\fd\times \R^d,\R)$ satisfy for all $\theta,x\in \R^\fd$  that 
     \begin{equation}\llabel{def: loss}
        \smalll(\theta,x)=\|A\theta-x\|^2+\delta\|\theta\|^{r},
    \end{equation}
    let $(\Omega,\cF,\P)$ be a probability space,
 let $(\gamma_n)_{n\in \N}\subseteq \R$ be non-increasing, assume $\sum_{n=1}^\infty(\gamma_n)^2<\infty$, 
     for every $n,m\in \N$ let $X_{n,m}\colon \Omega\to \setX$ be a random variable, for every $k,M,n\in \N_0$ let $\Theta_n^{k,M}=(\Theta_{n,1}^{k,M},\dots,\allowbreak\Theta_{n,\fd}^{k,M})\colon \Omega\to\R^d$ be a random variable, and assume for all $k\in \{1,2\}$, $M,n\in \N$, $i\in \{1,2,\dots,\fd\}$ that 
    \begin{equation} \label{def: V: cor: priori bound stochastic Adam 2 explosion}
    \Theta^{k,M}_{0,i}= 0,\qquad \Theta_{n,i}^{k,M}=\beta_k \Theta_{n-1,i}^{k,M}+(1-\beta_k)\bigl[\textstyle \frac 1M \sum_{m=1}^M(\frac{\partial}{\partial\theta_i} \smalll)(\Theta_{n-1}^{0,M},X_{n,m})\bigr]^k,
    \end{equation}
    \begin{equation}\llabel{def: Theta}
\text{and}\qquad\textstyle\Theta_{n,i}^{0,M}=\Theta_{n-1,i}^{0,M} -\gamma_n\Bigl[\frac{\Theta_{n,i}^{1,M}}{1-(\beta_1)^n}\Bigr] \Bigl[\varepsilon+\Bigl[\frac{\Theta_{n,i}^{2,M}}{1-(\beta_2)^n}\Bigr]^{\nicefrac{1}{2}}\Bigr]^{-1}.
    \end{equation}
 Then there exists $c\in \R$ such that for all $M,n \in \N$ it holds that  
 \begin{equation}\llabel{conclude}
    \begin{split}
     \| \Theta_n^{0,M} \|\leq c(1+\|\Theta_0^{0,M}\|).
    \end{split} 
 \end{equation}
\end{athm}
\end{tcolorbox}
\begin{aproof}
    \argument{\cref{lem: priori bound stochastic Adam non explosion};\cref{lem: verify 1}}{\lref{conclude}\dott}
\end{aproof}
\subsection{Qualitative a priori bounds for Adam for general strongly convex SOPs}\label{subsec: deterministic bound}
 \begin{tcolorbox}[colback=white!95!gray,
                  colframe=black,
                  boxrule=0.5pt,
                  sharp corners,
                  enhanced,
                  breakable,
                 ]
\begin{athm}{lemma}{lem: estimate fn}
    Let $\alpha\in [0,1)$, $\beta\in (\alpha^2,1)$. Then
    \begin{equation}\llabel{conclude}
    \begin{split}
        \log_\beta\biggl(\frac{\beta-\alpha^2}{1-\alpha^2}\biggr)\leq\max\{4(1-\alpha)^{-1},4(\beta-\alpha^2)^{-1}\}.
        \end{split}
    \end{equation}
\end{athm}
\end{tcolorbox}
\begin{aproof}
Throughout this proof let $f\colon (0,1)\to\R$ satisfy for all $x\in (0,1)$ that
\begin{equation}\llabel{def: f}
    f(x)=\frac{\ln(1-x)}{x}.
\end{equation}
\argument{\lref{def: f}}{that for all $x\in (0,1)$ it holds that
\begin{equation}\llabel{eq1}
    f'(x)=\frac{\frac{-x}{1-x}-\ln(1-x)}{x^2}=\frac{1-\frac{1}{1-x}-\ln(1-x)}{x^2}=\frac{\ln(\frac{1}{1-x})+(1-\frac{1}{1-x})}{x^2}.
\end{equation}}
\argument{\lref{eq1};the fact that for all $x\in (1,\infty)$ it holds that $\ln(x)<x-1$;}{that for all $x\in (0,1)$ it holds that
\begin{equation}\llabel{eq2}
    f'(x)<\frac{\frac{1}{1-x}-1+(1-\frac{1}{1-x})}{x^2}=0.
\end{equation}}
\argument{\lref{eq2};\lref{def: f};the fact that $\lim_{x \to 0}\frac{\ln(1-x)}{x}=-1$}{that for all $y\in (0,1)$, $x\in (0,y]$ it holds that
\begin{equation}\llabel{eq3'}
  \frac{\ln(1-y)}{y}=f(y)\leq f(x)=  \frac{\ln(1-x)}{x}\leq -1.  
\end{equation}}
\argument{\lref{eq3'}}{for all $y\in (0,1)$, $x\in (0,y]$ that
\begin{equation}\llabel{eq3}
  \frac{-\ln(1-y)}{y}\geq  \frac{-\ln(1-x)}{x}\geq 1.  
\end{equation}}
    \argument{\lref{eq3}; the fact that $0<1-\beta$; the fact that $0<\frac{1-\beta}{1-\alpha^2}\leq \frac{1-\min\{\beta,\frac{1+\alpha^2}{2}\}}{1-\alpha^2}<1$; the fact that $1-\alpha\leq 1-\alpha^2$}{that
    \begin{equation}\llabel{eq4}
    \begin{split}
        &\log_\beta\biggl(\frac{\beta-\alpha^2}{1-\alpha^2}\biggr)=\log_{\beta}(e)\ln\biggl(\frac{\beta-\alpha^2}{1-\alpha^2}\biggr)=\frac{\ln\bigl(\frac{\beta-\alpha^2}{1-\alpha^2}\bigr)}{\ln(\beta)}=\frac{\ln\bigl(1-\frac{1-\beta}{1-\alpha^2}\bigr)}{\ln(1-(1-\beta))}\\
        &=\biggl(\frac{-\ln(1-(1-\beta))}{1-\beta}\biggr)^{-1}\frac{-\ln\bigl(1-\frac{1-\beta}{1-\alpha^2}\bigr)}{\frac{1-\beta}{1-\alpha^2}}\frac{1}{1-\alpha^2}\\
        &\leq \frac{-\ln\bigl(1-\frac{1-\beta}{1-\alpha^2}\bigr)}{\frac{1-\beta}{1-\alpha^2}}\frac{1}{1-\alpha^2}\leq \frac{-\ln\bigl(1-\frac{1-\min\{\beta,\frac{1+\alpha^2}{2}\}}{1-\alpha^2}\bigr)}{\frac{1-\min\{\beta,\frac{1+\alpha^2}{2}\}}{1-\alpha^2}}\frac{1}{1-\alpha^2}\\
        &=\frac{-\ln\Bigl(\frac{\min\{\beta-\alpha^2,\frac{1-\alpha^2}{2}\}}{1-\alpha^2}\Bigr)}{1-\min\{\beta,\frac{1+\alpha^2}{2}\}}\leq \frac{-2\ln\bigl(\min\{\frac{\beta-\alpha^2}{1-\alpha^2},\frac 12\}\bigr)}{1-\alpha^2}\leq \frac{-2\ln\bigl(\min\{\frac{\beta-\alpha^2}{1-\alpha^2},\frac 12\}\bigr)}{1-\alpha}.
        \end{split}
    \end{equation}}
    \argument{\lref{eq4}; the fact that for all $x\in (0,1)$ it holds that $-\ln(x)\leq x^{-1}$}{that
    \begin{equation}\llabel{eq5}
    \begin{split}
        &\log_\beta\biggl(\frac{\beta-\alpha^2}{1-\alpha^2}\biggr)\leq 2(1-\alpha)^{-1}\max\biggl\{2,\frac{1-\alpha^2}{\beta-\alpha^2}\biggr\}=\max\biggl\{4(1-\alpha)^{-1},\frac{2(1+\alpha)}{\beta-\alpha^2}\biggr\}\\
        &\leq \max\{4(1-\alpha)^{-1},4(\beta-\alpha^2)^{-1}\}.
        \end{split}
    \end{equation}}
    \end{aproof}
        \begin{tcolorbox}[colback=white!95!gray,
                  colframe=black,
                  boxrule=0.5pt,
                  sharp corners,
                  enhanced,
                  breakable,
                 ]
\begin{athm}{cor}{cor: priori bound stochastic Adam 1 non explosion}
      Let $\fd,\dimX\in \N$, $\kappa,\cK,\varepsilon\in (0,\infty)$, $\tau\in (0,\nicefrac{1}{2})$, let $O\subseteq \R^\dimX$ be open, let $\setX\subseteq O$ be compact and convex, let $\smalll=(\smalll(\theta,x))_{(\theta,x)\in \R^\fd\times O}\allowbreak\in C^{1}(\R^\fd\times O,\R)$ satisfy for all $\theta,\vartheta\in \R^\fd$, $x\in \setX$ that $\spro{ \theta-\vartheta, (\nabla_\theta\smalll)(\theta,x)- (\nabla_\theta\smalll)(\vartheta,x)}\geq \kappa\|\theta-\vartheta\|^2$ and
 \begin{equation}\llabel{assumption2}
 \textstyle\max\bigl\{\|(\nabla_\theta\smalll)(\theta,x)- (\nabla_\theta\smalll)(\vartheta,x)\|,\frac{\|(\nabla_x\smalll)(\theta,x)-(\nabla_x\smalll)(\vartheta,x)\|}{1+\|\theta\|^{\tau}+\|\vartheta\|^{\tau}}\bigr\}\leq  \cK\|\theta-\vartheta\|,
 \end{equation} 
 let $(\gamma_n)_{n\in \N}\subseteq \R$ be non-increasing, and assume $\lim_{n\to\infty}\gamma_n=0$. Then there exists $c\in \R$ such that for every $M\in \N$, $\alpha\in [0,1)$, $\beta\in (\alpha^2,1)$, every
     $(X_{n,m})_{(n,m)\in \N^2}\subseteq \setX$, every $\Theta=(\Theta^{1},\dots,\Theta^{\fd}\allowbreak)\colon\N_0\to\R^d$, every $\bfm=(\bfm^{1},\dots,\bfm^{n})\colon\N_0\to\R^d$, and every $\democrator=(\democrator^{1},\dots,\allowbreak\democrator^{n})\colon\N_0\to\R^\fd$ with the property that for all $n\in \N$, $i\in \{1,2,\dots,\fd\}$ it holds that 
 \begin{equation} \llabel{def: bfm}
    \bfm^{i}_{0}= 0,\qquad \bfm^{i}_{n}=\alpha \bfm_{n-1}^{i}+(1-\alpha)\textstyle \frac 1M \sum_{m=1}^M(\frac{\partial}{\partial\theta_i} \smalll)(\Theta_{n-1},X_{n,m}),
    \end{equation}
    \begin{equation} \llabel{def: V}
    \democrator^{i}_{0}= 0,\qquad \democrator_{n}^{i}=\beta \democrator_{n-1}^{i}+(1-\beta)\bigl[\textstyle \frac 1M \sum_{m=1}^M(\frac{\partial}{\partial\theta_i} \smalll)(\Theta_{n-1},X_{n,m})\bigr]^2,
    \end{equation}
    \begin{equation}\llabel{def: Theta}
\text{and}\qquad\textstyle\Theta_{n}^{i}=\Theta_{n-1}^{i} -\gamma_n\Bigl[\frac{\bfm_{n}^{i}}{1-\alpha^n}\Bigr] \Bigl[\varepsilon+\Bigl[\frac{\democrator_{n}^{i}}{1-\beta^n}\Bigr]^{\nicefrac{1}{2}}\Bigr]^{-1}, 
    \end{equation}
 we have for all $n\in\N$ that
 \begin{equation}\llabel{conclude}
    \begin{split}
       \| \Theta_n\|&\leq \textstyle c(1-\alpha)^{-1/2}\| \Theta_0\|+c(\beta-\alpha^2)^{-2}+c(1-\alpha)^{-2}+c(1-\alpha)^{-1/(1-2\tau)}.
    \end{split} 
 \end{equation}
\end{athm}
\end{tcolorbox}
\begin{aproof}
Throughout this proof let $\bfx\in \setX$, $\scrc\in [1,\infty)$ satisfy $\setX\subseteq\{x\in \R^\dimX\colon \|x\|\leq \scrc\}$ and let  $N\in \N$ satisfy
    \begin{equation}\llabel{arg1}
      \textstyle\sup_{n\in \N\cap[N,\infty)}\max\{4\cK\varepsilon^{-1}\gamma_n,2\cK\scrc d^{\tau}(1+\tau)(2\varepsilon^{-1}\gamma_n)^{\tau}\}\leq \frac 18
    \end{equation}
     \argument{\lref{assumption2};\unskip, \eg, \cite[Proposition 5.7.23]{ArBePhi2024}}{that \llabel{argt1} $\R^d\ni\theta\mapsto \smalll(\theta,\bfx)\in \R$ is strongly convex\dott}
    \argument{\lref{argt1};\unskip, \eg, \cite[Corollary 5.7.22]{ArBePhi2024}}{that there exists $\vartheta\in \R^\fd$ which satisfies
    \begin{equation}\llabel{def: vartheta}
        \smalll(\vartheta,\bfx)=\inf_{\theta\in \R^\fd}\smalll(\theta,\bfx).
    \end{equation}}
    \startnewargseq
    \argument{the assumption that $\smalll\in C^{1}(\R^\fd\times O,\R)$; the assumption that $\setX$ is compact}{that
    \begin{equation}\llabel{def: rho}
        \textstyle 1+\sup_{x\in \setX}\|\nabla_\theta\smalll(\vartheta,x)\|<\infty.
    \end{equation}}
    In the following let $\rho\in \R$ satisfy $\rho=1+\sup_{x\in \setX}\|\nabla_\theta\smalll(\vartheta,x)\|$, for every $\alpha\in [0,1)$, $\beta\in (\alpha^2,1)$ let $\fn_{\alpha,\beta}\in \N$, $\eta_{\alpha,\beta},\delta_{\alpha,\beta},\chi_{\alpha,\beta},\cA_3^{\alpha,\beta},\cA_4^{\alpha,\beta},\cA_5^{\alpha,\beta}\in \R$ satisfy
 \begin{equation}\llabel{def: eta}
     \eta_{\alpha,\beta}=\frac{d}{\sqrt{(1-\alpha^2\beta^{-1})(1-\beta)}},
 \end{equation}
  \begin{equation}\llabel{def: fn}
      \fn_{\alpha,\beta}=\ceil{\max\bigl\{\textstyle N,-[\log_{2}(\alpha+2\mathbbm 1_{\{0\}}(\alpha))]^{-1},\log_\beta(\frac{\beta-\alpha^2}{1-\alpha^2})+1\bigr\}},
  \end{equation}
    \begin{equation}\llabel{def: delta}
       \delta_{\alpha,\beta}=\frac{ (\fn_{\alpha,\beta})^{1/2} d}{\sqrt{1-\alpha^2\beta^{-1}}},\qquad \chi_{\alpha,\beta}=\max\biggl\{\frac{\alpha}{1-\alpha},4\biggr\},
    \end{equation}
  \begin{equation}\llabel{def: A3}
\cA_3^{\alpha,\beta}=\frac{\rho^{4/(1+2\tau)}\kappa}{(\chi_{\alpha,\beta} d\varepsilon^{-1}\scrc^2\cK^3)^{2/(1+2\tau)}},
  \end{equation}
  \begin{equation}\llabel{def: A4}
      \cA_4^{\alpha,\beta}=\max\bigl\{1610(1+\gamma_1)\cK^{3/2}\kappa^{-(5+2\tau)/4}(1+\kappa)\scrc\varepsilon^{-3/2}(1+\varepsilon)(1+(\chi_{\alpha,\beta})^{1/2}) d,1\bigr\},
  \end{equation}
  \begin{equation}\llabel{def: A5}
   \textstyle \text{and}\qquad  \cA_5^{\alpha,\beta}=\bigl(4\cK(\cA_4^{\alpha,\beta})^{2}\bigr)^{2/(1-2\tau)},
  \end{equation}
     for every $M\in \N$, $\alpha\in [0,1)$, $\beta\in (\alpha^2,1)$, $\xi\in\R^d$ let $\cA_1^{M,\alpha,\beta,\xi}\in \R$ satisfy
     \begin{equation}\llabel{def: A1}
  \begin{split}
      \cA_1^{M,\alpha,\beta,\xi}&=2\cK( \fn_{\alpha,\beta}\gamma_1\delta_{\alpha,\beta}+\|\xi-\vartheta\|)^2+\frac{d}{2(1-\alpha)} \bigl[ 4\cK(\fn_{\alpha,\beta}\gamma_1\delta_{\alpha,\beta}+\|\xi-\vartheta\|)+\rho\bigr]\gamma_1\delta_{\alpha,\beta},
        \end{split}
  \end{equation}
  let $\cA_2\in \R$ satisfy
  \begin{equation}\llabel{def: A2}
     \textstyle \cA_2=\max\bigl\{\frac{8\cK^2}{\kappa d^2},\frac{\kappa}{2},\frac{\kappa\|\vartheta\|^2}{2},\frac{\rho^2\kappa}{2\cK^2},\kappa\bigl(\frac{\varepsilon}{\cK}\bigr)^{2},\bigl(\frac{1+32\cK^2\scrc^2}{\cK^2\scrc^2}\bigr)^{1/\tau}\kappa\bigr\},
  \end{equation}
  and for every $\alpha\in [0,1)$ let $\cA_6^\alpha\in\R$ satisfy
   \begin{equation}\llabel{def: A6}
    \cA_6^\alpha=\bigl(328000(1-\alpha)^{-1}\gamma_1\cK^3\kappa^{-(1+2\tau)/2}\scrc^2d\varepsilon^{-2}\bigr)^{2/(1-2\tau)}.
  \end{equation}
  \startnewargseq
    \argument{\lref{assumption2};\lref{def: vartheta};\lref{def: rho};\lref{def: fn};\lref{def: delta};\lref{def: A3};\lref{def: A4};\lref{def: A5};\lref{def: A1};\lref{def: A2};\lref{def: A6};\cref{lem: priori bound stochastic Adam non explosion}}{that for every $M\in \N$, $\alpha\in [0,1)$, $\beta\in (\alpha^2,1)$, $\xi\in\R^d$, every
     $(X_{n,m})_{(n,m)\in \N^2}\subseteq \setX$, every $\Theta=(\Theta^{1},\dots,\Theta^{\fd}\allowbreak)\colon\N_0\to\R^d$, every $\bfm=(\bfm^{1},\dots,\bfm^{n})\colon\N_0\to\R^d$, and every $\democrator=(\democrator^{1},\dots,\allowbreak\democrator^{n})\colon\N_0\to\R^\fd$ with the property that for all $n\in \N$, $i\in \{1,2,\dots,\fd\}$ it holds that 
 \begin{equation} 
    \bfm^{i}_{0}= 0,\qquad \bfm^{i}_{n}=\alpha \bfm_{n-1}^{i}+(1-\alpha)\textstyle \frac 1M \sum_{m=1}^M(\frac{\partial}{\partial\theta_i} \smalll)(\Theta_{n-1},X_{n,m}),
    \end{equation}
    \begin{equation}
    \democrator^{i}_{0}= 0,\qquad \democrator_{n}^{i}=\beta \democrator_{n-1}^{i}+(1-\beta)\bigl[\textstyle \frac 1M \sum_{m=1}^M(\frac{\partial}{\partial\theta_i} \smalll)(\Theta_{n-1},X_{n,m})\bigr]^2,
    \end{equation}
    \begin{equation}
\Theta_n=\xi,\qquad \text{and}\qquad\textstyle\Theta_{n}^{i}=\Theta_{n-1}^{i} -\gamma_n\Bigl[\frac{\bfm_{n}^{i}}{1-\alpha^n}\Bigr] \Bigl[\varepsilon+\Bigl[\frac{\democrator_{n}^{i}}{1-\beta^n}\Bigr]^{\nicefrac{1}{2}}\Bigr]^{-1}, 
    \end{equation}
 we have for all $n\in\N$ that
    \begin{equation}\llabel{eqmain}
    \begin{split}
         \| \Theta_n \|^2&\leq 4\kappa^{-1}\max\{1,\cA_1^{M,\alpha,\beta,\xi},\cA_2,\cA_3^{\alpha,\beta},\cA_5^{\alpha,\beta},\cA_6^\alpha\}+2\|\vartheta\|^2\\
        &\leq  4+ 4\kappa^{-1}\cA_1^{M,\alpha,\beta,\xi}+4\kappa^{-1}\cA_2+4\kappa^{-1}\cA_3^{\alpha,\beta}+4\kappa^{-1}\cA_5^{\alpha,\beta}+4\kappa^{-1}\cA_6^\alpha+2\|\vartheta\|^2.
        \end{split}
    \end{equation}}
    \argument{\lref{def: fn};\cref{lem: estimate fn}}{that there exist $c_1,c_2\in \R$ such that for all $\alpha\in [0,1)$, $\beta\in (\alpha^2,1)$, $\varepsilon\in (0,\infty)$ it holds that
    \begin{equation}\llabel{estimate fn}
    \begin{split}
        \fn_{\alpha,\beta}&=\ceil{\max\{\textstyle N,-[\log_{2}(\alpha+2\mathbbm 1_{\{0\}}(\alpha))]^{-1},\log_\beta(\frac{\beta-\alpha^2}{1-\alpha^2})+1\}}\\
       &\leq N-[\log_{2}(\alpha+2\mathbbm 1_{\{0\}}(\alpha))]^{-1}+\max\{4(1-\alpha)^{-1},4(\beta-\alpha^2)^{-1}\}+1\\
       &\leq N+1+c_1(1-\alpha)^{-1}+4(1-\alpha)^{-1}+4(\beta-\alpha^2)^{-1}\\
       &\leq N+1+c_2(1-\alpha)^{-1}+c_2(\beta-\alpha^2)^{-1}.
        \end{split}
    \end{equation}}
    \argument{\lref{estimate fn};\lref{def: delta};the Cauchy-Schwarz inequality}{that there exists $c\in \R$ such that for all $\alpha\in [0,1)$, $\beta\in (\alpha^2,1)$ it holds that
    \begin{equation}\llabel{estimate delta}
    \begin{split}
        \delta_{\alpha,\beta}&\leq d(\fn_{\alpha,\beta})^{1/2}(\beta-\alpha^2)^{-1/2}\leq d\fn_{\alpha,\beta}+d(\beta-\alpha^2)^{-1}\\
        &\leq c+c(1-\alpha)^{-1}+c(\beta-\alpha^2)^{-1}
        \end{split}
    \end{equation}
    \begin{equation}\llabel{estimate chi}
        \text{and}\qquad 4\leq \chi_{\alpha,\beta}\leq c(1-\alpha)^{-1}.
    \end{equation}}
    \argument{\lref{estimate chi};\lref{estimate fn};Young inequality;the fact that for all $x_1,x_2,x_3,x_4\in [0,\infty)$ it holds that $(x_1+x_2+x_3+x_4)^3\leq 16\sum_{i=1}^4(x_i)^3$;the fact that for all $x_1,x_2,x_3,x_4\in [0,\infty)$ it holds that $(x_1+x_2+x_3+x_4)^4\leq 64\sum_{i=1}^4(x_i)^4$}{that there exists $c\in \R$ such that for all $\alpha\in [0,1)$, $\beta\in (\alpha^2,1)$ it holds that
    \begin{equation}\llabel{estimate fn delta}
       \max\{(\fn_{\alpha,\beta}\delta_{\alpha,\beta})^2, \fn_{\alpha,\beta}(\delta_{\alpha,\beta})^2,(\delta_{\alpha,\beta})^2\}\leq c+c(1-\alpha)^{-4}+c(\beta-\alpha^2)^{-4}.
    \end{equation}}
    \argument{\lref{def: A1};\lref{estimate delta};\lref{estimate fn delta};}{that there exist $c_1,c_2,c_3\in \R$ such that for all $M\in \N$, $\alpha\in [0,1)$, $\beta\in (\alpha^2,1)$, $\xi\in\R^d$ it holds that
    \begin{equation}\llabel{estimate cA1}
    \begin{split}
       & \textstyle\cA_1^{M,\alpha,\beta,\xi}\\
       &\textstyle\leq c_1(1-\alpha)^{-1}\|\xi-\vartheta\|^2+c_1(\fn_{\alpha,\beta}\delta_{\alpha,\beta})^2+c_1(1-\alpha)^{-1}\|\xi-\vartheta\|+c_1\fn_{\alpha,\beta}(\delta_{\alpha,\beta})^2+c_1(\delta_{\alpha,\beta})^2+c_1\\
        &\leq c_2(1-\alpha)^{-1}\|\xi-\vartheta\|^2+c_2+c_2(1-\alpha)^{-4}+c_2(\beta-\alpha^2)^{-4}\\
        &\leq c_3(1-\alpha)^{-1}\|\xi\|^2+c_3+c_3(1-\alpha)^{-4}+c_3(\beta-\alpha^2)^{-4}.
        \end{split}
    \end{equation}}
    \argument{\lref{def: A3};\lref{estimate chi}}{that there exists $c\in \R$ such that for all $\alpha\in [0,1)$, $\beta\in (\alpha^2,1)$ it holds that
    \begin{equation}\llabel{estimate cA3}
    \begin{split}
         \cA_3^{\alpha,\beta}\leq \frac{\rho^{4/(1+2\tau)}\kappa}{(4 d\varepsilon^{-1}\scrc^2\cK^3)^{2/(1+2\tau)}}\leq c.
        \end{split}
    \end{equation}}
     \argument{\lref{def: A4};\lref{estimate chi}}{that there exist $c_1,c_2\in \R$ such that for all $\alpha\in [0,1)$, $\beta\in (\alpha^2,1)$ it holds that
    \begin{equation}\llabel{estimate cA4}
    \begin{split}
        \cA_4^{\alpha,\beta}\leq c_1+c_1(\chi_{\alpha,\beta})^{1/2}\leq c_2+c_2(1-\alpha)^{-1/2}.
        \end{split}
    \end{equation}}
    \argument{\lref{def: A5};\lref{estimate cA4}; the fact that for all $x,y\in[0,\infty)$, $r\in [2,\infty)$ it holds that $(x+y)^r\leq 2^r(x^r+y^r)$}{that there exist $c_1,c_2,c_3\in \R$ such that for all $\alpha\in [0,1)$, $\beta\in (\alpha^2,1)$ it holds that
    \begin{equation}\llabel{estimate cA5}
    \begin{split}
        \cA_5^{\alpha,\beta}&\leq c_1(\cA_4^{\alpha,\beta})^{4/(1-2\tau)}\leq c_1  (c_2+c_2(1-\alpha)^{-1/2})^{4/(1-2\tau)}\leq c_3+c_3(1-\alpha)^{-2/(1-2\tau)}.
        \end{split}
    \end{equation}}
    \argument{\lref{def: A6};the fact that $1-\tau\geq \frac 12$}{that there exists $c\in (0,\infty)$ such that for all $\alpha\in [0,1)$ it holds that
    \begin{equation}\llabel{estimate cA6}
    \begin{split}
        \cA_6^\alpha\leq c(1-\alpha)^{-2/(1-2\tau)}.
        \end{split}
    \end{equation}}
    \argument{\lref{eqmain};\lref{estimate cA1};\lref{estimate cA3};\lref{estimate cA5};\lref{estimate cA6};}{that there exists $c\in \R$ such that for for every $M\in \N$, $\alpha\in [0,1)$, $\beta\in (\alpha^2,1)$, $\xi\in\R^d$, every
     $(X_{n,m})_{(n,m)\in \N^2}\subseteq \setX$, every $\Theta=(\Theta^{1},\dots,\Theta^{\fd}\allowbreak)\colon\N_0\to\R^d$, every $\bfm=(\bfm^{1},\dots,\bfm^{n})\colon\N_0\to\R^d$, and every $\democrator=(\democrator^{1},\dots,\allowbreak\democrator^{n})\colon\N_0\to\R^\fd$ with the property that for all $n\in \N$, $i\in \{1,2,\dots,\fd\}$ it holds that 
 \begin{equation} 
    \bfm^{i}_{0}= 0,\qquad \bfm^{i}_{n}=\alpha \bfm_{n-1}^{i}+(1-\alpha)\textstyle \frac 1M \sum_{m=1}^M(\frac{\partial}{\partial\theta_i} \smalll)(\Theta_{n-1},X_{n,m}),
    \end{equation}
    \begin{equation}
    \democrator^{i}_{0}= 0,\qquad \democrator_{n}^{i}=\beta \democrator_{n-1}^{i}+(1-\beta)\bigl[\textstyle \frac 1M \sum_{m=1}^M(\frac{\partial}{\partial\theta_i} \smalll)(\Theta_{n-1},X_{n,m})\bigr]^2,
    \end{equation}
    \begin{equation}
\Theta_n=\xi,\qquad \text{and}\qquad\textstyle\Theta_{n}^{i}=\Theta_{n-1}^{i} -\gamma_n\Bigl[\frac{\bfm_{n}^{i}}{1-\alpha^n}\Bigr] \Bigl[\varepsilon+\Bigl[\frac{\democrator_{n}^{i}}{1-\beta^n}\Bigr]^{\nicefrac{1}{2}}\Bigr]^{-1}, 
    \end{equation}
 we have for all $n\in\N$ that
    \begin{equation}\llabel{eq10}
    \begin{split}
         \| \Theta_n\|^2&\leq c(1-\alpha)^{-1}\| \xi \|^2+c+c(1-\alpha)^{-4}+c(1-\alpha)^{-2/(1-2\tau)}+c(\beta-\alpha^2)^{-4}\\
         &=c(1-\alpha)^{-1}\| \Theta_0 \|^2+c+c(1-\alpha)^{-4}+c(1-\alpha)^{-2/(1-2\tau)}+c(\beta-\alpha^2)^{-4}.
        \end{split}
    \end{equation}}
    \argument{\lref{eq10};\lref{arg1};the fact that for all $x_1,x_2,x_3,x_4\in [0,\infty)$ it holds that $[(x_1)^2+(x_2)^2+(x_3)^2+(x_4)^2]^{1/2}\leq \sum_{i=1}^4x_i$}{\lref{conclude}\dott}
\end{aproof}
\newcommand{\openset}{O}
\subsection{Uniform a priori bounds for Adam for quadratic example SOPs}\label{subsec: Example 2}
\renewcommand{\setX}{V}
 \begin{tcolorbox}[colback=white!95!gray,
                  colframe=black,
                  boxrule=0.5pt,
                  sharp corners,
                  enhanced,
                  breakable,
                 ]
\begin{athm}{lemma}{lem: verify 2}
    Let $d,\dimX,
\numofsum\in \N$, $f\in C^1(\R^\dimX,\R^d)$, let $\openset\subseteq\R^\dimX$ be open, let $\setX\subseteq O$ be compact, for every $i\in \{1,2,\dots,
\numofsum\}$ let $g_i\in C^1(\R^d,\R)$ be convex and let $h_i\in C^1(\R^\dimX,\R)$ satisfy for all $x\in \setX$ that $h_i(x)\geq 0$, let $r,L\in [0,\infty)$ satisfy for all $\theta,\vartheta\in \R^d$, $i\in \{1,2,\dots,
\numofsum\}$ that
    \begin{equation}\llabel{assume}
       \max \biggl\{\frac{ \|g_i(\theta)-g_i(\vartheta)\|}{1+\|\theta\|^r+\|\vartheta\|^{r}},\|(\nabla g_i)(\theta)-(\nabla g_i)(\vartheta)\|\biggr\}\leq L\|\theta-\vartheta\|,
    \end{equation}  
    let $A\in \R^{d\times d}$ be invertible,
    and let $\smalll\colon\R^\fd\times\openset\to\R$ satisfy for all $\theta\in \R^\fd$, $x\in \openset$ that
    \begin{equation}\llabel{def: loss}
        \smalll(\theta,x)=\| A \theta - f( x ) \|^2 
    + 
    \sum\limits_{ i = 1 }^\numofsum h_i( x ) g_i(\theta) .
    \end{equation}
 Then there exist $\kappa,\cK\in (0,\infty)$ such that
 \begin{enumerate}[label=(\roman*)]
     \item \label{item 1: verify 2} it holds for all $\theta,\vartheta\in \R^\fd$, $x\in \setX$ that $\|(\nabla_\theta\smalll)(\theta,x)-(\nabla_\theta\smalll)(\vartheta,x)\|\leq \cK\|\theta-\vartheta\|$,
     \item \label{item 2: verify 2} it holds for all $\theta,\vartheta\in \R^\fd$, $x\in \setX$ that $\|(\nabla_x\smalll)(\theta,x)-(\nabla_x\smalll)(\vartheta,x)\|\leq \cK\|\theta-\vartheta\|(1+\|\theta\|^{r}+\|\vartheta\|^{r})$, and
     \item \label{item 3: verify 2} it holds for all $\theta,\vartheta\in \R^\fd$, $x\in \setX$ that $ \spro{ \theta-\vartheta, (\nabla_\theta\smalll)(\theta,x)- (\nabla_\theta\smalll)(\vartheta,x)}\geq \kappa\|\theta-\vartheta\|^2$.
 \end{enumerate} 
\end{athm}
\end{tcolorbox}
\begin{aproof}
    \argument{\lref{def: loss};}{that for all $\theta\in \R^\fd$, $x\in \openset$ it holds that
    \begin{equation}\llabel{eq1}
        (\nabla_\theta\smalll)(\theta,x)=2A^{*}(A\theta-f(x))+\sum_{i=1}^\numofsum(\nabla g_i)(\theta)h_i(x)
    \end{equation}
    and
    \begin{equation}\llabel{eq2}
        (\nabla_x\smalll)(\theta,x)=-2(f'(x))^*(A\theta-f(x))+\sum_{i=1}^\numofsum g_i(\theta) (\nabla h_i)(x).
    \end{equation}}
    \argument{\lref{eq2};\lref{assume};the Cauchy-Schwarz inequality;}{that for all $\theta,\vartheta\in \R^{\fd}$, $x\in \setX$ it holds that
    \begin{equation}\llabel{evidence 1}
    \begin{split}
       &\textstyle \|(\nabla_\theta\smalll)(\theta,x)-(\nabla_\theta)\smalll(\vartheta,x)\|\\
       &\textstyle=\bigl\|2A^{*}A(\theta-\vartheta)+\sum_{i=1}^\numofsum[(\nabla g_i)(\theta)h_i(x)-(\nabla g_i)(\vartheta)h_i(x)]\bigr\|\\
        &\textstyle\leq 2\|A^{*}A(\theta-\vartheta)\|+\sum_{i=1}^
\numofsum\|(\nabla g_i)(\theta)h_i(x)-(\nabla g_i)(\vartheta)h_i(x)\|\\
        &\textstyle\leq 2\|A^{*}A\|\|\theta-\vartheta\|+\sum_{i=1}^
\numofsum\bigl[\sup_{y\in \setX}\max_{j\in \{1,2,\dots,\numofsum\}}h_j(y)\bigr]\|(\nabla g_i)(\theta)-(\nabla g_i)(\vartheta)\|\\
        &\textstyle \leq 2\|A^{*}A\|\|\theta-\vartheta\|+L\numofsum\bigl[\sup_{y\in \setX}\max_{i\in \{1,2,\dots,\numofsum\}}h_i(y)\bigr]\|\theta-\vartheta\|\\
        &\textstyle \leq \|\theta-\vartheta\|\bigl(2\|A^{*}A\|+L\numofsum\bigl[\sup_{y\in \setX}\max_{i\in \{1,2,\dots,\numofsum\}}h_i(y)\bigr]\bigr).
        \end{split}
    \end{equation}}
    \argument{\lref{assume};\lref{eq2};the Cauchy-Schwarz inequality}{for all $\theta,\vartheta\in \R^\fd$, $x\in \setX$ that
    \begin{equation}\llabel{evidence 2}
    \begin{split}
              & \textstyle \|(\nabla_x\smalll)(\theta,x)-(\nabla_x\smalll)(\vartheta,x)\|\\
              &\textstyle= \bigl\|-2(f'(x))^* A(\theta-\vartheta)+\sum_{i=1}^\numofsum [g_i(\theta) (\nabla h_i)(x)-g_i(\vartheta) (\nabla h_i)(x)]\bigr\|\\
              &\textstyle \leq 2\bigl[\sup_{x\in \setX}\|(f'(x))^* A\|\bigr]\|\theta-\vartheta\|+ L\sum_{i=1}^\numofsum \|(\nabla h_i)(x)\|\|\theta-\vartheta\|(1+ \|\theta\|^r+\|\vartheta\|^r)\\
              &\textstyle \leq \|\theta-\vartheta\|\bigl(2\bigl[\sup_{x\in \setX}\|(f'(x))^* A\|\bigr]+L\numofsum\bigl[\sup_{y\in \setX}\max_{i\in \{1,2,\dots,\numofsum\}}\|(\nabla h_i)(y)\|\bigr](1+ \|\theta\|^r+\|\vartheta\|^r)\bigr).
               \end{split}
    \end{equation}}
    \argument{the assumption that for all $i\in \{1,2,\dots,\numofsum\}$ it holds that $g_i$ is convex;\cite[Lemma 5.7.3]{ArBePhi2024}}{that for all $\theta,\vartheta\in \R^d$, $i\in \{1,2,\dots,\numofsum\}$ it holds that
    \begin{equation}\llabel{evd3.1}
        \spro{\theta-\vartheta,(\nabla g_i)(\theta)-(\nabla g_i)(\vartheta)}\geq 0.
    \end{equation}}
    \argument{\lref{evd3.1};\lref{eq1};the fact that $\operatorname{det}(A)\neq 0$; the assumption that for all $i\in \{1,2,\dots,\numofsum\}$, $x\in \setX$ it holds that $h_i(x)\geq 0$; the Cauchy-Schwarz inequality;}{that for all $\theta,\vartheta\in \R^\fd$, $x\in \setX$ it holds that
    \begin{equation}\llabel{evidence 3}
    \begin{split}
       & \spro{ \theta-\vartheta, (\nabla_\theta\smalll)(\theta,x)- (\nabla_\vartheta\smalll)(\vartheta,x)}\\
       &=\spro{\theta-\vartheta,2A^{*}A(\theta-\vartheta)}+\sum_{i=1}^\numofsum \spro{\theta-\vartheta,(\nabla g_i)(\theta)h_i(x)-(\nabla g_i)(\vartheta)h_i(x)}\\
       &\geq 2\|A(\theta-\vartheta)\|^2\geq 2(\|A^{-1}\|)^{-2}\|\theta-\vartheta\|^2.
        \end{split}
    \end{equation}}
    \argument{\lref{evidence 1};\lref{evidence 2};\lref{evidence 3};the fact that for all $i\in\{1,2,\dots,\numofsum\}$ it holds that $h_i$ is continuous; the fact that for all $i\in\{1,2,\dots,\numofsum\}$ it holds that $\nabla h_i$ is continuous; the fact that $f'$ is continuous}{items \ref{item 1: verify 2}, \ref{item 2: verify 2}, and \ref{item 3: verify 2}\dott}
\end{aproof}
 \begin{tcolorbox}[colback=white!95!gray,
                  colframe=black,
                  boxrule=0.5pt,
                  sharp corners,
                  enhanced,
                  breakable,
                 ]
    \begin{athm}{lemma}{lem: verify verify 2}
    Let $d,\dimofsum\in \N$, $r\in [\nicefrac 12,1)$, $A\in \R^{\dimofsum\times d}$, $\beta\in (0,\infty)$ and let $g\colon \R^d\to\R$ satisfy for all $\theta\in \R^d$ that
    \begin{equation}\llabel{def: loss}
        g(\theta)=(\|A\theta\|^2+\beta)^{r}.
    \end{equation}
    Then
    \begin{enumerate}[label=(\roman*)]
        \item \label{item 1: verify verify 2} there exists $L\in \R$ such that for all $\theta,\vartheta\in \R^d$ it holds that
    \begin{equation}\llabel{conclude}
       \frac{ \|g(\theta)-g(\vartheta)\|}{1+\|\theta\|^{2r-1}+\|\vartheta\|^{2r-1}}+\|(\nabla g)(\theta)-(\nabla g)(\vartheta)\|\leq L\|\theta-\vartheta\|
    \end{equation}  
    and
    \item \label{item 2: verify verify 2} it holds that $g$ is convex.
    \end{enumerate}
\end{athm}
\end{tcolorbox}
\begin{aproof}
Throughout this proof let $I=(I_{m,n})_{(m,n)\in \{1,2,\dots,\d\}^2}\in \R^{d\times d}$ satisfy for all $m,n\in \{1,2,\dots,d\}$ that $I_{m,n}=\mathbbm 1_{\{m\}}(n)$.
     \argument{\lref{def: loss};}{that for all $\theta\in \R^\fd$ it holds that
    \begin{equation}\llabel{eq1}
        (\nabla g)(\theta)= 2rA^*A\theta (\|A\theta\|^2+\beta)^{r-1}. 
    \end{equation}
    and 
    \begin{equation}\llabel{eq2}
        (\operatorname{Hess} g)(\theta)=2rA^*A (\|A\theta\|^2+\beta)^{r-1} I_d+4r(r-1)(A^*A\theta)(A^*A\theta)^* (\|A\theta\|^2+\beta)^{r-2}.
    \end{equation}}
    \argument{\lref{eq2};the fact that $r\geq \frac{1}{2}$}{that for all $\theta\in \R^d$ it holds that
    \begin{equation}\llabel{eq3}
        \| (\nabla g)(\theta)\|=2r\|A^*A\theta\|(\|A\theta\|^2+\beta)^{r-1}\leq 2r\|A^*\| \|A\theta\|^{2r-1}\leq 2r\|A^*\|\|A\|^{2r-1}\|\theta\|^{2r-1}.
    \end{equation}}
    \argument{\lref{eq3};the fundametal theorem of calculus}{that there exists $L\in \R$ such that for all $\theta,\vartheta\in \R^d$ that
    \begin{equation}\llabel{evd1}
        \|g(\theta)-g(\vartheta)\|\leq L\|\theta-\vartheta\| (1+\|\theta\|^{2r-1}+\|\vartheta\|^{2r-1})
    \end{equation}}
    \argument{\lref{eq2};the triangle inequality; the fact that $r<1$}{that for all $\theta\in \R^d$ it holds that
    \begin{equation}\llabel{eq4}
    \begin{split}
        \|(\operatorname{Hess} g)(\theta)\|
       &\leq 2r\|A^*A\| (\|A\theta\|^2+\beta)^{r-1} +4r(1-r)\|A^*\|^2 (\|A\theta\|^2+\beta)^{r-2}\|A\theta\|^2\\
      & \leq 2r\|A^*A\|\beta^{r-1}+4r(1-r)\beta^{r-1}\|A^*\|^2.
        \end{split}
    \end{equation}}
    \argument{\lref{eq4};the fundamental theorem of calculus}{that there exists $L\in \R$ such that for all $\theta,\vartheta\in \R^d$ that
    \begin{equation}\llabel{evd2}
        \|(\nabla g)(\theta)-(\nabla g)(\vartheta)\|\leq L\|\theta-\vartheta\|.
    \end{equation}}
    \argument{\lref{evd1};\lref{evd2}}{\cref{item 1: verify verify 2}\dott}
    \startnewargseq
    \argument{\lref{eq2};the fact that $\frac 12\leq r<1$;the Cauchy-Schwarz inequality}{that for all $\theta,x\in \R^d$ it holds that
    \begin{equation}\llabel{eq5}
    \begin{split}
       x^* (\operatorname{Hess} g)(\theta) x&=  2r\|Ax\|^2(\|A\theta\|^2+\beta)^{r-1}-4r(1-r)|\spro{A\theta,Ax}|^2(\|A\theta\|^2+\beta)^{r-2}\\
       &\geq  2r \|Ax\|^2(\|A\theta\|^2+\beta)^{r-1}-4r(1-r)\|Ax\|^2\|A\theta\|^2(\|A\theta\|^2+\beta)^{-1}(\|A\theta\|^2+\beta)^{r-1}\\
       &\geq 2r \|Ax\|^2(\|A\theta\|^2+\beta)^{r-1}-4r(1-r)\|Ax\|^2(\|A\theta\|^2+\beta)^{r-1}\geq 0.
       \end{split}
    \end{equation}}
    \argument{\lref{eq5};\cite[Proposition 5.7.24]{ArBePhi2024}}{\cref{item 1: verify verify 2}\dott}
    \end{aproof}
     \begin{tcolorbox}[colback=white!95!gray,
                  colframe=black,
                  boxrule=0.5pt,
                  sharp corners,
                  enhanced,
                  breakable,
                 ]
    \begin{athm}{lemma}{cor: verify main}
     Let $\fd,\dimX,\numofsum,\dimofsum,p\in \N$, $f_0\in C^p(\R^\dimX,\R^d)$, let $\setX\subseteq \R^\dimX$ be compact, let $A_0\in \R^{d\times d}$ be invertible, for every $i\in\N$ let $r_i\in [\nicefrac{1}{2},\nicefrac{3}{4})$, $A_i\in \R^{\dimofsum\times d}$, $\mu_i\in (0,\infty)$ and let $f_i\in C^p(\R^\dimX,\R)$ satisfy for all $x\in \setX$ that $f_i(x)\geq 0$, let $\fr\in \R$ satisfy $\fr=2\max\{r_1,r_2,\dots,r_\numofsum\}-1$, let $\smalll=(\smalll(\theta,x))_{(\theta,x)\in \R^\fd\times \R^\dimX}\allowbreak\colon \R^\fd\times \R^\dimX\to\R$ satisfy for all $\theta\in \R^\fd$, $x\in \R^\dimX$ that 
     \begin{equation}\llabel{def: loss}
        \textstyle \smalll(\theta,x)=\| A_0 \theta - f_0( x ) \|^2 
    + 
    \sum\nolimits_{ i = 1 }^\numofsum  ( \| A_i \theta \|^2 + \mu_i )^{ r_i }f_i( x),
    \end{equation}
 Then 
 \begin{enumerate}[label=(\roman*)]
     \item \label{item 1: verify main} it holds that $\smalll\in C^p(\R^d\times\R^\dimX,\R)$ and
     \item \label{item 2: verify main} there exist $\kappa,\cK\in (0,\infty)$ such that for all $\theta,\vartheta\in \R^\fd$, $x\in \setX$ it holds that
 \begin{equation}\llabel{conclude 1}
     \|(\nabla_\theta\smalll)(\theta,x)-(\nabla_\theta\smalll)(\vartheta,x)\|\leq \cK\|\theta-\vartheta\|,\quad \spro{ \theta-\vartheta, (\nabla_\theta\smalll)(\theta,x)- (\nabla_\theta\smalll)(\vartheta,x)}\geq \kappa\|\theta-\vartheta\|^2,
 \end{equation}
  \begin{equation}\llabel{conclude 2}
     \text{and}\qquad \|(\nabla_x\smalll)(\theta,x)-(\nabla_x\smalll)(\vartheta,x)\|\leq \cK\|\theta-\vartheta\|(1+\|\theta\|^{\fr}+\|\vartheta\|^{\fr}).
 \end{equation}
 \end{enumerate}
\end{athm}
\end{tcolorbox}
\begin{aproof}
\argument{the fact that $(\R^d\times\R^d\ni(\theta,x)\mapsto \| A_0 \theta - x\|^2\in \R)\in C^\infty(\R^d\times\R^d,\R)$; the assumption that $f_0\in C^p(\R^\dimX,\R^d)$; the fact that for all $i\in \{1,2,\dots,\numofsum\}$ it holds that $(\R^d\ni \theta \mapsto\| A_i \theta \|^2\in \R)\in C^\infty(\R^d,\R)$; the fact that for all $i\in \{1,2,\dots,\numofsum\}$ it holds that $([0,\infty)\ni x\mapsto (x+\mu_i)^{r_i}\in \R)\in C^\infty([0,\infty),\R)$; the assumption that for all $i\in \{1,2,\dots,\numofsum\}$ it holds that $f_i\in C^p(\R^\dimX,\R)$; the chain rule}{\cref{item 1: verify main}\dott}
\startnewargseq
    \argument{\cref{lem: verify 2};\cref{lem: verify verify 2}}{\cref{item 2: verify main}\dott}
\end{aproof}
\renewcommand{\setX}{U}
 \begin{tcolorbox}[colback=white!95!gray,
                  colframe=black,
                  boxrule=0.5pt,
                  sharp corners,
                  enhanced,
                  breakable,
                 ]
 \begin{athm}{lemma}{lem: reconstruct}
      Let $\fd,\dimX,\numofsum,\dimofsum\in \N$, $p,\varepsilon\in (0,\infty)$, let $A_0\in \R^{d\times d}$ be invertible, let $f_0\colon \R^\dimX\to\R^d$ be measurable and locally bounded, for every $i\in\N$ let $A_i\in \R^{\dimofsum\times d}$, $r_i\in [\nicefrac{1}{2},\nicefrac{3}{4})$, $\mu_i\in (0,\infty)$, $\gamma_i\in \R$ and let $f_i\colon \R^\dimX\to[0,\infty)$ be measurable and locally bounded, let $\smalll=(\smalll(\theta,x))_{(\theta,x)\in \R^\fd\times \R^\dimX}\allowbreak \colon \R^\fd\times \R^\dimX\to\R$ satisfy for all $\theta\in \R^\fd$, $x\in \R^\dimX$ that 
     \begin{equation}\llabel{def: loss}
       \textstyle \smalll(\theta,x)=\| A_0 \theta - f_0( x ) \|^2 
    + 
    \sum_{ i = 1 }^\numofsum ( \| A_i \theta \|^2 + \mu_i )^{ r_i }f_i( x ) ,
    \end{equation}
     for every $n,m\in \N$ let $X_{n,m}\in [-p,p]^\dimX$, $\bbX_{n,m}\in \R^{d+\numofsum}$ satisfy
         $\bbX_{n,m}=(f_0(X_{n,m}),f_1(X_{n,m}),\dots,f_\numofsum(X_{n,m}))$,
     for every $M\in \N$, $\para=(\para_1,\para_2)\in (0,1)^2$ let 
    $
\mandV{k}{M}{\para}{n}=(\mandVcom{k}{M}{\para}{n}{1},\dots,\allowbreak \mandVcom{k}{M}{\para}{n}{\fd})\allowbreak \in\R^\fd$, $(k,n)\in (\N_0)^2$, satisfy for all $k\in \{1,2\}$, $n\in \N$, $i\in \{1,2,\dots,\fd\}$ that
    \begin{equation}\llabel{def: bbV}
    \mandV{k}{M}{\para}{0}=0,\qquad \mandVcom{k}{M}{\para}{n}{i}= \beta_k \mandVcom{k}{M}{\para}{n-1}{i}+(1-\beta_k)\bigl[\textstyle \frac 1M \sum_{m=1}^M(\nabla_{\theta_i} \smalll)(\mandV{0}{M}{\beta}{n-1},X_{n,m})\bigr]^k,
        \end{equation}
        \begin{equation}\llabel{def: Theta}
    \text{and}\qquad \mandVcom{0}{M}{\para}{n}{i}=\mandVcom{0}{M}{\para}{n-1}{i} -\gamma_n [1-(\para_1)^n]^{-1}\bigl[\varepsilon+\bigl[[1-(\para_2)^n]^{-1}\mandVcom{2}{M}{\para}{n}{i}\bigr]^{\nicefrac{1}{2}}\bigr]^{-1}\mandVcom{1}{M}{\para}{n}{i},
     \end{equation}
let $\cL=(\cL(\theta,x))_{(\theta,x)\in \R^\fd\times \R^{d+\numofsum}}\colon \R^{d}\times \R^{d+\numofsum}\to\R$ satisfy for all $\theta\in \R^d$, $x=(x_1,\dots,x_{d+\numofsum})\in \R^{d+\numofsum}$ that
\begin{equation}\llabel{def: cL1}
    \cL(\theta,x)=\textstyle\| A_0 \theta - (x_1,x_2,\ldots,x_d) \|^2 
    + 
    \sum\nolimits_{ i = 1 }^\numofsum  ( \| A_i \theta \|^2 + \mu_i )^{ r_i }x_{i+d},
\end{equation}
and let $\fr\in \R$ satisfy $\fr=2\max\{r_1,r_2,\dots,r_\numofsum\}-1$.
 Then 
 \begin{enumerate}[label=(\roman*)]
 \item \llabel{item 1} it holds that $0\leq \fr<\nicefrac 12$, 
     \item \llabel{item 2} there exist $c,\kappa,\cK\in (0,\infty)$ which satisfy for all $a,b\in \R^\fd$, $x\in [-c,c]^{d}\times[0,c]^\numofsum$, $n,m\in \N$ that
 \begin{equation}
     \|(\nabla_\theta\cL)(a,x)-(\nabla_\theta\cL)(b,x)\|\leq \cK\|a-b\|,\quad \spro{ a-b, (\nabla_\theta\cL)(a,x)- (\nabla_\theta\cL)(b,x)}\geq \kappa\|a-b\|^2,
 \end{equation}
  \begin{equation}
     \bbX_{n,m}\in [-c,c]^d\times[0,c]^\numofsum,\quad\text{and}\quad \|(\nabla_x\cL)(a,x)-(\nabla_x\cL)(b,x)\|\leq \cK\|a-b\|(1+\|a\|^{\fr}+\|b\|^{\fr}),
 \end{equation}
 \item \llabel{item 3} it holds for all $k\in \{1,2\}$, $M,n\in \N$, $i\in\{1,2,\dots,d\}$, $\beta=(\beta_1,\beta_2)\in (0,1)^2$ that
 \begin{equation}
    \mandV{k}{M}{\para}{0}=0,\qquad \mandVcom{k}{M}{\para}{n}{i}= \beta_k \mandVcom{k}{M}{\para}{n-1}{i}+(1-\beta_k)\bigl[\textstyle \frac 1M \sum_{m=1}^M(\nabla_{\theta_i} \cL)(\mandV{0}{M}{\beta}{n-1},\bbX_{n,m})\bigr]^k,
        \end{equation}
        \begin{equation}
 \text{and}\qquad\mandVcom{0}{M}{\para}{n}{i}=\mandVcom{0}{M}{\para}{n-1}{i} -\gamma_n [1-(\para_1)^n]^{-1}\bigl[\varepsilon+\bigl[[1-(\para_2)^n]^{-1}\mandVcom{2}{M}{\para}{n}{i}\bigr]^{\nicefrac{1}{2}}\bigr]^{-1}\mandVcom{1}{M}{\para}{n}{i}.
     \end{equation}
 \end{enumerate}
\end{athm}
\end{tcolorbox}
\begin{aproof}
     Throughout this proof let $F_{i}\colon \R^{d+\numofsum}\to\R^{d-\min\{i,1\}(d-1)}$, $i\in \{0,1,\dots,\numofsum\}$, satisfy for all $i\in \{1,2,\dots,\numofsum\}$, $x=(x_1,\dots,x_{d+\numofsum})\in \R^{d+\numofsum}$ that
\begin{equation}\llabel{def: F}
    F_0(x)=(x_1,x_2,\dots,x_d)\qqandqq F_i(x)=x_{i+d},
\end{equation}
and for every $n,m\in\N$ let $\bfX_{n,m}=(\bfX_{n,m}^1,\dots,\bfX_{n,m}^{d+\numofsum})\in \R^{d+\numofsum}$ satisfy $\bfX_{n,m}=\bbX_{n,m}$.
\argument{the fact that for all $i\in \{1,2,\dots,\numofsum\}$ it holds that $\frac 12\leq r_i<\frac 34$}{\lref{item 1}\dott}
\startnewargseq
\argument{the fact that for all $i\in\{0,1,\dots,\numofsum\}$ it holds that $f_i$ is locally bounded; the fact that $\setX$ is bounded}{that there exists $c\in (0,\infty)$ which satisfies for all $i\in\{0,1,\dots,\numofsum\}$ that
\begin{equation}\llabel{def: c}
    \cup_{x\in [-p,p]^\dimX}\{f_i(x)\}\subseteq [-c,c]^{d-\min\{1,i\}(d-1)}.
\end{equation}}
\startnewargseq
\argument{\lref{def: c};the fact that for all $x\in\R^\dimX$, $i\in \{1,2,\dots,\numofsum\}$ it holds that $f_i(x)\geq 0$;}{that for all $i\in \{1,2,\dots,\numofsum\}$ it holds that 
\begin{equation}\llabel{arg2'}
    \cup_{x\in[-p,p]^\dimX}\{f_i(x)\}\subseteq [0,c].
\end{equation}}
\argument{\lref{arg2'};\lref{def: c}}{that
\begin{equation}\llabel{def: V}
     \cup_{x\in [-p,p]^\dimX}\{(f_0(x),f_1(x),\ldots,f_\numofsum(x))\}\subseteq [-c,c]^d\times [0,c]^\numofsum.
\end{equation}}
\argument{assumption that for all $n,m\in \N$ it holds that $\bbX_{n,m}=(f_0(X_{n,m}),f_1(X_{n,m}),\dots,\allowbreak f_\numofsum(X_{n,m}))$}{that for all $n,m\in \N$ it holds that
\begin{equation}\llabel{def: bbX}
    \bfX_{n,m}=(f_0(X_{n,m}),f_1(X_{n,m}),\dots,f_\numofsum(X_{n,m})).
\end{equation}}
\argument{\lref{def: V};\lref{def: bbX};the fact that for all $n,m\in \N$ it holds that $X_{n,m}\in [-p,p]^\dimX$}{that for all $n,m\in \N$ it holds that
\begin{equation}\llabel{evd1}
    \bfX_{n,m}\in [-c,c]^d\times[0,c]^\numofsum.
\end{equation}}
\argument{\lref{def: F};}{that for all $x\in [-c,c]^d\times [0,c]^\numofsum$, $i\in \{1,2,\dots,\numofsum\}$ it holds that
\begin{equation}\llabel{arg2}
    F_i(x)\geq 0.
\end{equation}}
\argument{\lref{def: cL1};\lref{def: F}}{that for all $\theta\in \R^d$, $x\in \R^{d+\numofsum}$ it holds that
\begin{equation}\llabel{def: cL}
     \cL(\theta,x)=\textstyle\| A_0 \theta - F_0(x) \|^2 
    + 
    \sum\nolimits_{ i = 1 }^\numofsum  ( \| A_i \theta \|^2 + \mu_i )^{ r_i }F_i(x).
\end{equation}}
\argument{\lref{def: cL};\lref{arg2};the fact that $F_0\in C^2(\R^{d+\numofsum},\R^d)$; the fact that for all $i\in \{1,2,\dots,\numofsum\}$ it holds that $F_i\in C^2(\R^{d+\numofsum},\R)$; \cref{cor: verify main} (applied with $\dimX\curvearrowleft d+\numofsum$, $p\curvearrowleft 2$, $f_0\curvearrowleft F_0$, $(f_i)_{i\in \{1,2,\dots,\numofsum\}}\curvearrowleft (F_i)_{i\in \{1,2,\dots,\numofsum\}}$, $V\curvearrowleft [-c,c]^d\times [0,c]^\numofsum$ in the notation of \cref{cor: verify main})}{that
\begin{enumerate}[label=(\Roman*)]
  \item \llabel{Item 1} it holds that $\cL\in C^2(\R^d\times\R^{d+\numofsum},\R)$ and
     \item \llabel{Item 2} there exist $\kappa,\cK\in (0,\infty)$ which satisfy for all $a,b\in \R^\fd$, $x\in [-c,c]^d\times [0,c]^\numofsum$ that
 \begin{equation}\llabel{conclude 1}
     \|(\nabla_\theta\cL)(a,x)-(\nabla_\theta\cL)(b,x)\|\leq \cK\|a-b\|,\quad \spro{ a-b, (\nabla_\theta\cL)(a,x)- (\nabla_\theta\cL)(b,x)}\geq \kappa\|a-b\|^2,
 \end{equation}
  \begin{equation}\llabel{conclude 2}
     \text{and}\qquad \|(\nabla_x\cL)(a,x)-(\nabla_x\cL)(b,x)\|\leq \cK\|a-b\|(1+\|a\|^{\fr}+\|b\|^{\fr}).
 \end{equation}
\end{enumerate}}
\startnewargseq
 \argument{\lref{def: loss};\lref{def: F};\lref{def: bbX};\lref{def: cL};}{that for all $\theta\in \R^d$, $n,m\in \N$ it holds that
 \begin{equation}\llabel{eq2}
 \begin{split}
     (\nabla_\theta\smalll)(\theta,X_{n,m})&=2(A_0^{*})(A_0\theta-f_0(X_{n,m}))+\sum_{i=1}^\numofsum\nabla _\theta((\|A_i\theta\|^2+\mu_i)^{r_i})f_i(X_{n,m})\\
     &=2(A_0)^{*}(A_0\theta-(\bfX_{n,m}^1,\bfX_{n,m}^2,\dots,\bfX_{n,m}^d))+\sum_{i=1}^\numofsum\nabla _\theta((\|A_i\theta\|^2+\mu_i)^{r_i})\bfX_{n,m}^{d+i}\\
     &=2(A_0)^{*}(A_0\theta-F_0(\bfX_{n,m}))+\sum_{i=1}^\numofsum\nabla _\theta((\|A_i\theta\|^2+\mu_i)^{r_i})F_i(\bfX_{n,m})\\
     &=(\nabla_\theta\cL)(\theta,\bfX_{n,m}).
     \end{split}
 \end{equation}}
 \argument{\lref{eq2};\lref{def: bbV};}{that for all $k\in \{1,2\}$, $M,n\in \N$, $i\in \{1,2,\dots,d\}$, $\beta=(\beta_1,\beta_2)\in (0,1)^2$ it holds that
\begin{equation}\llabel{def: bbV2}
     \mandVcom{k}{M}{\para}{n}{i}= \beta_k \mandVcom{k}{M}{\para}{n-1}{i}+(1-\beta_k)\bigl[\textstyle \frac 1M \sum_{m=1}^M(\nabla_{\theta_i} \cL)(\mandV{0}{M}{\beta}{n-1},\bfX_{n,m})\bigr]^k.
        \end{equation}}
        \argument{\lref{def: bbV2};\lref{def: Theta};}{for all for all $k\in\{1,2\}$, $M,n\in \N$, $i\in\{1,2,\dots,d\}$, $\beta=(\beta_1,\beta_2)\in (0,1)^2$ that
 \begin{equation}
    \mandV{k}{M}{\para}{0}=0,\qquad \mandVcom{k}{M}{\para}{n}{i}= \beta_k \mandVcom{k}{M}{\para}{n-1}{i}+(1-\beta_k)\bigl[\textstyle \frac 1M \sum_{m=1}^M(\nabla_{\theta_i} \cL)(\mandV{0}{M}{\beta}{n-1},\bfX_{n,m})\bigr]^k,
        \end{equation}
        \begin{equation}\llabel{eq3}
    \text{and}\qquad \mandVcom{0}{M}{\para}{n}{i}=\mandVcom{0}{M}{\para}{n-1}{i} -\gamma_n [1-(\para_1)^n]^{-1}\bigl[\varepsilon+\bigl[[1-(\para_2)^n]^{-1}\mandVcom{2}{M}{\para}{n}{i}\bigr]^{\nicefrac{1}{2}}\bigr]^{-1}\mandVcom{1}{M}{\para}{n}{i}.
     \end{equation}}
     \argument{\lref{evd1};\lref{eq3};\lref{Item 1};\lref{Item 2}}{\lref{item 2,item 3}\dott}
\end{aproof}
\renewcommand{\setX}{U}
 \begin{tcolorbox}[colback=white!95!gray,
                  colframe=black,
                  boxrule=0.5pt,
                  sharp corners,
                  enhanced,
                  breakable,
                 ]
    \begin{athm}{cor}{cor: priori bound stochastic Adam 2 non explosion}
      Let $\fd,\dimX,\numofsum,\dimofsum\in \N$, $q\in (0,1)$, $p,\varepsilon\in (0,\infty)$, let $A_0\in \R^{d\times d}$ be invertible, let $f_0\colon \R^\dimX\to\R^d$ be measurable and locally bounded, for every $i\in\N$ let $A_i\in \R^{\dimofsum\times d}$, $r_i\in [\nicefrac{1}{2},\nicefrac{3}{4})$, $\mu_i\in (0,\infty)$ and let $f_i\colon \R^\dimX\to[0,\infty)$ be measurable and locally bounded, let $\smalll=(\smalll(\theta,x))_{(\theta,x)\in \R^\fd\times \R^\dimX}\allowbreak\colon\R^\fd\times \R^\dimX\to\R$ satisfy for all $\theta\in \R^\fd$, $x\in \R^\dimX$ that 
     \begin{equation}\llabel{def: loss}
       \textstyle \smalll(\theta,x)=\| A_0 \theta - f_0( x ) \|^2 
    + 
    \sum_{ i = 1 }^\numofsum ( \| A_i \theta \|^2 + \mu_i )^{ r_i }f_i( x ) ,
    \end{equation}
    let $(\Omega,\cF,\P)$ be a probability space,
 let $(\gamma_n)_{n\in \N}\subseteq \R$ be non-increasing, assume $\lim_{n\to\infty}\gamma_n=0$,
     for every $n,m\in \N$ let $X_{n,m}\colon \Omega\to [-p,p]^\dimX$ be a random variable, for every $k,M,n\in \N_0$, $\para\in \R^2$ let 
    $
\mandV{k}{M}{\para}{n}=(\mandVcom{k}{M}{\para}{n}{1},\dots,\allowbreak \mandVcom{k}{M}{\para}{n}{\fd})\colon\allowbreak \Omega\to\R^\fd$ be a random variable, and assume for all $k\in \{1,2\}$, $M,n\in \N$, $i\in \{1,2,\dots,\fd\}$, $\para=(\para_1,\para_2)\in (0,1)^2$ that
    \begin{equation}\label{def: bbV: cor: priori bound stochastic Adam 2 non explosion}
    \mandV{k}{M}{\para}{0}=0,\qquad \mandVcom{k}{M}{\para}{n}{i}= \beta_k \mandVcom{k}{M}{\para}{n-1}{i}+(1-\beta_k)\bigl[\textstyle \frac 1M \sum_{m=1}^M(\nabla_{\theta_i} \smalll)(\mandV{0}{M}{\beta}{n-1},X_{n,m})\bigr]^k,
        \end{equation}
        \begin{equation}\llabel{def: Theta}
    \text{and}\qquad \mandVcom{0}{M}{\para}{n}{i}=\mandVcom{0}{M}{\para}{n-1}{i} -\gamma_n [1-(\para_1)^n]^{-1}\bigl[\varepsilon+\bigl[[1-(\para_2)^n]^{-1}\mandVcom{2}{M}{\para}{n}{i}\bigr]^{\nicefrac{1}{2}}\bigr]^{-1}\mandVcom{1}{M}{\para}{n}{i}.
     \end{equation}
 Then there exists $c\in \R$ such that for all $M,n \in \N$, $\beta=(\beta_1,\beta_2)  \in [q, 1 )^2$ with $ (\beta_1)^2 + q  \leq \beta_2 $ it holds that  
 \begin{equation}\llabel{conclude}
    \begin{split}
     \| \Theta_n^{0,M,\beta} \|\leq c(1+\|\Theta_0^{0,M,\beta}\|).
    \end{split} 
 \end{equation}
\end{athm}
\end{tcolorbox}
\begin{aproof}
Throughout this proof let $\fr,\fR\in \R$ satisfy $\fr=2\max\{r_1,r_2,\dots,r_\numofsum\}-1$ and $\fR=\max\{2\max\{r_1,r_2,\dots,r_\numofsum\}-1,\nicefrac{1}{4}\}$, let $\cL=(\cL(\theta,x))_{(\theta,x)\in \R^\fd\times \R^{d+\numofsum}}\colon \R^{d}\times \R^{d+\numofsum}\to\R$ satisfy for all $\theta\in \R^d$, $x=(x_1,\dots,x_{d+\numofsum})\in \R^{d+\numofsum}$ that
\begin{equation}\llabel{def: cL}
    \cL(\theta,x)=\textstyle\| A_0 \theta - (x_1,x_2,\ldots,x_d) \|^2 
    + 
    \sum\nolimits_{ i = 1 }^\numofsum  ( \| A_i \theta \|^2 + \mu_i )^{ r_i }x_{i+d},
\end{equation}
and for every $n,m\in \N$ 
let $\bbX_{n,m}\in \R^{d+\numofsum}$ satisfy
         \begin{equation}\llabel{def: bbX}
             \bbX_{n,m}=(f_0(X_{n,m}),f_1(X_{n,m}),\dots,f_\numofsum(X_{n,m})).
         \end{equation}
\argument{\lref{def: cL};\lref{def: bbX};\cref{lem: reconstruct};}{that
 \begin{enumerate}[label=(\roman*)]
     \item \llabel{item 1} it holds that $0\leq \fr<\nicefrac 12$, 
     \item \llabel{item 2} there exist $c,\kappa,\cK\in (0,\infty)$ which satisfy for all $a,b\in \R^\fd$, $x\in [-c,c]^{d}\times[0,c]^\numofsum$, $n,m\in \N$ that
 \begin{equation}\llabel{item 1.1}
     \|(\nabla_\theta\cL)(a,x)-(\nabla_\theta\cL)(b,x)\|\leq \cK\|a-b\|,\quad \spro{ a-b, (\nabla_\theta\cL)(a,x)- (\nabla_\theta\cL)(b,x)}\geq \kappa\|a-b\|^2,
 \end{equation}
  \begin{equation}\llabel{item 1.2}
     \bbX_{n,m}\in [-c,c]^d\times[0,c]^\numofsum,\quad\text{and}\quad \|(\nabla_x\cL)(a,x)-(\nabla_x\cL)(b,x)\|\leq \cK\|a-b\|(1+\|a\|^{\fr}+\|b\|^{\fr}),
 \end{equation}
 \item \llabel{item 3} it holds for all $k\in \{1,2\}$, $M,n\in \N$, $i\in\{1,2,\dots,d\}$, $\beta=(\beta_1,\beta_2)\in (0,1)^2$ that
 \begin{equation}
    \mandV{k}{M}{\para}{0}=0,\qquad \mandVcom{k}{M}{\para}{n}{i}= \beta_k \mandVcom{k}{M}{\para}{n-1}{i}+(1-\beta_k)\bigl[\textstyle \frac 1M \sum_{m=1}^M(\nabla_{\theta_i} \cL)(\mandV{0}{M}{\beta}{n-1},\bbX_{n,m})\bigr]^k,
        \end{equation}
        \begin{equation}
 \text{and}\qquad\mandVcom{0}{M}{\para}{n}{i}=\mandVcom{0}{M}{\para}{n-1}{i} -\gamma_n [1-(\para_1)^n]^{-1}\bigl[\varepsilon+\bigl[[1-(\para_2)^n]^{-1}\mandVcom{2}{M}{\para}{n}{i}\bigr]^{\nicefrac{1}{2}}\bigr]^{-1}\mandVcom{1}{M}{\para}{n}{i}.
     \end{equation}
 \end{enumerate}}
 \startnewargseq
 \argument{\lref{item 1.2};the fact that $\fR\geq \fr\geq 0$}{that there exists $\fC\in (0,\infty)$ such that for all $a,b\in  \R^d$, $x\in [-c,c]^d\times[0,c]^{\numofsum}$ it holds that
 \begin{equation}\llabel{eq2}
     \|(\nabla_x\cL)(a,x)-(\nabla_x\cL)(b,x)\|\leq \fC\|a-b\|(1+\|a\|^{\fR}+\|b\|^{\fR})
 \end{equation}}
\argument{\lref{eq2};\lref{item 1.1};\lref{item 2};\lref{item 3};the fact that $0<\fR<\nicefrac 12$;\cref{cor: priori bound stochastic Adam 1 non explosion};}{\lref{conclude}\dott}
\end{aproof}
 \begin{tcolorbox}[colback=white!95!gray,
                  colframe=black,
                  boxrule=0.5pt,
                  sharp corners,
                  enhanced,
                  breakable,
                 ]
\begin{athm}{cor}{cor: priori bound stochastic Adam 2 non explosion copy}
      Let $\fd\in \N$, $p,\varepsilon\in (0,\infty)$, $q\in (0,1)$, $r\in [\nicefrac{1}{2},\nicefrac{3}{4})$, let $A\in \R^{d\times d}$ be invertible, let $\smalll=(\smalll(\theta,x,y))_{(\theta,x,y)\in \R^\fd\times \R^{d}\times\R}\allowbreak\colon \R^\fd\times \R^{d}\times\R\to\R$ satisfy for all $\theta\in \R^\fd$, $x\in \R^d$, $y\in \R$ that 
     \begin{equation}\llabel{def: loss}
        \smalll(\theta,x,y)=\|A\theta-x\|^2+(\|\theta\|^2+1)^r |y|,
    \end{equation}
    let $(\Omega,\cF,\P)$ be a probability space
     for every $n,m\in \N$ let $X_{n,m}\colon \Omega\to [-p,p]^{d+1}$ be a random variable, let $(\gamma_n)_{n\in \N}\subseteq \R$ be non-increasing, assume $\lim_{n\to\infty}\gamma_n=0$, for every $k,M,n\in \N_0$, $\para\in \R^2$ let 
    $
\mandV{k}{M}{\para}{n}=(\mandVcom{k}{M}{\para}{n}{1},\dots,\allowbreak \mandVcom{k}{M}{\para}{n}{\fd})\colon\allowbreak \Omega\to\R^\fd$ be a random variable, and assume for all $k\in \{1,2\}$, $M,n\in \N$, $i\in \{1,2,\dots,\fd\}$, $\para=(\para_1,\para_2)\in (0,1)^2$ that
    \begin{equation}\label{def: bbV: cor: priori bound stochastic Adam 2 non explosion copy}
    \mandV{k}{M}{\para}{0}=0,\qquad \mandVcom{k}{M}{\para}{n}{i}= \beta_k \mandVcom{k}{M}{\para}{n-1}{i}+(1-\beta_k)\bigl[\textstyle \frac 1M \sum_{m=1}^M(\nabla_{\theta_i} \smalll)(\mandV{0}{M}{\beta}{n-1},X_{n,m})\bigr]^k,
        \end{equation}
        \begin{equation}\llabel{def: Theta}
    \text{and}\qquad \mandVcom{0}{M}{\para}{n}{i}=\mandVcom{0}{M}{\para}{n-1}{i} -\gamma_n [1-(\para_1)^n]^{-1}\bigl[\varepsilon+\bigl[[1-(\para_2)^n]^{-1}\mandVcom{2}{M}{\para}{n}{i}\bigr]^{\nicefrac{1}{2}}\bigr]^{-1}\mandVcom{1}{M}{\para}{n}{i}.
     \end{equation}
 Then there exists $c\in \R$ such that for all $M,n \in \N$, $\beta=(\beta_1,\beta_2)  \in [q, 1 )^2$ with $ (\beta_1)^2 + q  \leq \beta_2 $ it holds that  
 \begin{equation}\llabel{conclude}
    \begin{split}
     \| \Theta_n^{0,M,\beta} \|\leq c(1+\|\Theta_0^{0,M,\beta}\|).
    \end{split} 
 \end{equation}
\end{athm}
\end{tcolorbox}
\begin{aproof}
    \argument{\cref{cor: priori bound stochastic Adam 2 non explosion}}{\lref{conclude}\dott}
\end{aproof}
\section{Convergence rates for strongly convex SOPs}\label{sec: convergence of Adam}
In this section we combine the conditional error analysis for \Adam\ in \cite[Corollary 2.11]{DeDoArPhi2025} that relies on the assumption that \Adam\ does not diverge to infinity but stays bounded with the a priori bounds for \Adam\ established in \cref{cor: priori bound stochastic Adam 1 non explosion} in \cref{sec: non-explosion bound} above to develop in \cref{cor: convegence of Adam3} and its follow-up results (cf.\ \cref{cor: convegence of Adam2} and \cref{cor: convegence of Adam} below) an unconditioned error analysis for \Adam\ when applied to strongly convex \SOPs. \cref{main theorem} in \cref{sec: introduction} above is an immediate consequence of \cref{cor: convegence of Adam3}.
\subsection{Convergence rates for Adam without assuming L-smoothness}\label{subsec: convergence without L}\label{subsec: assume without L smooth}
 \begin{tcolorbox}[colback=white!95!gray,
                  colframe=black,
                  boxrule=0.5pt,
                  sharp corners,
                  enhanced,
                  breakable,
                 ]
\begin{athm}{theorem}{cor: convegence of Adam3}
    Let $( \Omega, \cF,\P )$ be a probability space, let $d, \dimX \in \N$, $\varepsilon,p\in (0,\infty)$, $q,r\in (0,\nicefrac{1}{2})$, let $O \subseteq \R^{ \dimX }$ be open, let $\setX \subseteq O$ be compact and convex, let $X_{ n,m } \colon \Omega \to \setX$, $(n,m ) \in \N^2$, be \iid\ random variables, let $\smalll=(\smalll(\theta,x))_{(\theta,x)\in \R^\fd\times O}\allowbreak\in C^2(\R^\fd\times O,\R)$ satisfy for all $\theta,\vartheta\in \R^\fd$, $x\in \setX$ that 
 \begin{equation}\llabel{assumption2}
\|(\nabla_\theta\smalll)(\theta,x)- (\nabla_\theta\smalll)(\vartheta,x)\|+(1+\|\theta\|+\|\vartheta\|)^{-r}\|(\nabla_x\smalll)(\theta,x)-(\nabla_x\smalll)(\vartheta,x)\|\leq p \|\theta-\vartheta\|,
 \end{equation}
 let $(\gamma_n)_{n\in \N}\subseteq (0,\infty)$ be non-increasing, for every $k,M,n\in \N_0$, $\para\in \R^2$ let 
    $
\mandV{k}{M}{\para}{n}=(\mandVcom{k}{M}{\para}{n}{1},\dots,\allowbreak \mandVcom{k}{M}{\para}{n}{\fd})\colon\allowbreak \Omega\to\R^\fd$ be a random variable, assume for all $k\in \{1,2\}$, $M,n\in \N$, $i\in \{1,2,\dots,\fd\}$, $\para=(\para_1,\para_2)\in (0,1)^2$ that
    \begin{equation}\llabel{def: bbV}
    \mandV{k}{M}{\para}{0}=0,\qquad \mandVcom{k}{M}{\para}{n}{i}= \beta_k \mandVcom{k}{M}{\para}{n-1}{i}+(1-\beta_k)\bigl[\textstyle \frac 1M \sum_{m=1}^M(\nabla_{\theta_i} \smalll)(\mandV{0}{M}{\beta}{n-1},X_{n,m})\bigr]^k,
        \end{equation}
        \begin{equation}\llabel{def: Theta}
     \mandVcom{0}{M}{\para}{n}{i}=\mandVcom{0}{M}{\para}{n-1}{i} -\gamma_n [1-(\para_1)^n]^{-1}\bigl[\varepsilon+\bigl[[1-(\para_2)^n]^{-1}\mandVcom{2}{M}{\para}{n}{i}\bigr]^{\nicefrac{1}{2}}\bigr]^{-1}\mandVcom{1}{M}{\para}{n}{i}, 
     \end{equation}
      and $\P( \| \mandV{0}{M}{\beta}{0} \| \leq p )=1$, assume for all $M\in \N$, $\para\in \R^2$ that $\mandV{0}{M}{\para}{0}$ and $(X_{n,m})_{(n,m)\in \N^2}$ are independent, assume 
       $\limsup_{n\to\infty}( ( \gamma_n )^{ - 2 } ( \gamma_n - \gamma_{ n + 1 } ) +\sum_{ m = n }^{ \infty } ( \gamma_m )^p ) =0$, assume for all $x \in \setX$ that $\R^d \ni \theta \mapsto \smalll( \theta, x ) - q \| \theta \|^2 \in \R$ is convex, and let $\globalmin \in \R^\fd$ satisfy $
           \E[\smalll(\globalmin,X_{1,1})]=\inf_{\theta\in \R^\fd}\E[\smalll(\theta,X_{1,1})]$. Then there exists $(\scrc_{\para})_{\para\in \R^2}\subseteq \R$ such that for all $M \in \N$, $n \in \N_0$, $\para = ( \para_1, \para_2 ) \in   [ q, 1 )^2$ with $( \beta_1 )^2 + q\leq \beta_2 $ it holds that
\begin{equation} \llabel{conclude}
   \bigl( \E\bigl[ \| \mandV{0}{M}{\para}{n} - \globalmin\|^p \bigr] \bigr)^{ 1/p } \leq  \scrc_0 M^{ - 1 } ( 1 - \beta_2 \mathbbm 1_{ [\scrc_0,\infty)} ( M ) ) + \scrc_{ \beta } ( \gamma_{ n + 1 } )^{ 1 / 2 }.
\end{equation}
\end{athm}
\end{tcolorbox}
\begin{aproof}
    \argument{\cref{cor: priori bound stochastic Adam 1 non explosion};\cite[Corollary 2.11]{DeDoArPhi2025}}{\lref{conclude}\dott}
\end{aproof}
\cref{main theorem} in the introduction is a direct consequence of \cref{cor: convegence of Adam3}. In the next result, \cref{cor: convegence of Adam4}, we illustrate \cref{main theorem} by means of the example \SOPs\ from \cref{cor: priori bound stochastic Adam 2 non explosion} above.
 \begin{tcolorbox}[colback=white!95!gray,
                  colframe=black,
                  boxrule=0.5pt,
                  sharp corners,
                  enhanced,
                  breakable,
                 ]
\begin{athm}{cor}{cor: convegence of Adam4}
    Let $\fd,\dimX,\numofsum,\dimofsum\in \N$, $q\in (0,1)$, $p,\varepsilon\in (0,\infty)$, $\xi\in \R^\fd$, let $A_0\in \R^{d\times d}$ be invertible, let $f_0\colon \R^\dimX\to\R^d$ be measurable and locally bounded, for every $i\in\N$ let $r_i\in [\nicefrac{1}{2},\nicefrac{3}{4})$, $A_i\in \R^{\dimofsum\times d}$, $\mu_i\in (0,\infty)$ and let $f_i\colon\R^\dimX\to[0,\infty)$ be measurable and locally bounded, let $( \Omega, \cF,\P )$ be a probability space, 
     let $X_{ n,m } \colon \Omega \to [-p,p]^\dimX$, $(n,m ) \in \N^2$, be \iid\ random variables, let $\smalll=(\smalll(\theta,x))_{(\theta,x)\in \R^\fd\times \R^\dimX}\colon\R^\fd\times \R^\dimX\to\R$ satisfy for all $\theta\in \R^\fd$, $x\in \R^\dimX$ that 
     \begin{equation}\llabel{def: loss}
        \textstyle \smalll(\theta,x)=\| A_0 \theta - f_0( x ) \|^2 
    + 
    \sum\nolimits_{ i = 1 }^\numofsum  ( \| A_i \theta \|^2 + \mu_i )^{ r_i }f_i( x),
    \end{equation}
 let $(\gamma_n)_{n\in \N}\subseteq (0,\infty)$ be non-increasing, for every $M\in \N$, $\para=(\para_1,\para_2)\in (0,1)^2$ let 
    $
\mandV{k}{M}{\para}{n}=(\mandVcom{k}{M}{\para}{n}{1},\dots,\mandVcom{k}{M}{\para}{n}{\fd})\colon\allowbreak \Omega\to\R^\fd$, $(k,n)\in (\N_0)^2$, satisfy for all $k\in \{1,2\}$, $n\in \N$, $i\in \{1,2,\dots,\fd\}$ that
    \begin{equation}\llabel{def: bbV}
    \mandV{k}{M}{\para}{0}=0,\qquad \mandVcom{k}{M}{\para}{n}{i}= \beta_k \mandVcom{k}{M}{\para}{n-1}{i}+(1-\beta_k)\bigl[\textstyle \frac 1M \sum_{m=1}^M(\nabla_{\theta_i} \smalll)(\mandV{0}{M}{\beta}{n-1},X_{n,m})\bigr]^k,
        \end{equation}
        \begin{equation}\llabel{def: Theta}
     \mandVcom{0}{M}{\para}{n}{i}=\mandVcom{0}{M}{\para}{n-1}{i} -\gamma_n [1-(\para_1)^n]^{-1}\bigl[\varepsilon+\bigl[[1-(\para_2)^n]^{-1}\mandVcom{2}{M}{\para}{n}{i}\bigr]^{\nicefrac{1}{2}}\bigr]^{-1}\mandVcom{1}{M}{\para}{n}{i}, 
     \end{equation}
      and $\mandV{0}{M}{\beta}{0} =\xi$, assume 
       $\limsup_{n\to\infty}( ( \gamma_n )^{ - 2 } ( \gamma_n - \gamma_{ n + 1 } ) +\sum_{ m = n }^{ \infty } ( \gamma_m )^p ) =0$, and let $\globalmin \in \R^\fd$ satisfy $
           \E[\smalll(\globalmin,X_{1,1})]=\inf_{\theta\in \R^\fd}\E[\smalll(\theta,X_{1,1})]$. Then there exists $(\scrc_{\para})_{\para\in \R^2}\subseteq \R$ such that for all $M ,n\in \N$, $\para = ( \para_1, \para_2 ) \in   [ q, 1 )^2$ with $( \beta_1 )^2 + q \leq \beta_2 $ it holds that
\begin{equation} \llabel{conclude}
   \bigl( \E\bigl[ \| \mandV{0}{M}{\para}{n} - \globalmin\|^p \bigr] \bigr)^{ 1/p } \leq  \scrc_0 M^{ - 1 } ( 1 - \beta_2 \mathbbm 1_{ [\scrc_0,\infty)} ( M ) ) + \scrc_{ \beta } ( \gamma_{ n } )^{ 1 / 2 }.
\end{equation}
\end{athm}
\end{tcolorbox}
\renewcommand{\setX}{V}
\begin{aproof}
Throughout this proof let $\fr,\fR\in \R$ satisfy $\fr=2\max\{r_1,r_2,\dots,r_\numofsum\}-1$ and $\fR=\max\{2\max\{r_1,r_2,\dots,r_\numofsum\}-1,\nicefrac{1}{4}\}$, let $\cL=(\cL(\theta,x))_{(\theta,x)\in \R^\fd\times \R^{d+\numofsum}}\colon \R^{d}\times \R^{d+\numofsum}\to\R$ satisfy for all $\theta\in \R^d$, $x=(x_1,\dots,x_{d+\numofsum})\in \R^{d+\numofsum}$ that
\begin{equation}\llabel{def: cL}
    \cL(\theta,x)=\textstyle\| A_0 \theta - (x_1,x_2,\ldots,x_d) \|^2 
    + 
    \sum\nolimits_{ i = 1 }^\numofsum  ( \| A_i \theta \|^2 + \mu_i )^{ r_i }x_{i+d},
\end{equation}
and for every $n,m\in \N$ 
let $\bbX_{n,m}\colon \Omega\to \R^{d+\numofsum}$ satisfy
         \begin{equation}\llabel{def: bbX}
             \bbX_{n,m}=(f_0(X_{n,m}),f_1(X_{n,m}),\dots,f_\numofsum(X_{n,m})).
         \end{equation}
  \argument{\lref{def: cL};\lref{def: bbX};\cref{lem: reconstruct};}{that there exist $c,\kappa,\cK\in (0,\infty)$ which satisfy that
 \begin{enumerate}[label=(\roman*)]
     \item \llabel{item 1} it holds that $0\leq \fr<\nicefrac 12$, 
     \item \llabel{item 2} there exist $c,\kappa,\cK\in (0,\infty)$ which satisfy for all $a,b\in \R^\fd$, $x\in [-c,c]^{d}\times[0,c]^\numofsum$, $n,m\in \N$ that
 \begin{equation}\llabel{conclude 1}
     \|(\nabla_\theta\cL)(a,x)-(\nabla_\theta\cL)(b,x)\|\leq \cK\|a-b\|,\quad \spro{ a-b, (\nabla_\theta\cL)(a,x)- (\nabla_\theta\cL)(b,x)}\geq \kappa\|a-b\|^2,
 \end{equation}
  \begin{equation}\llabel{conclude 2}
     \bbX_{n,m}\in [-c,c]^d\times[0,c]^\numofsum,\quad\text{and}\quad \|(\nabla_x\cL)(a,x)-(\nabla_x\cL)(b,x)\|\leq \cK\|a-b\|(1+\|a\|^{\fr}+\|b\|^{\fr}),
 \end{equation}
 \item \llabel{item 3} it holds for all $k\in \{1,2\}$, $M,n\in \N$, $i\in\{1,2,\dots,d\}$, $\beta=(\beta_1,\beta_2)\in (0,1)^2$ that
 \begin{equation}
    \mandV{k}{M}{\para}{0}=0,\qquad \mandVcom{k}{M}{\para}{n}{i}= \beta_k \mandVcom{k}{M}{\para}{n-1}{i}+(1-\beta_k)\bigl[\textstyle \frac 1M \sum_{m=1}^M(\nabla_{\theta_i} \cL)(\mandV{0}{M}{\beta}{n-1},\bbX_{n,m})\bigr]^k,
        \end{equation}
        \begin{equation}
 \text{and}\qquad\mandVcom{0}{M}{\para}{n}{i}=\mandVcom{0}{M}{\para}{n-1}{i} -\gamma_n [1-(\para_1)^n]^{-1}\bigl[\varepsilon+\bigl[[1-(\para_2)^n]^{-1}\mandVcom{2}{M}{\para}{n}{i}\bigr]^{\nicefrac{1}{2}}\bigr]^{-1}\mandVcom{1}{M}{\para}{n}{i}.
     \end{equation}
 \end{enumerate}}
\startnewargseq
 \argument{\lref{conclude 2};the fact that $\fR\geq \fr\geq 0$}{that there exists $\fC\in (0,\infty)$ such that for all $a,b\in  \R^d$, $x\in [-c,c]^d\times[0,c]^{\numofsum}$ it holds that
 \begin{equation}\llabel{eq2}
     \|(\nabla_x\cL)(a,x)-(\nabla_x\cL)(b,x)\|\leq \fC\|a-b\|(1+\|a\|^{\fR}+\|b\|^{\fR}).
 \end{equation}}
 \argument{\lref{conclude 1};\lref{eq2};the fact that $\cL\in C^2(\R^d\times\R^{d+\numofsum},\R)$}{that there exists $\fC\in (0,\infty)$ which satisfies for all $\theta\in \R^\fd$, $x\in [-c,c]^d\times [0,c]^\numofsum$ that 
 \begin{equation}\llabel{eq1}
\textstyle\sum_{i=1}^d\bigl(\|(\nabla_{\theta_i}\nabla_\theta\cL)(\theta,x)\|+(1+\|\theta\|)^{-\fR}\|(\nabla_{\theta_i}\nabla_x\cL)(\theta,x)\|\bigr)\leq \fC.
 \end{equation} }
 \startnewargseq
\argument{\lref{def: bbX};the fact that for all $i\in \{0,1,\dots,\numofsum\}$ it holds that $f_\numofsum$ is measurable}{that $\bbX_{n,m}$, $(n,m)\in \N^2$, are \iid\ }
\argument{\lref{conclude 1};\cite[Proposition 5.7.23]{ArBePhi2024}}{that for all $x\in [-c,c]^d\times [0,c]^\numofsum$ it holds that \llabel{arg1} $\R^d \ni \theta \mapsto \cL( \theta, x ) - \frac{\kappa}{2} \| \theta \|^2 \in \R$ is convex\dott}
    \argument{\lref{arg1};\lref{item 2};\lref{item 3};\lref{eq1};the fact that $\bbX_{n,m}$, $(n,m)\in \N^2$, are \iid; the fact that $0<\fR<\nicefrac{1}{2}$;\cref{main theorem} (applied with $\Omega\curvearrowleft\Omega$, $\cF\curvearrowleft\cF$, $\P\curvearrowleft\P$, $d\curvearrowleft d$, $\dimX\curvearrowleft d+\numofsum$, $\varepsilon\curvearrowleft\varepsilon$, $p\curvearrowleft \max\{p,\fC\}$, $q\curvearrowleft\min\{ \frac{\kappa}{2},q,\nicefrac{1}{4}\}$, $r\curvearrowleft\fR$, $\xi\curvearrowleft\xi$, $(\gamma_n)_{n\in \N}\curvearrowleft(\gamma_n)_{n\in \N}$, $U\curvearrowleft [-c,c]^d\times [0,c]^\numofsum$, $\smalll\curvearrowleft\cL$, $(X_{n,m})_{(n,m)\in \N^2}\curvearrowleft (\bbX_{n,m})_{(n,m)\in \N^2}$, $\vartheta\curvearrowleft\vartheta$ in the notation of \cref{main theorem})}{that there exists $(\scrc_{\para})_{\para\in \R^2}\allowbreak\subseteq \R$ such that for all $M ,n\in \N$, $\para = ( \para_1, \para_2 ) \in   [ \min\{ \frac{\kappa}{2},q,\nicefrac{1}{4}\}, 1 )^2$ with $( \beta_1 )^2 + \min\{ \kappa,q,\nicefrac{1}{4}\} \leq \beta_2 $ it holds that
\begin{equation} \llabel{conclude'}
   \bigl( \E\bigl[ \| \mandV{0}{M}{\para}{n} - \globalmin\|^{\max\{p,\fC\}} \bigr] \bigr)^{ 1/(\max\{p,\fC\})} \leq  \scrc_0 M^{ - 1 } ( 1 - \beta_2 \mathbbm 1_{ [\scrc_0,\infty)} ( M ) ) + \scrc_{ \beta } ( \gamma_{ n } )^{ 1 / 2 }.
\end{equation}}
\argument{\lref{conclude'};}{\lref{conclude}\dott}
\end{aproof}
\renewcommand{\setX}{K}
\subsection{Convergence rates for Adam with assuming L-smoothness}
In the next result, \cref{cor: convegence of Adam2}, we specialize \cref{cor: convegence of Adam3} to the situation where the gradient of the loss function $\smalll \colon\R^d\times O \to\R$ in \cref{cor: convegence of Adam3} is Lipschitz continuous.
 \begin{tcolorbox}[colback=white!95!gray,
                  colframe=black,
                  boxrule=0.5pt,
                  sharp corners,
                  enhanced,
                  breakable,
                 ]
\begin{athm}{cor}{cor: convegence of Adam2}
    Let $( \Omega, \cF,\P )$ be a probability space, let $d, \dimX \in \N$, $\varepsilon,p,q \in (0,\infty)$, let $O \subseteq \R^{ \dimX }$ be open, let $\setX \subseteq O$ be compact and convex, let $X_{ n,m } \colon \Omega \to \setX$, $(n,m ) \in \N^2$, be \iid\ random variables, let $\smalll=(\smalll(\theta,x))_{(\theta,x)\in \R^\fd\times O}\allowbreak\in C^2(\R^\fd\times O,\R)$ have a Lipschitz continuous first derivative, let $(\gamma_n)_{n\in \N}\subseteq (0,\infty)$ be non-increasing, for every $k\in \{0,1,2\}$, $M,n\in \N_0$, $\para\in \R^2$ let 
    $
\mandV{k}{M}{\para}{n}=(\mandVcom{k}{M}{\para}{n}{1},\dots,\mandVcom{k}{M}{\para}{n}{\fd})\colon\allowbreak \Omega\to\R^\fd$ be a random variable, assume for all $k\in \{1,2\}$, $M,n\in \N$, $i\in \{1,2,\dots,\fd\}$, $\para=(\para_1,\para_2)\in (0,1)^2$ that
    \begin{equation}\llabel{def: bbV}
    \mandV{k}{M}{\para}{0}=0,\qquad \mandVcom{k}{M}{\para}{n}{i}= \beta_k \mandVcom{k}{M}{\para}{n-1}{i}+(1-\beta_k)\bigl[\textstyle \frac 1M \sum_{m=1}^M(\nabla_{\theta_i} \smalll)(\mandV{0}{M}{\beta}{n-1},X_{n,m})\bigr]^k,
        \end{equation}
        \begin{equation}\llabel{def: Theta}
     \mandVcom{0}{M}{\para}{n}{i}=\mandVcom{0}{M}{\para}{n-1}{i} -\gamma_n [1-(\para_1)^n]^{-1}\bigl[\varepsilon+\bigl[[1-(\para_2)^n]^{-1}\mandVcom{2}{M}{\para}{n}{i}\bigr]^{\nicefrac{1}{2}}\bigr]^{-1}\mandVcom{1}{M}{\para}{n}{i}, 
     \end{equation}
      and $\P( \| \mandV{0}{M}{\beta}{0} \| \leq p )=1$, assume for all $M\in \N$, $\para\in \R^2$ that $\mandV{0}{M}{\para}{0}$ and $(X_{n,m})_{(n,m)\in \N^2}$ are independent, assume 
       $\limsup_{n\to\infty}( ( \gamma_n )^{ - 2 } ( \gamma_n - \gamma_{ n + 1 } ) +\sum_{ m = n }^{ \infty } ( \gamma_m )^p ) =0$, assume for all $x \in \setX$ that $\R^d \ni \theta \mapsto \smalll( \theta, x ) - q \| \theta \|^2 \in \R$ is convex, and let $\globalmin \in \R^\fd$ satisfy $
           \E[\smalll(\globalmin,X_{1,1})]=\inf_{\theta\in \R^\fd}\E[\smalll(\theta,X_{1,1})]$. Then there exists $(\scrc_{\para})_{\para\in \R^2}\subseteq \R$ such that for all $M \in \N$, $n \in \N_0$, $\para = ( \para_1, \para_2 ) \in   [ q, 1 )^2$ with $( \beta_1 )^2 + q \leq \beta_2 $ it holds that
\begin{equation} \llabel{conclude}
   \bigl( \E\bigl[ \| \mandV{0}{M}{\para}{n} - \globalmin\|^p \bigr] \bigr)^{ 1/p } \leq  \scrc_0 M^{ - 1 } ( 1 - \beta_2 \mathbbm 1_{ [\scrc_0,\infty)} ( M ) ) + \scrc_{ \beta } ( \gamma_{ n + 1 } )^{ 1 / 2 }.
\end{equation}
\end{athm}
\end{tcolorbox}
\begin{aproof}
    \argument{\cref{cor: convegence of Adam3};}{\lref{conclude}\dott}
\end{aproof}

In the next result, \cref{cor: convegence of Adam}, we specialize \cref{cor: convegence of Adam2} to the situation where the \Adam\ optimization process $( \Theta^{ 0, M, \beta }_n )_{ n \in \N_0 }$ starts at a fixed deterministic vector $\xi \in \R^d$ instead of at a random variable as in \cref{cor: convegence of Adam2}.
 \begin{tcolorbox}[colback=white!95!gray,
                  colframe=black,
                  boxrule=0.5pt,
                  sharp corners,
                  enhanced,
                  breakable,
                 ]
\begin{athm}{cor}{cor: convegence of Adam}
    Let $( \Omega, \cF,\P )$ be a probability space, let $d, \dimX \in \N$, $\xi\in \R^d$, $\varepsilon,p,q\in (0,\infty)$, let $X_{ n,m } \colon \Omega \to[-p,p]^\dimX$, $(n,m ) \in \N^2$, be \iid\ random variables, let $\smalll=(\smalll(\theta,x))_{(\theta,x)\in \R^\fd\times \R^\dimX}\allowbreak\in C^2(\R^\fd\times \R^\dimX,\R)$ have a bounded second derivative, let $(\gamma_n)_{n\in \N}\subseteq (0,\infty)$ be non-increasing, for every $M\in \N$, $\para=(\para_1,\para_2)\in (0,1)^2$ let 
    $
\mandV{k}{M}{\para}{n}=(\mandVcom{k}{M}{\para}{n}{1},\dots,\mandVcom{k}{M}{\para}{n}{\fd})\colon\allowbreak \Omega\to\R^\fd$, $(k,n)\in (\N_0)^2$, satisfy for all $k\in \{1,2\}$, $n\in \N$, $i\in \{1,2,\dots,\fd\}$ that
    \begin{gather}
    \mandV{k}{M}{\para}{0}=0,\quad \mandVcom{k}{M}{\para}{n}{i}= \beta_k \mandVcom{k}{M}{\para}{n-1}{i}+(1-\beta_k)\bigl[\textstyle \frac 1M \sum_{m=1}^M(\nabla_{\theta_i} \smalll)(\mandV{0}{M}{\beta}{n-1},X_{n,m})\bigr]^k, \llabel{def: Theta}\\
      \mandVcom{0}{M}{\para}{n}{i}=\mandVcom{0}{M}{\para}{n-1}{i} -\gamma_n [1-(\para_1)^n]^{-1}\bigl[\varepsilon+\bigl[[1-(\para_2)^n]^{-1}\mandVcom{2}{M}{\para}{n}{i}\bigr]^{\nicefrac{1}{2}}\bigr]^{-1}\mandVcom{1}{M}{\para}{n}{i},
     \end{gather}
     and  $\mandV{0}{M}{\para}{0}=\xi$,
     assume
       $\limsup_{n\to\infty}( ( \gamma_n )^{ - 2 } ( \gamma_n- \gamma_{ n + 1 } ) +\sum_{ m = n }^{ \infty } ( \gamma_m)^p ) =0$, assume for all $x \in \R^\dimX$ that $\R^d \ni \theta \mapsto \smalll( \theta, x ) - q \| \theta \|^2 \in \R$ is convex, and let $\globalmin \in \R^\fd$ satisfy $
           \E[\smalll(\globalmin,X_{1,1})]=\inf_{\theta\in \R^\fd}\E[\smalll(\theta,X_{1,1})]$.
       Then there exists $(\scrc_{\para})_{\para\in \R^2}\subseteq \R$ such that for all $M,n \in \N$, $\para = ( \para_1, \para_2 ) \in   [ q, 1 )^2$ with $( \beta_1 )^2 + q \leq \beta_2 $ it holds that
      
\begin{equation} \llabel{conclude}
  \bigl( \E\bigl[ \| \mandV{0}{M}{\para}{n} - \globalmin\|^p \bigr] \bigr)^{ 1/p } \leq  \scrc_0 M^{ - 1 } ( 1 - \beta_2 \mathbbm 1_{ [\scrc_0,\infty)} ( M ) ) + \scrc_{ \beta }  \sqrt{ \gamma_{ n  } }.
\end{equation}
\end{athm}
\end{tcolorbox}
\begin{aproof}
    \argument{\cref{cor: convegence of Adam2}}{\lref{conclude}\dott}
\end{aproof}
\subsubsection*{Acknowledgements}
This work has been partially supported by the National Science Foundation of China (NSFC) under grant number W2531010. Moreover,
this work has been partially funded by the Deutsche Forschungsgemeinschaft (DFG, German Research Foundation) under Germany’s Excellence Strategy EXC 2044/2-390685587, Mathematics Münster: Dynamics-Geometry-Structure. Furthermore, this work has been partially funded by the European Union (ERC, MONTECARLO, 101045811). The views and the opinions expressed in this work are however those of the authors only and do not necessarily reflect those of the European Union or the European Research Council (ERC). Neither the European Union nor the granting authority can be held responsible for them. Most of the specific formulations in the proofs of this work have been created using \cite{Bennoargumentcommand}.

\subsection*{Use of large language models}

{\sc GPT 5.6 Pro} have helped us to reveal several inaccuracies and errors within this manuscript. The entire article (including all statements and all arguments in this article) has been carefully written by the authors and the authors take full responsibility for each of the sentences/statements made in the paper.
\bibliographystyle{acm}
\bibliography{bibfileAdamprioriI}

\end{document}